%% file: main.tex
\documentclass[nohyperref]{article}

\usepackage[accepted]{icml2022}

\input{math_commands.tex}

\usepackage{url}

\usepackage{xcolor}
\usepackage{hyperref}
\definecolor{cite_color}{HTML}{114083}
\definecolor{link_color}{RGB}{153, 0,0} 
\definecolor{url_color}{RGB}{153, 102,  0}
\definecolor{emp_color}{RGB}{0,0,255}

\hypersetup{
 colorlinks,
 citecolor=cite_color,
 linkcolor=link_color,
 urlcolor=url_color}

\usepackage{amsmath}
\usepackage{amssymb}
\usepackage{amsthm}
\usepackage{mathtools}
\usepackage{amscd}
\usepackage{amsfonts}
\usepackage{graphicx}
\usepackage{tabularborder}
\usepackage{enumerate}
\usepackage{bm}
\usepackage{natbib}
\usepackage{multirow}
\usepackage{algorithm}
\usepackage{algorithmic}
\usepackage{fancyhdr}
\usepackage{subcaption}
\usepackage{wrapfig}
\usepackage{framed}
\usepackage{color}
\definecolor{shadecolor}{rgb}{0.94, 0.97, 1.0}

\usepackage[capitalize,noabbrev]{cleveref}

\theoremstyle{plain}

\newtheorem{theorem}{Theorem}
\newtheorem{proposition}{Proposition}

\theoremstyle{definition}
\newtheorem{definition}{Definition}

\theoremstyle{remark}
\newtheorem{remark}{Remark}

\usepackage{cleveref} 
\crefname{section}{Sec.}{Sections}
\crefname{theorem}{Theorem}{Theorems}
\crefname{lemma}{Lemma}{Lemmas}
\crefname{equation}{Eq.}{Equations}
\crefname{proposition}{Proposition}{Propositions}
\crefname{claim}{Claim}{Claims}
\crefname{appendix}{Appendix}{Appendices}
\crefname{algorithm}{Alg.}{Algorithms}
\crefname{figure}{Fig.}{Figures}
\crefname{table}{Table}{Tables}
\crefname{remark}{Remark}{Remarks}
\crefname{definition}{Def.}{Definitions}
\crefname{corollary}{Corollary}{Corollaries}

\usepackage[textsize=tiny]{todonotes}

\usepackage{xspace}
\newcommand{\pgnn}{$^p$GNN\xspace}
\newcommand{\pgnns}{$^p$GNNs\xspace}
\newcommand{\appendixtitle}[1]{
	\begin{center}
		\LARGE \bf #1
	\end{center}
}

\usepackage{etoc}
\etocdepthtag.toc{mtchapter}
\etocsettagdepth{mtchapter}{subsection}
\etocsettagdepth{mtappendix}{none}

\icmltitlerunning{$p$-Laplacian Based Graph Neural Networks}

\begin{document}

\twocolumn[
\icmltitle{$p$-Laplacian Based Graph Neural Networks}




\begin{icmlauthorlist}
\icmlauthor{Guoji Fu}{tencent}
\icmlauthor{Peilin Zhao}{tencent}
\icmlauthor{Yatao Bian}{tencent}
\end{icmlauthorlist}

\icmlaffiliation{tencent}{Tencent AI Lab, Shenzhen, China}

\icmlcorrespondingauthor{Guoji Fu}{guoji.leo.fu@gmail.com}
\icmlcorrespondingauthor{Yatao Bian}{yatao.bian@gmail.com}

\icmlkeywords{Graph Neural Networks, $p$-Laplacian}

\vskip 0.3in
]



\printAffiliationsAndNotice{}

    \input{sections/0-Abs}
	
	\input{sections/1-Intro}
	\input{sections/2-Pre}
	\input{sections/3-pGNN}
	\input{sections/4-Analysis}
	\input{sections/5-Exp}

	\input{sections/6-Conclusion}
	
	\input{sections/6.1-Acknowledge}
	\input{sections/6.2-Statement}
	\bibliography{reference}
    \bibliographystyle{icml2022}
	
	\clearpage
	\appendix
	\onecolumn
	\appendixtitle{Appendix}
	
	\etocdepthtag.toc{mtappendix}
    \etocsettagdepth{mtchapter}{none}
    \etocsettagdepth{mtappendix}{subsection}
    
    {
        \hypersetup{linkcolor=black}
    	\footnotesize\tableofcontents
    }
    \clearpage
	\input{sections/7-App_related}
	\input{sections/8-App_discussion}
	\input{sections/9-App_theorem}
	\input{sections/10-App_exp}

\end{document}

%% file: math_commands.tex
\usepackage{amsmath,amsfonts,bm}

\def\eqref#1{equation~\ref{#1}}

\def\1{\bm{1}}

\def\rmM{{\mathbf{M}}}

\def\vu{{\mathbf{u}}}
\def\vv{{\mathbf{v}}}

\def\vx{{\mathbf{x}}}
\def\vy{{\mathbf{y}}}

\def\evu{{u}}

\def\evx{{x}}
\def\evy{{y}}

\def\mD{{\mathbf{D}}}

\def\mF{{\mathbf{F}}}

\def\mI{{\mathbf{I}}}

\def\mM{{\mathbf{M}}}

\def\mU{{\mathbf{U}}}

\def\mW{{\mathbf{W}}}
\def\mX{{\mathbf{X}}}

\def\mZ{{\mathbf{Z}}}

\def\mLambda{{\mathbf{\Lambda}}}

\DeclareMathAlphabet{\mathsfit}{\encodingdefault}{\sfdefault}{m}{sl}
\SetMathAlphabet{\mathsfit}{bold}{\encodingdefault}{\sfdefault}{bx}{n}

\def\gE{{\mathcal{E}}}
\def\gF{{\mathcal{F}}}
\def\gG{{\mathcal{G}}}
\def\gH{{\mathcal{H}}}

\def\gL{{\mathcal{L}}}
\def\gM{{\mathcal{M}}}
\def\gN{{\mathcal{N}}}

\def\gS{{\mathcal{S}}}

\def\gV{{\mathcal{V}}}

\def\sN{{\mathbb{N}}}

\def\emA{{A}}

\def\emD{{D}}

\def\emM{{M}}

\def\emW{{W}}

\def\emZ{{Z}}

\newcommand{\E}{\mathbb{E}}

\newcommand{\R}{\mathbb{R}}

\DeclareMathOperator*{\argmax}{arg\,max}
\DeclareMathOperator*{\argmin}{arg\,min}

%% file: sections/0-Abs.tex
\begin{abstract}
	Graph neural networks (GNNs) have demonstrated superior performance for semi-supervised node classification on graphs, as a result of their ability to exploit node features and topological information. However, most GNNs implicitly assume that the labels of nodes and their neighbors in a graph are the same or consistent, which does not hold in heterophilic graphs, where the labels of linked nodes are likely to differ. Moreover, when the topology is non-informative for label prediction, ordinary GNNs may work significantly worse than simply applying multi-layer perceptrons (MLPs) on each node. 
	To tackle the above problem, we propose a new $p$-Laplacian based GNN model, termed as \pgnn, whose message passing mechanism is derived from a discrete regularization framework and can be theoretically explained as an approximation of a polynomial graph filter defined on the spectral domain of $p$-Laplacians. The spectral analysis shows that the new message passing mechanism works as low-high-pass filters, thus rendering \pgnns  effective on both homophilic and heterophilic graphs. 
	Empirical studies on real-world and synthetic datasets validate our findings and demonstrate that \pgnns significantly outperform several state-of-the-art GNN architectures on heterophilic benchmarks while achieving competitive performance on homophilic benchmarks. Moreover, \pgnns can adaptively learn aggregation weights and are robust to noisy edges.
\end{abstract}

%% file: sections/1-Intro.tex
\section{Introduction}\label{sec:sec1}

In this paper, we explore  the usage of graph neural networks (GNNs) for semi-supervised node classification on graphs, especially when the graphs admit strong heterophily or noisy edges.  
Semi-supervised learning on graphs is ubiquitous in a lot of real-world scenarios, such as user classification in social media~\citep{DBLP:conf/iclr/KipfW17}, protein classification in biology~\citep{DBLP:conf/iclr/VelickovicCCRLB18}, molecular property prediction in chemistry~\citep{,DBLP:conf/nips/DuvenaudMABHAA15}, and many others~\citep{DBLP:conf/emnlp/MarcheggianiT17,DBLP:conf/iclr/SatorrasE18}. 
Recently, GNNs have become the de facto choice for processing graph structured data. They can exploit the node features and the graph topology by propagating and transforming the features over the topology in each layer and thereby learn refined node representations.
A series of GNN architectures have been proposed, including graph convolutional networks~\citep{DBLP:journals/iclr/BrunaZSL14,DBLP:journals/corr/HenaffBL15,DBLP:conf/icml/NiepertAK16,DBLP:conf/iclr/KipfW17,DBLP:conf/icml/WuSZFYW19}, 
graph attention networks~\citep{DBLP:conf/iclr/VelickovicCCRLB18,DBLP:journals/corr/abs-1803-03735}, 
and  other representatives~\citep{DBLP:conf/nips/HamiltonYL17,DBLP:conf/icml/XuLTSKJ18,DBLP:conf/iclr/PeiWCLY20}.

Most of the existing GNN architectures work under the homophily assumption, i.e. the labels of nodes and their neighbors in a graph are the same or consistent, which is also commonly used in graph clustering~\citep{bach2004learning,DBLP:journals/sac/Luxburg07,DBLP:books/crc/aggarwal13/LiuH13} and semi-supervised learning on graphs~\citep{DBLP:conf/colt/BelkinMN04,DBLP:conf/colt/Hein06,nadler2009semi}. 
%
However, recent studies~\citep{DBLP:conf/nips/ZhuYZHAK20,DBLP:conf/aaai/ZhuR0MLAK21,DBLP:conf/iclr/ChienP0M21} show that in contrast to their success on homophilic graphs, most GNNs fail to work well on heterophilic graphs, in which linked nodes are more likely to have distinct labels. 
Moreover, GNNs could even fail on graphs where their topology is not helpful for label prediction. In these cases, propagating and transforming node features over the graph topology could lead to worse performance than simply applying multi-layer perceptrons (MLPs) on each of the nodes independently. 
Several recent works were proposed to deal with the heterophily issues of GNNs. 
\citet{DBLP:conf/nips/ZhuYZHAK20} finds that heuristically combining ego-, neighbor, and higher-order embeddings improves GNN performance on heterophilic graphs. \citet{DBLP:conf/aaai/ZhuR0MLAK21} uses a compatibility matrix to model the graph homophily or heterphily level. \citet{DBLP:conf/iclr/ChienP0M21} incorporates the generalized PageRank algorithm with graph convolutions so as to jointly optimize node feature and topological information extraction for both homophilic and heterophilic graphs.
However, the problem of GNNs on graphs with non-informative topologies (or noisy edges) remains open.


Unlike previous works, we tackle the above issues of GNNs by proposing the discrete $p$-Laplacian based message passing scheme, termed as $p$-Laplacian message passing.
It is derived from a discrete regularization framework and is theoretically verified as an approximation of a polynomial graph filter defined on the spectral domain of the $p$-Laplacian. 
Spectral analysis of $p$-Laplacian message passing shows that it works as low-high-pass filters\footnote{Note that if the low frequencies and high frequencies dominate the middle frequencies (the frequencies around the cutoff frequency), we say that the filter works as a low-high-pass filter.} and thus is applicable to  both homophilic and heterophilic graphs. Moreover, when $p \neq 2$, our theoretical results indicate that it can adaptively learn aggregation weights in terms of the variation of node embeddings on edges (measured by the graph gradient~\citep{DBLP:journals/arscom/Amghibech03,DBLP:conf/dagm/ZhouS05,DBLP:journals/ml/LuoHDN10}), and work as low-pass or low-high-pass filters on a node according to the local variation of node embeddings around the node (measured by the norm of graph gradients). 

Based on $p$-Laplacian message passing, we propose a new GNN architecture, called \pgnn, to enable GNNs to work with  heterophilic graphs and graphs with non-informative topologies. 
Several existing GNN architectures, including SGC~\citep{DBLP:conf/icml/WuSZFYW19}, APPNP~\citep{DBLP:conf/iclr/KlicperaBG19} and GPRGNN~\citep{DBLP:conf/iclr/ChienP0M21}, can be shown to be analogical to \pgnn with $p = 2$. 
Our empirical studies on real-world benchmark datasets (homophilic and heterophilic datasets) and synthetic datasets (cSBM~\citep{DBLP:conf/nips/DeshpandeSMM18}) demonstrate that \pgnns obtain the best performance on heterophilic graphs and  competitive performance on homophilic graphs against state-of-the-art GNNs.
Moreover, experimental results on graphs with different levels of noisy edges show that \pgnns work much more robustly than GNN baselines and even as well as MLPs on graphs with completely random edges.
Additional experiments (reported in \cref{App:plug_exp}) illustrate that intergrating \pgnns with existing GNN architectures (i.e. GCN~\citep{DBLP:conf/iclr/KipfW17}, JKNet~\citep{DBLP:conf/icml/XuLTSKJ18}) can significantly improve their performance on heterophilic graphs. In conclusion,  our contributions can be summarized as below:

(1) \textbf{New methodologies.} We propose $p$-Laplacian message passing and \pgnn to adapt GNNs to heterophilic graphs and graphs where the topology  is non-informative for label prediction. (2) \textbf{Superior performance.}
We empirically demonstrate that \pgnns is superior on heterophilic graphs and competitive on homophilic graphs against state-of-the-art GNNs.
Moreover, \pgnns work  robustly on graphs with noisy edges or non-informative topologies. (3) \textbf{Theoretical justification.} We theoretically demonstrate that $p$-Laplacian message passing works as low-high-pass filters and 
the message passing iteration is guaranteed to converge with proper settings.
(4) \textbf{New paradigm of  designing GNN architectures.} We bridge the gap between discrete regularization framework and GNNs, which could further inspire researchers to develop new graph convolutions or message passing schemes using other regularization techniques with explicit assumptions on graphs. 
Due to  space limit, we defer the discussions on related work and future work and all proofs to the Appendix. Code available at \url{https://github.com/guoji-fu/pGNNs}.

%% file: sections/2-Pre.tex
\section{Preliminaries and Background}\label{sec:sec2}

\textbf{Notation.} Let $\gG = (\gV, \gE, \mW)$ be an undirected graph, where $\gV = \{1, 2, \dots, N\}$ is the set of nodes, $\gE \subseteq \gV \times \gV$ is the set of edges, $\mW \in \R^{N \times N}$ is the adjacency matrix and $\emW_{i,j} = \emW_{j,i}$, $\emW_{i,j} > 0$ for $[i, j] \in \gE$, $\emW_{i, j} = 0$, otherwise. $\gN_i = \{j\}_{[i, j] \in \gE}$ denotes the set of neighbors of node $i$, $\mD \in \R^{N \times N} = \mathrm{diag}(\emD_{1,1}, \dots \emD_{N,N})$ denotes the diagonal degree matrix with $\emD_{i,i} = \sum_{j=1}^N\emW_{i,j}$, for $i = 1, \dots, N$. $f: \gV \rightarrow \R$ and $g: \gE \rightarrow \R$ are functions defined on the vertices  and edges of $\gG$, respectively. $\gF_\gV$ denotes the Hilbert space of functions endowed with the inner product $\langle f, \tilde{f}\rangle_{\gF_\gV} := \sum_{i \in \gV}f(i)\tilde{f}(i)$. Similarly define $\gF_\gE$. We also denote by $[K] = \{1, 2, \dots, K\}, \forall K \in \sN$ and we use $\|\vx\| = \|\vx\|_2 = (\sum_{i=1}^d\evx_i^2)^{1/2}, \forall \vx \in \R^d$ to denote the Frobenius norm of a vector.

\textbf{Problem Formulation.} Given a graph $\gG = (\gV, \gE, \mW)$, each node $i \in \gV$ has a feature vector $\mX_{i,:}$ which is the $i$-th row of $\mX$ and a subset of nodes in $\gG$ have labels from a label set $\gL = \{1, \dots, L\}$. The goal of GNNs for semi-supervised node classification on $\gG$ is to train a GNN $\gM: \gV \rightarrow \gL$ and predict the labels of unlabeled nodes.

\textbf{Homophily and Heterophily.}
The homophily or heterophily of a graph is used to describe the relation of labels between linked nodes in the graphs. The level of homophily of a graph can be measured by $\gH(\gG) = \E_{i \in \gV}\left[\left|\{j\}_{j \in \gN_i, y_i = y_j}\right| \left/ \left|\gN_i\right|\right.\right]$~\citep{DBLP:conf/iclr/PeiWCLY20,DBLP:conf/iclr/ChienP0M21}, where $\left|\{j\}_{j \in \gN_i, y_i = y_j}\right|$ denotes the number of neighbors of $i \in V$ that share  the same label as $i$ (i.e. $y_i = y_j$) and $\gH(\gG) \rightarrow 1$ corresponds to strong homophily while $\gH(\gG) \rightarrow 0$ indicates strong heterophily. We say that a graph is a homophilic (heterophilic) graph if it has strong homophily (heterophily).

\textbf{Graph Gradient.} The graph gradient of an edge $[i, j], i, j \in \gV$ is defined to be a measurement of the variation of a function $f\footnote{$f$ can be a vector function: $f: \gV \rightarrow \R^c$ for some $c \in \sN$ and here we use $f: \gV \rightarrow \R$ for better illustration.}: \gV \rightarrow \R$ on the edge $[i, j]$.
\begin{definition}[Graph Gradient]\label{def:def1}
	Given a graph $\gG = (\gV, \gE)$ and a function $f: \gV \rightarrow \R$, the graph gradient is an operator $\nabla: \gF_\gV \rightarrow \gF_\gE$ defined as for all $[i, j] \in \gE$, 
	\begin{equation}\label{eq:eq1}
		(\nabla f)([i,j]) := \sqrt{\frac{\emW_{i,j}}{\emD_{j,j}}}f(j) - \sqrt{\frac{\emW_{i,j}}{\emD_{i,i}}}f(i).
	\end{equation}
\end{definition}

For $[i, j] \notin \gE$, $(\nabla f)([i,j]) := 0$. The graph gradient of a function $f$ at a vertex $i, i \in [N]$ is defined to be $\nabla f(i) := ((\nabla f)([i, 1]), \dots, (\nabla f)([i, N]))$ and its Frobenius norm is given by $\|\nabla f(i)\|_2 := (\sum_{j=1}^N(\nabla f)^2([i,j]))^{1/2}$, which measures the variation of $f$ around node $i$. We measure the variation of $f$ over the whole graph $\gG$ by $\mathcal{S}_p(f)$ where it is defined as for $p \geq 1$,
\begin{align}
	\mathcal{S}_p(f) := {} & \frac{1}{2}\sum_{i=1}^N\sum_{j=1}^N\left\|(\nabla f)([i, j])\right\|^p \notag \\
	= {} & \frac{1}{2}\sum_{i=1}^N\sum_{j=1}^N\left\|\sqrt{\frac{\emW_{i,j}}{\emD_{j,j}}}f(j) - \sqrt{\frac{\emW_{i,j}}{\emD_{i,i}}}f(i)\right\|^p. \label{eq:eq2}
\end{align}
Note that the definition of $\gS_p$ here is different with the $p$-Dirichlet form in \citet{DBLP:conf/dagm/ZhouS05}.

\textbf{Graph Divergence.} The graph divergence is defined to be the adjoint of the graph gradient:
\begin{definition}[Graph Divergence]\label{def:def2}
	Given a graph $\gG = (\gV, \gE)$, and functions $f:\gV \rightarrow \R$, $g: \gE \rightarrow \R$, the graph divergence is an operator $\mathrm{div}: \gF_\gE \rightarrow \gF_\gV$ which satisfies
	\begin{equation}\label{eq:eq3}
		\langle \nabla f, g \rangle = \langle f, -\mathrm{div}g \rangle.
	\end{equation}
    The graph divergence can be computed by
    \begin{equation}\label{eq:eq4}
    	(\mathrm{div}g)(i) = \sum_{j=1}^N\sqrt{\frac{\emW_{i,j}}{\emD_{i,i}}}\left(g([i,j]) - g([j,i])\right).
    \end{equation}
\end{definition}
\vspace{-5pt}
\cref{App:fig:example} in \cref{App:subsec:example} gives a tiny example of illustration of graph gradient and graph divergence.

\textbf{Graph $p$-Laplacian Operator.} By the definitions of graph gradient and graph divergence, we reach the definition of graph $p$-Laplacian operator as below.
\begin{definition}[Graph $p$-Laplacian\footnote{Note that the definition  adopted is slightly different with the one used in \citet{DBLP:conf/dagm/ZhouS05} where $\|\cdot\|^{p-2}$ is not element-wise and the one used in some literature such as \citet{DBLP:journals/arscom/Amghibech03,DBLP:conf/icml/BuhlerH09}, where  $(\Delta_pf)(i) = \sum_{j=1}^N\frac{\emW_{i,j}}{\emD_{i,i}}\left|f(i) - f(j)\right|^{p-2}\left(f(i) - f(j)\right)$ for $p > 1$ and $p=1$ is not allowed.}]\label{def:def3}
	Given a graph $\gG = (\gV, \gE)$ and a function $f: \gV \rightarrow \R$, the graph $p$-Laplacian is an operator $\Delta_p: \gF_\gV \rightarrow \gF_\gV$ defined by
	\begin{equation}\label{eq:eq5}
		\Delta_pf := -\frac{1}{2}\mathrm{div}(\|\nabla f\|^{p-2}\nabla f), \text{ for } p \geq 1.
	\end{equation}
	where $\|\cdot\|^{p-2}$ is element-wise, i.e. $\|\nabla f(i)\|^{p-2} = (\|(\nabla f)([i, 1])\|^{p-2}, \dots, \|(\nabla f)([i, N])\|^{p-2})$.
\end{definition}
\vspace{-5pt}
Substituting \cref{eq:eq1} and \cref{eq:eq4} into \cref{eq:eq5}, we obtain
\begin{align}
	(\Delta_pf)(i) = {} & \sum_{j=1}^N\sqrt{\frac{\emW_{i,j}}{\emD_{i,i}}}\|(\nabla f)([j, i])\|^{p-2}\left(\sqrt{\frac{\emW_{i,j}}{\emD_{i,i}}}f(i)\right. \notag \\
	{} & \qquad \left.- \sqrt{\frac{\emW_{i,j}}{\emD_{j,j}}}f(j)\right). \label{eq:eq6}
\end{align}
The graph $p$-Laplacian is semi-definite: $\langle f, \Delta_pf \rangle = \mathcal{S}_p(f) \geq 0$ and we have
\begin{equation}\label{eq:eq7}
	\left.\frac{\partial \mathcal{S}_p(f)}{\partial f}\right|_i = p (\Delta_pf) (i).
\end{equation}
When $p=2$, $\Delta_2$ is refered as the ordinary Laplacian operator and $\Delta_2 = \mI - \mD^{-1/2}\mW\mD^{-1/2}$ and when $p=1$, $\Delta_1$ is refered as the  Curvature operator and $\Delta_1f := -\frac{1}{2}\mathrm{div}(\|\nabla f\|^{-1}\nabla f)$. Note that Laplacian $\Delta_2$ is a linear operator, while in general for $p \neq 2$, $p$-Laplacian is nonlinear since $\Delta_p(af) \neq a\Delta_p(f)$ for $a \in \R$.

%% file: sections/3-pGNN.tex
\section{$p$-Laplacian based Graph Neural Networks}\label{sec:sec3}
\vspace{-5pt}
In this section, we derive the $p$-Laplacian message passing scheme from a $p$-Laplacian regularization framework and present \pgnn, a new GNN architecture developed upon the new message passing scheme. 
We demonstrate that $p$-Laplacian message passing scheme is guaranteed to converge with proper settings and provide an upper-bounding risk of $^p$GNNs in \cref{App:the:the3}.
\vspace{-5pt}
\subsection{$p$-Laplacian Regularization Framework}\label{subsec:plap_reg}
\vspace{-5pt}
Given an undirected graph $\gG = (\gV, \gE)$ and a signal function with $c$ ($c \in \sN$) channels 
$f: \gV \rightarrow \R^c$,
let $\mX = (\mX_{1,:}^\top, \dots, \mX_{N,:}^\top)^\top \in \R^{N \times c}$ with $\mX_{i,:} \in \R^{1 \times c}, i \in [N]$ denoting the node features of $\gG$ and $\mF = (\mF_{1,:}^\top, \dots, \mF_{N,:}^\top)^\top \in \R^{N \times c}$ be a matrix whose $i^{th}$ row vector $\mF_{i,:} \in \R^{1 \times c}, i \in [N]$ represents the function value of $f$ at the $i$-th vertex in $\gG$. We present a $p$-Laplacian regularization problem whose cost function is defined to be
\begin{equation}\label{eq:eq8}
	\gL_p(\mF) := \min_{\mF} \; \mathcal{S}_p(\mF) + \mu\sum_{i=1}^N\|\mF_{i,:} - \mX_{i,:}\|^2, 
\end{equation}
where $\mu \in (0, \infty)$. The first term of the right-hand side in \cref{eq:eq8} is a measurement of variation of the signal over the graph based on $p$-Laplacian. As we will discuss later, different choices of $p$ result in different smoothness constraint on the signals. The second term is the constraint that the optimal signals $\mF^* = \argmin\gL_p(\mF)$ should not change too much from the input signal $\mX$, and $\mu$ provides a trade-off between these two constraints.  

\textbf{Regularization with $p = 2$.} If $p = 2$, the solution of \cref{eq:eq8} satisfies $\Delta_2\mF^* + \mu(\mF^* - \mX) = \mathbf{0}$ and we can obtain the closed form~\citep{DBLP:conf/dagm/ZhouS05}
\begin{equation}\label{eq:eq9}
	\mF^* = \mu(\Delta_2 + \mu\mI_N)^{-1}\mX. 
\end{equation}
Then, we could use the following iteration algorithm to get an approximation of \cref{eq:eq9}:
\begin{equation}\label{eq:eq10}
	\mF^{(k+1)} = \alpha\mD^{-1/2}\mW\mD^{-1/2}\mF^{(k)} + \beta \mX,
\end{equation}
where $k$ represents the iteration index, $\alpha = \frac{1}{1+\mu}$ and $\beta = \frac{\mu}{1+\mu} = 1 - \alpha$. The iteration converges to a closed-form solution as $k$ goes to infinity~\citep{DBLP:conf/nips/ZhouBLWS03,DBLP:conf/dagm/ZhouS05}.
We could relate the  the result here with the personalized PageRank (PPR)~\citep{page1999pagerank,DBLP:conf/iclr/KlicperaBG19} algorithm: 
\begin{theorem}[Relation to personalized PageRank~\citep{DBLP:conf/iclr/KlicperaBG19}]\label{the:the1}
	$\mu(\Delta_2 + \mu\mI_N)^{-1}$ in the closed-form solution of \cref{eq:eq9} is equivalent to the personalized PageRank matrix.
\end{theorem}
\textbf{Regularization with $p > 1$.} For $p > 1$, the solution of \cref{eq:eq8} satisfies $p\Delta_p\mF^* + 2\mu(\mF^* - \mX) = \mathbf{0}$. By \cref{eq:eq6} we have that, for all $i \in [N]$,
\begin{align*}
	{} & \sum_{j=1}^N\frac{\emW_{i,j}}{\sqrt{\emD_{i,i}}}\|(\nabla f^*)([j, i])\|^{p-2}\left(\frac{1}{\sqrt{\emD_{i,i}}}\mF_{i,:}^* - \frac{1}{\sqrt{\emD_{j,j}}}\mF_{j,:}^*\right) \\ 
	{} & \qquad \qquad \qquad \qquad + \frac{2\mu}{p}\left(\mF_{i,:}^* - \mX_{i,:}\right) = \mathbf{0}.
\end{align*}
Let $\mM^{(k)} \in \R^{N \times N}$, $\bm{\alpha}^{(k)} = \mathrm{diag}(\alpha_{1,1}^{(k)}, \dots, \alpha_{N,N}^{(k)})$, $\bm{\beta}^{(k)} = \mathrm{diag}(\beta_{1,1}^{(k)}, \dots, \beta_{N,N}^{(k)})$.
Based on which we can construct a similar iterative algorithm to obtain a solution~\citep{DBLP:conf/dagm/ZhouS05}: for all $i \in [N]$,
\begin{equation}\label{eq:eq11}
	\mF_{i,:}^{(k+1)} = \alpha_{i,i}^{(k)}\sum_{j=1}^N\frac{\emM_{i,j}^{(k)}}{\sqrt{\emD_{i,i}\emD_{j,j}}}\mF_{j,:}^{(k)} + \beta_{i,i}^{(k)}\mX_{i,:},
\end{equation}
for all $i, j \in [N]$,
\begin{equation}\label{eq:eq12}
	\emM_{i,j}^{(k)} = \emW_{i, j}\left\|\sqrt{\frac{\emW_{i,j}}{\emD_{i,i}}}\mF_{i,:}^{(k)} - \sqrt{\frac{\emW_{i,j}}{\emD_{j,j}}}\mF_{j,:}^{(k)}\right\|^{p-2},
\end{equation}
\begin{equation}\label{eq:eq13}
	\alpha_{i,i}^{(k)} = 1 \big/ \left(\sum_{j=1}^N\frac{\emM_{i,j}^{(k)}}{\emD_{i,i}} + \frac{2\mu}{p}\right), \quad \beta_{i,i}^{(k)} = \frac{2\mu}{p}\alpha_{i,i}^{(k)}.
\end{equation}
Note that when $\left\|\sqrt{\frac{\emW_{i,j}}{\emD_{i,i}}}\mF_{i,:}^{(k)} - \sqrt{\frac{\emW_{i,j}}{\emD_{j,j}}}\mF_{j,:}^{(k)}\right\| = 0$, we set $M_{i,j}^{(k)} = 0$. It is easy to see that \cref{eq:eq10} is the special cases of \cref{eq:eq15} with $p=2$. 
\begin{remark}[Discussion on $p = 1$]
    For $p = 1$, when $f$ is a real-valued function ($c = 1$), $\Delta_1 f$ is a step function, which could make the stationary condition of the objective function \cref{eq:eq8} become problematic. Additionally, $\Delta_1f$ is not continuous at $\|(\nabla f)([i, j])\| = 0$. Therefore, $p = 1$ is not allowed when $f$ is a real value function. On the other hand, note that there is a Frobenius norm in $\Delta_pf$. When $f$ is a vector-valued function ($c > 1$), the step function in $\Delta_1f$ only exists on the axes. The stationary condition will be fine if the node embeddings $\mF$ are not a matrix of vectors that has only one non-zero element, which is true for many graphs. $p = 1$ may work for these graphs. Overall, we suggest to use $p > 1$ in practice but $p = 1$ may work for graphs with multiple channel signals as well. We conduct experiments for $p > 1$ ($p = 1.5, 2, 2.5$) and $p = 1$ in \cref{sec:sec5}. 
\end{remark}

\subsection{$p$-Laplacian Message Passing and \pgnn Architecture}\label{subsec:pgnn}
\begin{snugshade}
\textbf{$p$-Laplacian Message Passing.} Rewrite \cref{eq:eq11} in a matrix form we obtain
\begin{equation}\label{eq:eq15}
	\mF^{(k+1)} = \bm{\alpha}^{(k)}\mD^{-1/2}\mM^{(k)}\mD^{-1/2}\mF^{(k)} + \bm{\beta}^{(k)}\mX.
\end{equation}
\end{snugshade}
\vspace{-5pt}
\cref{eq:eq15} provides a new message passing mechanism, named $p$-Laplacian message passing.
\begin{remark}
    $\bm{\alpha}\mD^{-1/2}\mM\mD^{-1/2}$ in \cref{eq:eq15} can be regarded as the learned aggregation weights at each step for message passing. It suggests that $p$-Laplacian message passing could adaptively tune the aggregation weights during the course of learning, which will be demonstrated theoretically and empirically in the later. $\beta\mX$ in \cref{eq:eq15} can be regarded as a residual unit, which  helps the model escape from the oversmoothing issue~\citep{DBLP:conf/iclr/ChienP0M21}.
\end{remark}

We present the following theorem to show the shrinking property of $p$-Laplacian message passing.
\begin{snugshade}
\begin{theorem}[Shrinking Property of $p$-Laplacian Message Passing]\label{the:the2}
	Given a graph $\gG = (\gV, \gE, \mW)$ with node features $\mX$,   $\bm{\beta}^{(k)},\mF^{(k)}, \mM^{(k)}, \bm{\alpha}^{(k)}$ are updated accordingly to \cref{eq:eq11,eq:eq12,eq:eq13} for $k = 0, 1, \dots, K$ and $\mF^{(0)} = \mX$.  Then there exist some positive real value $\mu > 0$ which depends on $\mX, \gG, p$ and $p > 1$ such that
	\begin{equation*}
		\gL_p(\mF^{(k+1)}) \leq \gL_p(\mF^{(k)}).
	\end{equation*}
\end{theorem}
\end{snugshade}
\vspace{-5pt}
Proof see \cref{App:the2}. \cref{the:the2} shows that with some proper positive real value $\mu$ and $p > 1$, the loss of the objective function~\cref{eq:eq8} is guaranteed to decline after taking one step $p$-Laplacian message passing. \cref{the:the2} also demonstrates that the iteration \cref{eq:eq11,eq:eq12,eq:eq13} is guaranteed to converge for $p > 1$ with some proper $\mu$ which is chosen depends on the input graph and $p$.
\begin{snugshade}
\textbf{\pgnn Architecture.} We design the architecture of \pgnns using $p$-Laplacian message passing. Given node features $\mX \in \R^{N \times c}$, the number of node labels $L$, the number of hidden units $h$, the maximum number of iterations $K$, and $\mM$, $\bm{\alpha}$, and $\bm{\beta}$ updated by \cref{eq:eq12,eq:eq13} respectively, we give the \pgnn architecture as following:
\begin{align}
	& \mF^{(0)}  = \text{ReLU}(\mX\Theta^{(1)}), \label{eq:eq16} \\
	& \mF^{(k+1)}  =  \bm{\alpha}^{(k)}\mD^{-1/2}\mM^{(k)}\mD^{-1/2}\mF^{(k)} + \bm{\beta}^{(k)}\mF^{(0)}, \label{eq:eq17} \\
	& \mZ  =  \text{softmax}(\mF^{(K)}\Theta^{(2)}), \label{eq:eq18}
\end{align}
\end{snugshade}
\vspace{-10pt}
where $k =0, 1, \dots, K-1$, $\mZ \in \R^{N \times L}$ is the output propbability matrix with $\emZ_{i,j}$ is the estimated probability that the label at node $i \in [N]$ is $j \in [L]$ given features $\mX$ and graph $\gG$, $\Theta^{(1)} \in \R^{c \times h}$ and $\Theta^{(2)} \in \R^{h \times L}$ are the first- and the second-layer parameters of the neural network.

\begin{remark}[Connection to existing GNN variants]\label{remk:remk2}
	The message passing scheme of $^p$GNNs is different from that of several GNN variants (say, GCN, GAT, and GraphSage), which repeatedly stack message passing layers.    
	In contrast, it is similar with  SGC~\citep{DBLP:conf/icml/WuSZFYW19}, APPNP~\citep{DBLP:conf/iclr/KlicperaBG19}, and GPRGNN~\citep{DBLP:conf/iclr/ChienP0M21}. SGC is an approximation to the closed-form in \cref{eq:eq9}~\citep{DBLP:journals/corr/abs-2006-04386}. By \cref{the:the1}, it is easy to see that APPNP, which uses PPR to propagate the node embeddings, is analogical to \pgnn with $p = 2$, termed as $^{2.0}$GNN. APPNP and $^{2.0}$GNN work analogically and effectively on homophilic graphs. $^{2.0}$GNN can also work effectively on heterophilic graphs by letting $\Theta^{(2)}$ be negative. However, APPNP fails on heterophilic graphs as its PPR weights are fixed~\citep{DBLP:conf/iclr/ChienP0M21}. Unlike APPNP, GPRGNN, which adaptively learn the generalized PageRank (GPR) weights, works similarly to $^{2.0}$GNN on both homophilic and heterophilic graphs. However, GPRGNN needs more supervised information in order to learn optimal GPR weights. On the contrary, \pgnns need less supervised information to obtain similar results because $\Theta^{(2)}$ acts like a hyperplane for classification. \pgnns could work better under weak supervised information. Our analysis is also verified by the experimental results in \cref{sec:sec5}.
\end{remark}

We also provide an upper-bounding risk of \pgnns by \cref{the:the3} in \cref{App:the:the3} to study the effect of the hyperparameter $\mu$ on the performance of \pgnns. \cref{the:the3} shows that the risk of $^p$GNNs is upper-bounded by the sum of three terms: the risk of label prediction using only the original node features $\mX$, the norm of $p$-Laplacian diffusion on $\mX$, and the magnitude of the noise in $\mX$. $\mu$ controls the trade-off between these three terms. The smaller $\mu$, the more weights on the $p$-Laplacian diffusion term and the noise term and the less weights on the the other term and vice versa.

%% file: sections/4-Analysis.tex
\section{Spectral Views of $p$-Laplacian Message Passing}\label{sec:sec4}

In this section, we theoretically demonstrate that $p$-Laplacian message passing is an approximation of a polynomial graph filter defined on the spectral domain of $p$-Laplacian. We show by spectral analysis that $p$-Laplacian message passing works as low-high-pass filters.

\textbf{$p$-Eigenvalues and $p$-Eigenvectors of the Graph $p$-Laplacian.} We first introduce the definitions of $p$-eigenvalues and $p$-eigenvectors of $p$-Laplacian. Let $\phi_p: \R \rightarrow \R$ defined as $\phi_p(u) = \|u\|^{p-2}u, \text{ for } u \in \R, u \neq 0$.	Note that $\phi_2(u) = u$. For notational simplicity, we denote by $\phi_p(\vu) = (\phi_p(\evu_1), \dots, \phi_p(\evu_N))^\top$ for $\vu \in \R^N$ and $\Phi_p(\mU) = (\phi_p(\mU_{:,1}), \dots, \phi_p(\mU_{:,N}))$ for $\mU \in \R^{N \times N}$ and $\mU_{:,i} \in \R^N$ is the $i^{th}$ column vector of $\mU$.

\begin{definition}[$p$-Eigenvector and $p$-Eigenvalue]\label{def:def5}
	A vector $\vu \in \R^N$ is a $p$-eigenvector of $\Delta_p$ if it satisfies the equation
	\begin{equation*}
		\left(\Delta_p\vu\right)_i = \lambda\phi_p(\evu_i), \quad \text{for all } i \in [N],
	\end{equation*}
	where $\lambda \in \R$ is a real value referred as a $p$-eigenvalue of $\Delta_p$ associated with the $p$-eigenvector $\vu$.
\end{definition}

\begin{definition}[$p$-Orthogonal~\citep*{DBLP:journals/ml/LuoHDN10}]\label{def:def6}
	Given two vectors $\vu, \vv \in \R^N$ with $\vu, \vv \neq \bm{0}$, we call that $\vu$ and $\vv$ is $p$-orthogonal if 
	\begin{equation*}
		\phi_p(\vu)^\top\phi_p(\vv) = \sum_{i=1}^N\phi_p(\evu_i)\phi_p(v_i) = 0.
	\end{equation*}
\end{definition}


\citet{DBLP:journals/ml/LuoHDN10} demonstrated that the $p$-eigenvectors of $\Delta_p$ are $p$-orthogonal to each other (see \cref{the:the4} in \cref{App:the:the4} for details). Therefore, the space spanned by the multiple $p$-eigenvectors of $\Delta_p$ is $p$-orthogonal. Additionally, we demonstrate that the $p$-eigen-decomposition of $\Delta_p$ is given by: $\Delta_p = \Phi_p(\mU) \mLambda \Phi_p(\mU)^\top$ (see \cref{the:the5} in \cref{App:the:the5} for details), where $\mU$ is a matrix of $p$-eigenvectors of $\Delta_p$ and $\mLambda$ is a diagonal matrix in which the diagonal is the $p$-eigenvalues of $\Delta_p$. 
\textbf{Graph Convolutions based on $p$-Laplacian.} Based on \cref{the:the4}, the graph Fourier Transform $\hat{f}$ of any function $f$ on the vertices of $\gG$ can be defined as the expansion of $f$ in terms of $\Phi(\mU)$ where $\mU$ is the matrix of $p$-eigenvectors of $\Delta_p$: $\hat{f} = \Phi_p(\mU)^{\top}f$. Similarly, the inverse graph Fourier transform is then given by: $f = \Phi_p(\mU)\hat{f}$. Therefore, a signal $\mX \in \R^{N \times c}$ being filtered by a spectral filter $g_{\theta}$ can be expressed formally as: $g_{\bm{\theta}} \star \mX = \Phi_p(\mU)\hat{g}_{\bm{\theta}}(\mLambda)\Phi_p(\mU)^\top\mX$,
where $\mLambda$ denotes a diagonal matrix in which the diagonal corresponds to the $p$-eigenvalues $\{\lambda_l\}_{l = 0, \dots, N-1}$ of $\Delta_p$ and $\hat{g}_{\bm{\theta}}(\mLambda)$ denotes a diagonal matrix in which the diagonal corresponds to spectral filter coefficients. Let $\hat{g}_{\bm{\theta}}$ be a polynomial filter defined as $\hat{g}_{\bm{\theta}} = \sum_{k=0}^{K-1}\theta_k\lambda_l^k$, where the parameter $\bm{\theta} = [\theta_0, \dots, \theta_{K-1}]^\top \in \R^K$ is a vector of polynomial coefficients. By the $p$-eigen-decomposition of $p$-Laplacian, we have
\begin{footnotesize}
\begin{equation}\label{eq:eq22}
	g_{\bm{\theta}} \star \mX \approx \sum_{k=0}^{K-1}\theta_k\Phi_p(\mU)\mLambda^k\Phi_p(\mU)^\top\mX = \sum_{k=0}^{K-1}\theta_k\Delta_p^k\mX 
\end{equation}
\end{footnotesize}
\begin{snugshade}
\begin{theorem}\label{the:the6}
	The $K$-step $p$-Laplacian message passing is a $K$-order polynomial approximation to the graph filter given by \cref{eq:eq22}.
\end{theorem}
\end{snugshade}
Proof see \cref{App:the6}. \cref{the:the6} indicates that $p$-Laplacian message passing is implicitly a polynomial spectral filter defined on the spectral domain of $p$-Laplacian.

\textbf{Spectral Analysis of $p$-Laplacian Message Passing.} Here, we analyze the spectral propecties of $p$-Laplacian message passing. We can approximately view $p$-Laplacian message pasing as a filter of a linear combination of $K$ spectral filters $g(\mLambda)^{(0)}, g(\mLambda)^{(1)}, \dots, g(\mLambda)^{(K-1)}$ with each spectral filter defined to be $g(\mLambda)^{(k)} : = (\bm{\alpha}\mD^{-1/2}\mM\mD^{-1/2})^k$ where $\emM_{i,j} = \emW_{i, j}\|\sqrt{\frac{\emW_{i,j}}{\emD_{i,i}}}\mF_{i,:} - \sqrt{\frac{\emW_{i,j}}{\emD_{j,j}}}\mF_{j,:}\|^{p-2}$ for $i, j \in [N]$ and $\mF$ is the matrix of node embeddings. We can study the properties of $p$-Laplacian message passing by studying the spectral properties of $\bm{\alpha}\mD^{-1/2}\mM\mD^{-1/2}$ as given below.
\begin{snugshade}
\begin{proposition}\label{the:prop1}
	Given a connected graph $\gG = (\gV, \gE, \mW)$ with node embeddings $\mF$ and the $p$-Laplacian $\Delta_p$ with its $p$-eigenvectors $\{\vu^{(l)}\}_{l=0, 1, \dots, N-1}$ and the $p$-eigenvalues $\{\lambda_l\}_{l=0, 1, \dots, N-1}$. Let $g_p(\lambda_{i-1}) := \alpha_{i,i}\sum_j\emD_{i,i}^{-1/2}\emM_{i,j}\emD_{j,j}^{-1/2}$ for $i \in [N]$ be the filters defined on the spectral domain of $\Delta_p$, where $\emM_{i,j} = \emW_{i,j}\|\nabla f([i,j])\|^{p-2}$, $(\nabla f)([i,j])$ is the graph gradient of the edge between node $i$ and $j$ and $\|\nabla f(i)\|$ is the norm of graph gradient at $i$. $N_i$ denotes the number of edges connected to $i$, $N_{\text{min}} = \min\{N_j\}_{j \in [N]}$, and $k = \argmax_j(\{|\evu_j^{(l)}| / \sqrt{\emD_{l,l}}\}_{j \in [N]; l=0, \dots, N-1})$, then 
	\begin{enumerate}
		\item When $p = 2$, $g_p(\lambda_{i-1})$ works as low-high-pass filters.
		
		\item When $p > 2$, if $\|\nabla f(i)\| \leq 2^{(p-1)/(p-2)}$, $g_p(\lambda_{i-1})$ works as low-high-pass filters on node $i$ and $g_p(\lambda_{i-1})$ works as low-pass filters on $i$ when $\|\nabla f(i)\| \geq 2^{(p-1)/(p-2)}$.
		
		\item When $1 \leq p < 2$, if $0 \leq \|\nabla f(i)\| \leq 2(2\sqrt{N_k})^{1/(p-2)}$, $g_p(\lambda_{i-1})$ works as low-pass filters on node $i$ and $g_p(\lambda_{i-1})$ works as low-high-pass filters on $i$ when $\|\nabla f(i)\| \geq 2\left(2\sqrt{N_k}\right)^{1/(p-2)}$. Specifically, when $p = 1$, $N_k$ can be replaced by $N_{\text{min}}$.
	\end{enumerate}
\end{proposition}
\end{snugshade}
Proof see \cref{App:prop1}. \cref{the:prop1} shows that when $p \neq 2$, $p$-Laplacian message passing adaptively works as low-pass or low-high-pass filters on node $i$ in terms of the degree of local node embedding variation around $i$, i.e. the norm of the graph gradient $\|\nabla f(i)\|$ at node $i$. When $p = 2$, $p$-Laplacian message passing works as low-high-pass filters on node $i$ regardless of the value of $\|\nabla f(i)\|$. When $p > 2$, $p$-Laplacian message passing works as low-pass filters on node $i$ for large $\|\nabla f(i)\|$ and works as low-high-pass filters for small $\|\nabla f(i)\|$. Therefore, \pgnns with $p > 2$ can work very effectively on graphs with strong homophily.
When $1 \leq p < 2$, $p$-Laplacian message passing works as low-pass filters for small $\|\nabla f(i)\|$ and works as low-high-pass filters for large $\|\nabla f(i)\|$. Thus, \pgnns with $1\leq p < 2$ can work effectively on graphs with low homophily, i.e. heterophilic graphs. 

\begin{remark}
    We say that a filter works as a low-pass filter if the low frequencies dominate the other frequencies and it works as a low-high-pass filter if the low frequencies and high frequencies dominate the middle frequencies, i.e., the frequencies around the cutoff frequency.
\end{remark}

\begin{remark}
    Previous works~\citep{DBLP:conf/icml/WuSZFYW19,DBLP:conf/iclr/KlicperaBG19} have shown that GCN, SGC, APPNP work as low-pass filters. The reason is that they have added the self-loop to the adjacency matrix, which will shrink the spectral range of the Laplacian from $[0, 2]$ to approximately $[0, 1.5]$ and causes them work as low-pass filters~\citep{DBLP:conf/icml/WuSZFYW19}. On the contrary, we did not add self-loop to the adjacancy matrix and therefore the spectral filter work as low-high-pass filters in our case for $p=2$.
\end{remark}

%% file: sections/5-Exp.tex
\section{Empirical Studies}\label{sec:sec5}
Here, we empirically study the effectiveness of \pgnns for semi-supervised node classification using and real-world benchmark and synthetic datasets with heterophily and strong homophily. The experimental results are also used to validate our theoretical findings presented previously.

%
%

\begin{table*}[htp]
	\centering
	\caption{Heterophilious results. Averaged accuracy (\%) for 100 runs. Best results outlined in bold and the results within $95\%$ confidence interval of the best results are outlined in underlined bold.}\label{tab:tab_heter_exp}
    \resizebox{\textwidth}{!}{
	\begin{tabular}{ccccccc}
		\toprule
		Method & Chameleon & Squirrel & Actor & Wisconsin & Texas & Cornell \\
		\midrule
		MLP & $48.02_{\pm 1.72}$ & $\bm{33.80}_{\pm 1.05}$ & $39.68_{\pm 1.43}$ & $93.56_{\pm 3.14}$ & $79.50_{\pm 10.62}$ & $80.30_{\pm 11.38}$ \\
		GCN & $34.54_{\pm 2.78}$ & $25.28_{\pm 1.55}$ & $31.28_{\pm 2.04}$ & $61.93_{\pm 3.00}$ & $56.54_{\pm 17.02}$ & $51.36_{\pm 4.59}$ \\
		SGC & $34.76_{\pm 4.55}$ & $25.49_{\pm 1.63}$ & $30.98_{\pm 3.80}$ & $66.94_{\pm 2.58}$ & $59.99_{\pm 9.95}$ & $44.39_{\pm 5.88}$ \\
		GAT & $45.16_{\pm 2.10}$ & $31.41_{\pm 0.98}$ & $34.11_{\pm 1.28}$ & $65.64_{\pm 6.29}$ & $56.41_{\pm 13.01}$ & $43.94_{\pm 7.33}$ \\
		JKNet & $33.28_{\pm 3.59}$ & $25.82_{\pm 1.58}$ & $29.77_{\pm 2.61}$ & $61.08_{\pm 3.71}$ & $59.65_{\pm 12.62}$ & $55.34_{\pm 4.43}$ \\
		APPNP & $36.18_{\pm 2.81}$ & $26.85_{\pm 1.48}$ & $31.26_{\pm 2.52}$ & $64.59_{\pm 3.49}$ & $82.90_{\pm 5.08}$ & $66.47_{\pm 9.34}$ \\
		GPRGNN & $43.67_{\pm 2.27}$ & $31.27_{\pm 1.76}$ & $36.63_{\pm 1.22}$ & $88.54_{\pm 4.94}$ & $80.74_{\pm 6.76}$ & $78.95_{\pm 8.52}$ \\
		\midrule
		$^{1.0}$GNN & $\bm{48.86}_{\pm 1.95}$ & $\bm{\underline{33.75}}_{\pm 1.50}$ & $\bm{40.62}_{\pm 1.25}$ & $\bm{95.37}_{\pm 2.06}$ & $84.06_{\pm 7.41}$ & $\bm{82.16}_{\pm 8.62}$ \\
		$^{1.5}$GNN & $\bm{\underline{48.74}}_{\pm 1.62}$ & $33.33_{\pm 1.45}$ & $\bm{\underline{40.35}}_{\pm 1.35}$ & $\bm{\underline{95.24}}_{\pm 2.01}$ & $84.46_{\pm 7.79}$ & $78.47_{\pm 6.87}$ \\
		$^{2.0}$GNN & $\bm{\underline{48.77}}_{\pm 1.87}$ & $33.60_{\pm 1.47}$ & $40.07_{\pm 1.17}$ & $91.15_{\pm 2.76}$ & $\bm{87.96}_{\pm 6.27}$ & $72.04_{\pm 8.22}$ \\
		$^{2.5}$GNN & $\bm{\underline{48.80}}_{\pm 1.77}$ & $\bm{\underline{33.79}}_{\pm 1.45}$ & $39.80_{\pm 1.31}$ & $87.08_{\pm 2.69}$ & $83.01_{\pm 6.80}$ & $70.31_{\pm 8.84}$ \\
		\bottomrule
	\end{tabular}}
\end{table*}

\textbf{Datasets and Experimental Setup.} We use seven homophilic benchmark datasets: citation graphs Cora, CiteSeer, PubMed~\citep{DBLP:journals/aim/SenNBGGE08}, Amazon co-purchase graphs Computers, Photo, coauthor graphs CS, Physics~\citep{DBLP:journals/corr/abs-1811-05868}, and six heterophilic benchmark datasets: Wikipedia graphs Chameleon, Squirrel~\citep{DBLP:journals/compnet/RozemberczkiAS21}, the Actor co-occurrence graph, webpage graphs Wisconsin, Texas, Cornell~\citep{DBLP:conf/iclr/PeiWCLY20}. The node classification tasks are conducted in the transductive setting. Following \citet{DBLP:conf/iclr/ChienP0M21}, we use the sparse splitting ($2.5\%/2.5\%/95\%$) and the dense splitting ($60\%/20\%/20\%$) to randomly split the homophilic and heterophilic graphs into training$/$validation$/$testing sets, respectively. Note that we directly used the data from Pytorch Geometric library~\citep{DBLP:journals/corr/abs-1903-02428} where they did \textbf{not} transform Chameleon and Squirrel to undirected graphs, which is different from \citet{DBLP:conf/iclr/ChienP0M21} where they did so. Dataset statistics and their levels of homophily are presented in \cref{App:exp_setup}.

\textbf{Baselines.} We compare \pgnn with seven models, including MLP, GCN~\citep{DBLP:conf/iclr/KipfW17}, SGC~\citep{DBLP:conf/icml/WuSZFYW19}, GAT~\citep{DBLP:conf/iclr/VelickovicCCRLB18}, JKNet~\citep{DBLP:conf/icml/XuLTSKJ18}, APPNP~\citep{DBLP:conf/iclr/KlicperaBG19}, GPRGNN~\citep{DBLP:conf/iclr/ChienP0M21}. We use the Pytorch Geometric library to implement all baselines except GPRGNN. For GPRGNN, we use the code released by the authors\footnote{\url{https://github.com/jianhao2016/GPRGNN}}. The details of hyperparameter settings are deferred to \cref{App:subsec:param}.

\begin{figure*}[htp]
	\centering
	\begin{subfigure}[b]{\textwidth}
		\centering
		\includegraphics[width=\textwidth]{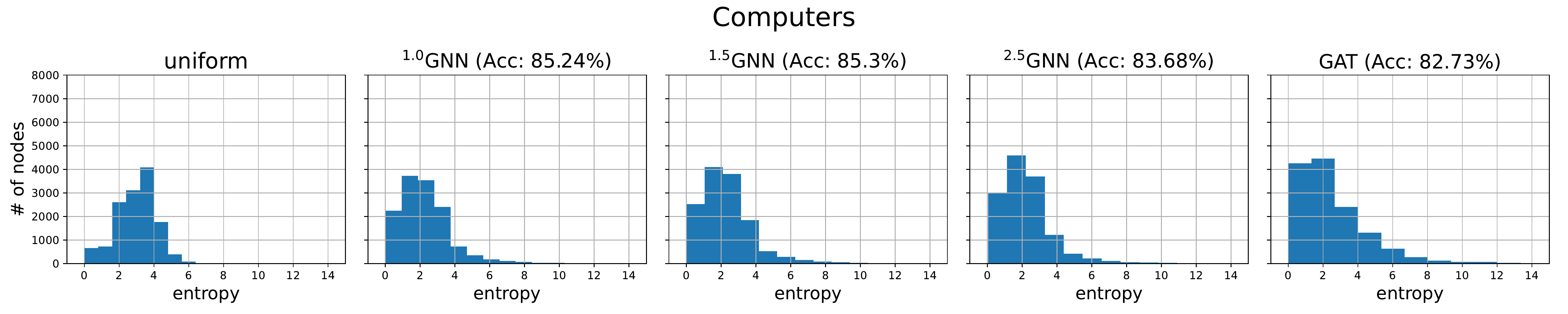}
	\end{subfigure}
	\begin{subfigure}[b]{\textwidth}
		\centering
		\includegraphics[width=\textwidth]{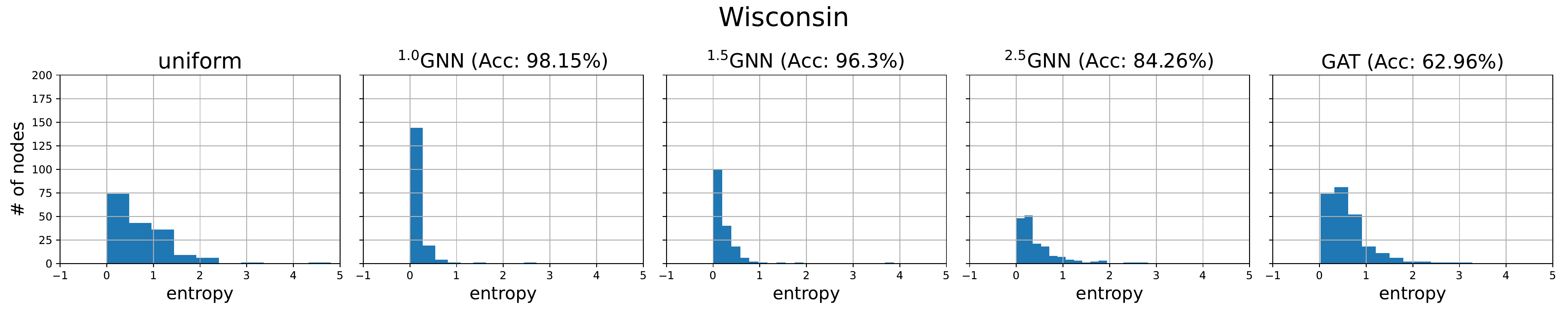}
	\end{subfigure}
	\caption{Aggregation weight entropy distribution of graphs. Low entropy means high degree of concentration, vice versa.  An entropy of zero means all aggregation weights are on one source node.}\label{fig:fig_aggr_exp}
\end{figure*}

\textbf{Superior Performance on Real-World  Heterophilic  Datasets.} The results on \emph{homophilic} benchmark datasets are deferred to \cref{App:homo_exp}, which show that \pgnns obtains competitive performance against state-of-the-art GNNs on homophilic datasets. \cref{tab:tab_heter_exp} summarizes the results on \emph{heterophilic} benchmark datasets. \cref{tab:tab_heter_exp} shows that \pgnns significantly dominate the baselines and $^{1.0}$GNN obtains the best performance on all heterophilic graphs except the Texas dataset. For Texas, $^{2.0}$GNN is the best. 
We also observe that MLP works very well and significantly outperforms most GNN baselines, which indicates that the graph topology is not informative for label prediction on these heterophilic graphs. Therefore, propagating and transforming node features over the graph topology could lead to worse performance than MLP. 
Unlike ordinary GNNs, \pgnns can adaptively learn aggregation weights and ignore edges that are not informative for label prediction and thus could work better. It confirms our theoretical findings presented in previous sections. Note that GAT can also learn aggregation weights, i.e. the attention weights. However, the aggregation weights learned by GAT are significantly distinct from that of \pgnns, as we will show following.

\textbf{Interpretability of the Learned Aggregation Weights of \pgnns.} We showcase the interpretability of the learned aggregation weights $\alpha_{i,i}\emD_{i,i}^{-1/2}\emM_{i,j}\emD_{j,j}^{-1/2}$ of \pgnns by studying its entropy distribution, along with the attention weights of GAT on real-world datasets. 
%
Denote $\{\emA_{i,j}\}_{j \in \gN_i}$ as the aggregation weights of node $i$ and its neighbors. For GAT, $\{\emA_{i,j}\}_{j \in \gN_i}$ are referred as the attention weights (in the first layer) and for \pgnns are $\alpha_{i,i}\emD_{i,i}^{-1/2}\emM_{i,j}\emD_{j,j}^{-1/2}$. For any node $i$, $\{\emA_{i,j}\}_{j \in \gN_i}$ forms a discrete probability distribution over all its neighbors with the entropy given by $H(\{\emA_{i,j}\}_{j \in \gN_i}) = -\sum_{j \in \gN_i}\emA_{i,j}\log(\emA_{i,j})$. Low entropy means high degree of concentration and vice versa. An entropy of zero means all aggregation weights or attentions are on one source node. The uniform distribution has the highest entropy of $log(\emD_{i,i})$. 
\cref{fig:fig_aggr_exp} reports the results on Computers, Wisconsin and we defer more results on other datasets to \cref{App:aggr} due to space limit. \cref{fig:fig_aggr_exp} shows that the aggregation weight entropy distributions of GAT and \pgnns on Computers (homophily) are both similar to the uniform case. It indicates the original graph topology of Computers is very helpful for label prediction and therefore GNNs could work very well on Computers. However, for Wisconsin (heterophily), the entropy distribution of \pgnns is significantly different from that of GAT and the uniform case. Most entropy of \pgnns is around zero, which means that most aggregation weights are on one source node. It states that the original graph topology of Wisconsin is not helpful for label prediction, which explains why MLP works well on Wisconsin.
On the contrary, the entropy distribution of GAT is similar to the uniform case and therefore GAT works similarly to GCN and is significantly worse than \pgnns on Wisconsin. 
Similar results can be observed on the experiments on more datasets in \cref{App:aggr}.

\begin{figure}[tp]
    \centering
    \includegraphics[width=0.48\textwidth]{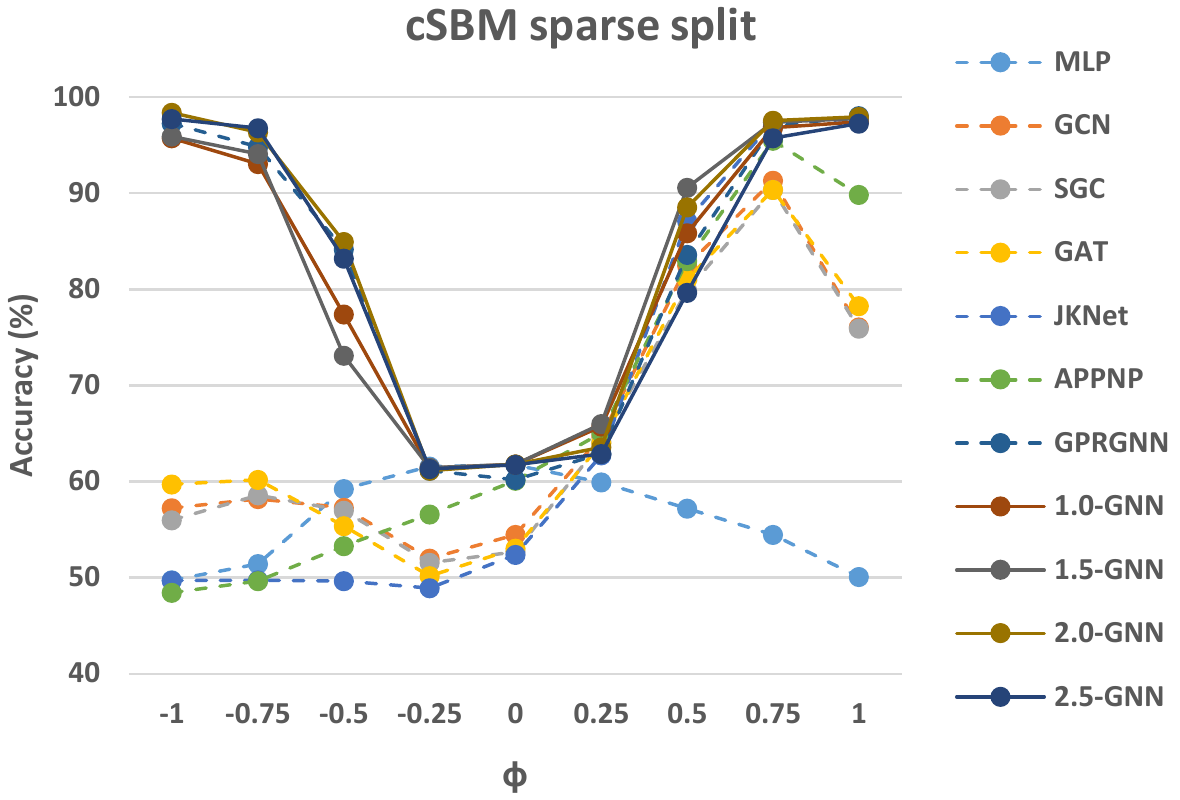}
    \caption{Averaged accuracy (\%) on cSBM (sparse split) for 20 runs. Best view in colors.}\label{fig:fig_csbm}
    \vspace{-5pt}
\end{figure}

\textbf{Results on cSBM Datasets.} We examine the performance of \pgnns on heterophilic graphs whose topology is informative for label prediction using synthetic graphs generated by cSBM~\citep{DBLP:conf/nips/DeshpandeSMM18} with $\phi \in \{-1, -0.75, \dots, 1\}$. We use the same settings of cSBM used in \citet{DBLP:conf/iclr/ChienP0M21}. Due to the space limit, we refer the readers to \citet{DBLP:conf/iclr/ChienP0M21} for more details of cSBM dataset. \cref{fig:fig_csbm} reports the results on cSBM using sparse splitting (for results on cSBM with dense splitting see \cref{App:csbm}). \cref{fig:fig_csbm} shows that when $\phi \leq -0.5$ (heterophilic graphs), $^{2.0}$GNN obtains the best performance and \pgnns and GPRGNN significantly dominate the others. It validates the effectiveness of \pgnns on heterophilic graphs. Moreover, $^{2.0}$GNN works better than GPRGNN and it again confirms that $^{2.0}$GNN is more superior under weak supervision (2.5\% training rate), as stated in \cref{remk:remk2}. 
Note that $^{1.0}$GNN and $^{1.5}$GNN are not better than $^{2.0}$GNN, the reason could be the iteration algorithms~\cref{eq:eq11} with $p=1, 1.5$ are not as stable as the one with $p=2$. When the graph topology is almost non-informative for label prediction ($\phi = -0.25, 0$), The performance of \pgnns is close to MLP and they outperform the other baselines. Again, it validates that \pgnns can erase non-informative edges and work as well as MLP and confirms the statements in \cref{the:the3}. When the graph is homophilic ($\phi \geq 0.25$), $^{1.5}$GNN is the best on weak homophilic graphs ($\phi = 0.25, 0.5$) and \pgnns work competitively with all GNN baselines on strong homophilic graphs ($\phi \geq 0.75$).

\begin{figure}[htp]
    \centering
	\begin{subfigure}[b]{0.45\textwidth}
		\includegraphics[width=\textwidth]{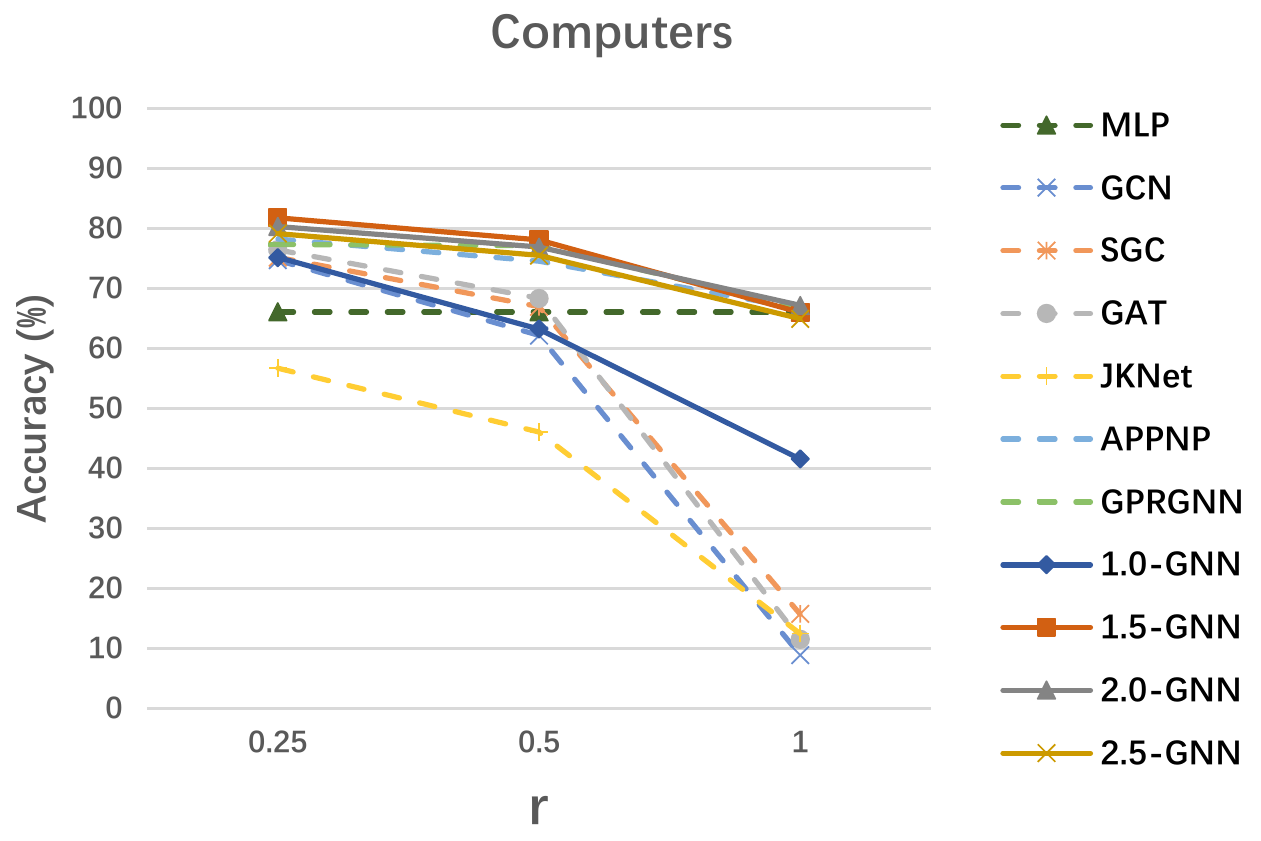}
	\end{subfigure}
	\begin{subfigure}[b]{0.45\textwidth}
		\includegraphics[width=\textwidth]{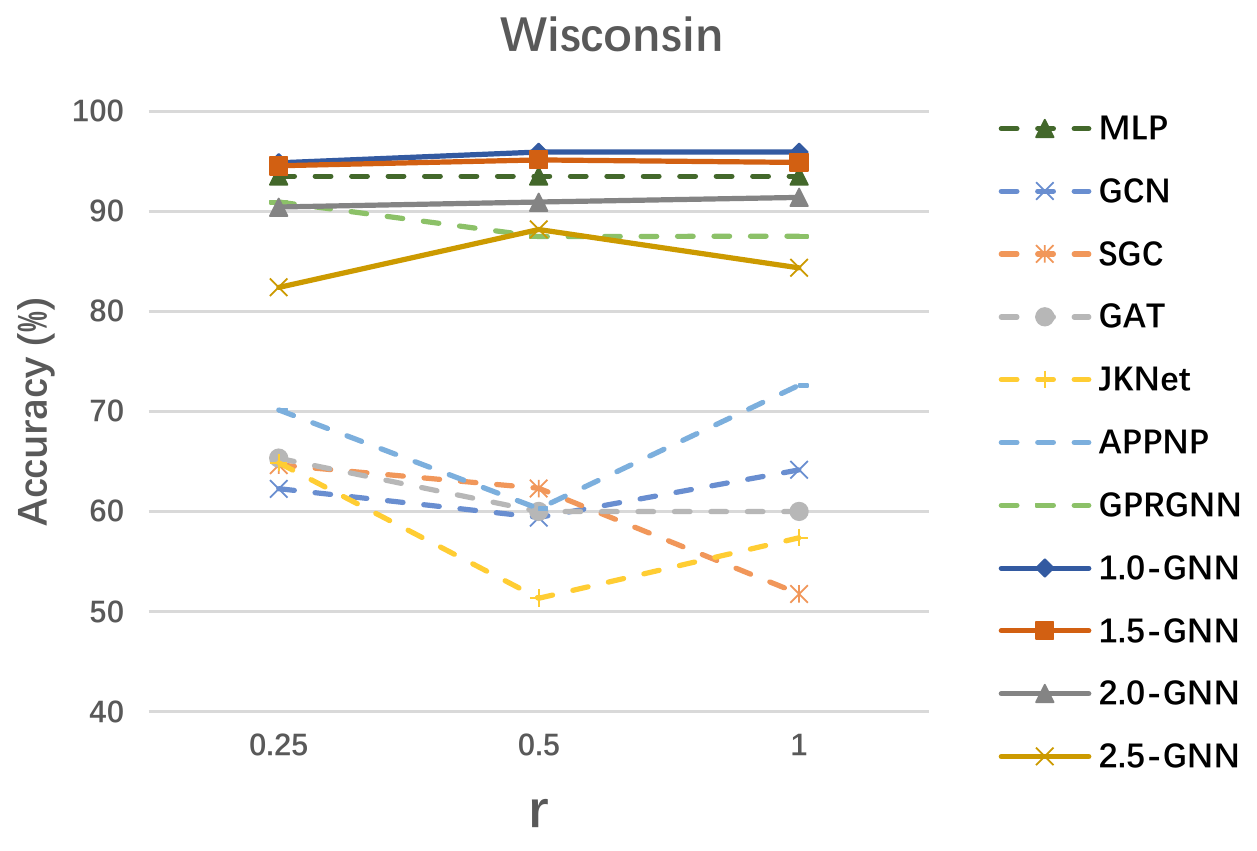}
	\end{subfigure}
	\caption{Averaged accuracy (\%) on graphs with noisy edges for 20 runs. Best view in colors.}\label{fig:noisy_exp}
	\vspace{-5pt}
\end{figure}

\textbf{Results on Datasets with Noisy Edges.} We conduct experiments to evaluate the performance of \pgnns on graphs with noisy edges by randomly adding edges to the graphs and randomly remove the same number of original edges. We define the random edge rate as $r := \frac{\#\text{random edges}}{\#\text{all edges}}$. The experiments are conducted on 4 homophilic datasets (Computers, Photo, CS, Physics) and 2 heterophilic datasets (Wisconsin, Texas) with $r = 0.25, 0.5, 1$. \cref{fig:noisy_exp} reports the results on Computers, Wisconsin and we defer more results to \cref{App:noisy_exp}. \cref{fig:noisy_exp} shows that \pgnns significantly outperform all baselines. Specifically, $^{1.5}$GNN obtains the best performance on Computers, and $^{1.5}$GNN and $^{2.0}$GNN even work as well as MLP on Computers with completely random edges ($r = 1$). For Wisconsin, $^{1.0}$GNN is the best, and $^{1.0}$GNN and $^{1.5}$GNN significantly dominate the others. We also observed that APPNP and GPRGNN, whose architectures are analogical to $^{2.0}$GNN, also work better than other GNNs. Nevertheless, they are significantly outperformed by \pgnns overall. Similar results can be observed in the experiments conducted on more datasets as presented in \cref{App:noisy_exp}.

%% file: sections/6-Conclusion.tex
\section{Conclusion}\label{sec:sec6}
We have addressed the problem of generalizing GNNs to heterophilic graphs and graphs with noisy edges.
To this end, we derived a novel $p$-Laplacian message passing scheme from a discrete regularization framework and proposed a new \pgnn architecture. We theoretically demonstrate our method works as low-high-pass filters and thereby applicable to both homophilic and heterophilic graphs. We empirically validate our theoretical results and show the advantages of our methods on heterophilic graphs and graphs with non-informative topologies.

Like most existing spectral based GNN models, e.g., GCN~\citep{DBLP:conf/iclr/KipfW17}, SGC~\citep{DBLP:conf/icml/WuSZFYW19}, the main restriction of $^p$GNNs is the relatively high space cost compared to GraphSage~\citep{DBLP:conf/nips/HamiltonYL17}, 
especially for extremely large graphs.
Integrating $^p$GNNs and $p$-Laplacian message passing with some popular subgraph sampling techniques so that $^p$GNNs (or its variants) could scale to large graphs would be an interesting future work. We refer the reader to \cref{App:sec:discussion} for further discussions on the potential extensions of \pgnns.

%% file: sections/6.1-Acknowledge.tex
\section*{Acknowledgements} 

Guoji would like to thank Prof. Xin Yao and Dr. Yunwen Lei for their sincere and selfless helps and supports.

%% file: sections/6.2-Statement.tex
\section*{Reproducibility Statement}

In order to ensure reproducibility, we have made the efforts in the following respects: (1) Provide our code as the supplementary material; (2) Provide self-contained proofs of the main claims in \cref{App:add_the,App:proof}; (3) Provide more details on experimental configurations in \cref{App:exp_setup} and experimental results in \cref{App:more_exp}.
All the datasets are publicly available as described in the main text. The code of \pgnns is available at \url{https://github.com/guoji-fu/pGNNs}.

%% file: sections/7-App_related.tex
\section{Related Work}\label{App:sec:related}

\textbf{Graph Neural Networks.} Graph neural networks (GNNs) are a variant of neural networks for graph-structured data, which can propagate and transform the node features over the graph topology and exploit the information in the graphs. Graph convolutional networks (GCNs) are one type of GNNs whose graph convolution mechanisms or the message passing schemes were mainly inspired by the field of graph signal processing. \citet{DBLP:journals/iclr/BrunaZSL14} defined a nonparametric graph filter using the Fourier coefficients. \citet{DBLP:conf/nips/DefferrardBV16} introduced Chebyshev polynomial to avoid computational expensive eigen-decomposition of Laplacian and obtain localized spectral filters. GCN~\citep{DBLP:conf/iclr/KipfW17} used the first-order approximation and reparameterized trick to simplify the spectral filters and obtain the layer-wise graph convolution. SGC~\citep{DBLP:conf/icml/WuSZFYW19} further simplify GCN by removing non-linear transition functions between each layer. \citet{DBLP:conf/iclr/ChenMX18} propose importance sampling to design an efficient variant of GCN. \citet{DBLP:conf/icml/XuLTSKJ18} explored a jumping knowledge architecture that flexibly leverages different neighborhood ranges for each node to enable better structure-aware representation. \citet{DBLP:conf/nips/AtwoodT16,DBLP:conf/iclr/LiaoZUZ19,DBLP:conf/uai/Abu-El-HaijaKPL19} exploited multi-scale information by diffusing multi-hop neighbor information over the graph topology. \citet{DBLP:journals/corr/abs-2002-06755} used label propagation to improve GCNs. \citet{DBLP:conf/iclr/KlicperaBG19} incorporated personalized PageRank with GCNs. \citet{DBLP:conf/icml/LiuJ0LLW0T21} introduced a $l_1$ norm-based graph smoothing term to enhance the local smoothnesss adaptivity of GNNs. 
\citet{DBLP:conf/nips/HamiltonYL17,DBLP:conf/iclr/ZengZSKP20} proposed sampling and aggregation frameworks to extent GCNs to inductive learning settings. Another variant of GNNs is graph attention networks~\citep{DBLP:conf/iclr/VelickovicCCRLB18,DBLP:journals/corr/abs-1803-03735,DBLP:conf/nips/Abu-El-HaijaPAA18}, which use attention mechanisms to adaptively learn aggregation weights based on the nodes features. There are many other works on GNNs~\citep{DBLP:conf/iclr/PeiWCLY20}~\citep{DBLP:conf/nips/YingY0RHL18,DBLP:conf/iclr/XinyiC19,DBLP:conf/iclr/VelickovicFHLBH19,DBLP:conf/iclr/ZengZSKP20}, we refer to \citet{DBLP:journals/aiopen/ZhouCHZYLWLS20,DBLP:journals/corr/abs-1806-01261,DBLP:journals/tnn/WuPCLZY21} for a  comprehensive review. Most GNN models implicitly assume that the labels of nodes and their neighbors should be the same or consistent, while it does not hold for heterophilic graphs. \citet{DBLP:conf/nips/ZhuYZHAK20} investigated the issues of GNNs on heterophilic graphs and proposed to separately learn the embeddings of ego-node and its neighborhood. \citet{DBLP:conf/aaai/ZhuR0MLAK21} proposed a framework to model the heterophily or homophily levels of graphs. \citet{DBLP:conf/iclr/ChienP0M21} incorporated generalized PageRank with graph convolution to adapt GNNs to heterophilic graphs. 

 There are also some works on the interpretability of GNNs proposed recently. \citet{DBLP:conf/aaai/LiHW18,DBLP:conf/nips/YingBYZL19,DBLP:journals/corr/abs-2006-04386} showed that spectral graph convolutions work as conducting Laplacian smoothing on the graph signals and \citet{DBLP:conf/icml/WuSZFYW19,DBLP:journals/corr/abs-1905-09550} demonstrated that GCN, SGC work as low-pass filters. \citet{DBLP:journals/tsp/GamaBR20} studied the stability properties of GNNs. \citet{DBLP:conf/iclr/XuHLJ19,DBLP:conf/iclr/OonoS20,DBLP:conf/iclr/Loukas20} studied the expressiveness of GNNs. \citet{DBLP:conf/kdd/VermaZ19,DBLP:conf/icml/GargJJ20} work on the generalization and representation power of GNNs.

\textbf{Graph based Semi-supervised Learning.}
Graph-based semi-supervised learning works under the assumption that the labels of a node and its neighbors shall be the same or consistent. Many methods have been proposed in the last decade, such as \citet{DBLP:conf/colt/SmolaK03,DBLP:conf/nips/ZhouBLWS03,DBLP:conf/colt/BelkinMN04} use Laplacian regularization techniques to force the labels of linked nodes to be the same or consistent. \citet{DBLP:conf/dagm/ZhouS05} introduce discrete regularization techniques to impose different regularizations on the node features based on $p$-Laplacian. Lable propagation~\citep{DBLP:conf/icml/ZhuGL03} recursively propagates the labels of labeled nodes over the graph topology and use the convergence results to make predictions. To mention but a few, we refer to \citet{DBLP:conf/dagm/ZhouS05,DBLP:journals/ml/EngelenH20} for a more comprehensive review.

\newpage

%% file: sections/8-App_discussion.tex
\section{Discussions and Future Work}\label{App:sec:discussion}
In this section, we discuss the future work of \pgnns. Our theoretical results and experimental results could lead to several potential extensions of \pgnns.

\textbf{New Paradigm of Designing GNN Architectures.} We bridge the gap between discrete regularization framework, graph-based semi-supervised learning, and GNNs, which provides a new paradigm of designing new GNN architectures. Following the new paradigm, researchers could introduce more regularization techniques, e.g., Laplacian regularization~\citep{DBLP:conf/colt/SmolaK03,DBLP:conf/colt/BelkinMN04}, manifold regularization~\citep{sindhwani2005linear,DBLP:conf/aistats/BelkinNS05,DBLP:journals/jmlr/Niyogi13}, high-order regularization~\citep{DBLP:journals/jmlr/ZhouB11}, Bayesian regularization~\citep{liu2014bayesian}, entropy regularization~\citep{DBLP:conf/nips/GrandvaletB04}, and consider more explicit assumptions on graphs, e.g. the homophily assumption, the low-density region assumption (i.e. the decision boundary is likely to lie in a low data density region), manifold assumption (i.e. the high dimensional data lies on a low-dimensional manifold), to develop new graph convolutions or message passing schemes for graphs with specific properties and generalize GNNs to a much broader range of graphs. Moreover, the paradigm also enables us to explicitly study the behaviors of the designed graph convolutions or message passing schemes from the theory of regularization~\citep{DBLP:journals/jcss/BelkinN08,DBLP:journals/jmlr/Niyogi13,DBLP:journals/corr/SlepcevT17}.

\textbf{Applications of \pgnns to learn on graphs with noisy topologies.} The empirical results (as shown in \cref{fig:noisy_exp} and \cref{App:tab:noisy_exp_homo,App:tab:noisy_exp_heter}) on graphs with noisy edges show that \pgnns are very robust to noisy edges, which suggests the applications of $p$-Laplacian message passing and \pgnns on the graph learning scenarios where the graph topology could potentially be seriously intervened.

\textbf{Integrating with existing GNN architectures.} As shown in \cref{App:tab:plug_exp_heter}, the experimental results on heterophilic benchmark datasets illustrate that integrating GCN, JKNet with \pgnns can significantly improve their performance on heterophilic graphs. It shows that \pgnn could be used as a plug-and-play component to be integrated into existing GNN architectures and improve their performance on real-world applications.

\textbf{Inductive learning for \pgnns.} \pgnns are shown to be very effective for inductive learning on PPI datasets as reported in \cref{App:tab:ppi_exp}. \pgnns even outperforms GAT on PPI, while using much fewer parameters than GAT. It suggests  the promising extensions of \pgnns to inductive learning on graphs.

\newpage

%% file: sections/9-App_theorem.tex
\section{Additional Theorems}\label{App:add_the}
\subsection{Theorem 4 (Upper-Bounding Risk of $^p$GNN)}\label{App:the:the3}
\begin{theorem}[Upper-Bounding Risks of \pgnns]\label{the:the3}
	Given a graph $\gG = (\gV, \gE, \mW)$ with $N$ nodes, let $\mX \in \R^{N \times c}$ be the node features and $\vy \in \R^N$ be the node labels and $\mM^{(k)}, \bm{\alpha}^{(k)}, \bm{\beta}^{k}, \mF^{k}$ are updated accordingly by \cref{eq:eq12,eq:eq13,eq:eq15} for $k = 0, 1, \dots, K-1$ and $\mF^{(0)} = \mX$, $K \in \sN$.
	Assume that $\gG$ is $d$-regular and the ground-truth node features $\mX^* = \mX + \bm{\epsilon}$, where $\bm{\epsilon} \in \R^{N \times c}$ represents the noise in the node features and there exists a $L$-Lipschitz function $\sigma: \R^{N \times c} \rightarrow \R^N$ such that $\sigma(\mX^*) = \vy$.  let $\tilde{\vy}^{(k+1)} = \bm{\alpha}^{(k)}\bm{\mD}^{-1/2}\mM^{(k)}\bm{\mD}^{-1/2}\sigma(\mF^{(k)}) + \bm{\beta}^{(k)}\sigma\left(\mF^{(0)}\right)$, we have
	\begin{align*}
		\frac{1}{N}\sum_{i=1}^N\left|\evy_i - \tilde{\evy}_i\right| \leq {} & \frac{1}{N}\sum_{i=1}^N\beta_{i,i}^{(K-1)}\left|\evy_i - \sigma(\mX_{i,:})\right| \\ {} & \qquad + \frac{L}{N}\sum_{i=1}^N\alpha_{i,i}^{(K-1)}\left\|\Delta_p^{(K-1)}\mF_{i,:}^{(K-1)} + \sum_{k=0}^{K-2}\prod_{l=k}^{K-2}\left(\sum_{j=1}^N\frac{\emM_{i,j}^{(l)}}{d}\right)\Delta_p^{(k)}\mX_{i,:}\right\| \\ {} & \qquad + \frac{L}{N}\sum_{i=1}^N(1 -\beta_{i,i}^{(K-1)})\left\|\bm{\epsilon}_{i,:}\right\|.
	\end{align*}
\end{theorem}
\vspace{-10pt}
Proof see \cref{App:proof:the3}. \cref{the:the3} shows that the risk of $^p$GNNs is upper-bounded by the sum of three terms: The first term of the r.h.s in the above inequation represents the risk of label prediction using only the original node features $\mX$, the second term is the norm of $p$-Laplacian diffusion on the node features $\mX$, and the third term is the magnitude of the noise in the node features. $\alpha_{i,i}$ and $\beta_{i,i}$ control the trade-off between these three terms and they are related to the hyperparameter $\mu$ in \cref{eq:eq10}. The smaller $\mu$, the smaller $\beta_{i,i}$ and larger $\alpha_{i,i}$, thus the more important of the $p$-Laplacian diffusion term but also the more effect from the noise. 
Therefore, for graphs whose topological information is not helpful for label prediction, we could impose more weights on the first term by using a large $\mu$ so that \pgnns work more like MLPs which simply learn on node features. While for graphs whose topological information is helpful for label prediction, we could impose more weights on the second term by using a small $\mu$ so that \pgnns can benefit from $p$-Laplacian smoothing on node features. 

In practice, to choose a proper value of $\mu$ one may first simply apply MLPs on the node features to have a glance at the helpfulness of the node features. If MLPs work very well, there is not much space for the graph's topological information to further improve the prediction performance and we may choose a large $\mu$. Otherwise, there could be a large chance for the graph's topological information to further improve the performance and we should choose a small $\mu$.

\subsection{Theorem 5 ($p$-Orthogonal Theorem (Luo et al., 2010))}\label{App:the:the4}
\begin{theorem}[$p$-Orthogonal Theorem~\citep{DBLP:journals/ml/LuoHDN10}]\label{the:the4}
	If $\vu^{(l)}$ and $\vu^{(r)}$ are two eigenvectors of $p$-Laplacian $\Delta_p$ associated with two different non-zero eigenvalues $\lambda_l$ and $\lambda_r$, $\mW$ is symmetric and $p \geq 1$, then $\vu^{(l)}$ and $\vu^{(r)}$ are $p$-orthogonal up to the second order Taylor expansion.
\end{theorem}
\cref{the:the4} implies that $\phi_p(\vu)^{(l)\top}\phi_p(\vu^{(r)}) \approx 0$, for all $l, r = 0, \dots, N-1$ and $\lambda_l \neq \lambda_r$. Therefore, the space spanned by the multiple eigenvectors of the graph $p$-Laplacian is $p$-orthogonal.

\subsection{Theorem 6 ($p$-Eigen-Decomposition of $\Delta_p$)}\label{App:the:the5}
\begin{theorem}[$p$-Eigen-Decomposition of $\Delta_p$]\label{the:the5}
	Given the $p$-eigenvalues $\{\lambda_l \in \R\}_{l=0, 1, \dots, N-1}$, and the $p$-eigenvectors $\{\vu^{(l)} \in \R^N\}_{l=0, 1, \dots, N-1}$ of $p$-Laplacian $\Delta_p$ and $\|\vu^{(l)}\|_p = (\sum_{i=1}^N(\evu_i^{(l)})^p)^{1/p} = 1$, let $\mU$ be a matrix of $p$-eigenvectors with $\mU = (\vu^{(0)}, \vu^{(1)}, \dots, \vu^{(N-1)})$ and $\mLambda$ be a diagonal matrix with $\mLambda = \mathrm{diag}(\lambda_0, \lambda_1, \dots, \lambda_{N-1})$, then the $p$-eigen-decomposition of $p$-Laplacian $\Delta_p$ is given by
	\begin{align*}
		\Delta_p = \Phi_p(\mU) \Lambda \Phi_p(\mU)^\top.
	\end{align*}
	When $p=2$, it reduces to the standard eigen-decomposition of the Laplacian matrix.
\end{theorem}
Proof see \cref{App:proof:the5}.

\subsection{Theorem 7 (Bounds of $p$-Eigenvalues)}\label{App:the:the7}
\begin{theorem}[Bounds of $p$-Eigenvalues]\label{the:the7}
	Given a graph $\gG = (\gV, \gE, \mW)$, if $\gG$ is connected and $\lambda$ is a $p$-eigenvalue associated with the $p$-eigenvector $\vu$ of $\Delta_p$, let $N_i$ denotes the number of edges connected to node $i$, $N_{\text{min}} = \min\{N_i\}_{i=1,2,\dots,N}$, and $k = \argmax(\{|\evu_i| / \sqrt{\emD_{i,i}}\}_{i=1, 2, \dots, N})$, then
	\begin{enumerate}
		\item for $p \geq 2$, $0 \leq \lambda \leq 2^{p-1}$; 
		
		\item for $1 < p < 2$, $0 \leq \lambda \leq 2^{p-1}\sqrt{N_k}$;
		
		\item for $p = 1$, $0 \leq \lambda \leq \sqrt{N_{\text{min}}}$.
	\end{enumerate}
\end{theorem}
Proof see \cref{App:proof:the7}.


\section{Proof of Theorems}\label{App:proof}

\subsection{Proof of Theorem 1}\label{App:the1}
\begin{proof}
    Let $\bm{i}$ be the one-hot indicator vector whose $i$-th element is one and the other elements are zero. Then, we can obtain the personalized PageRank on node $i$, denoted as $\bm{\pi}_{\text{PPR}}(\bm{i})$, by using the recurrent equation~\citep{DBLP:conf/iclr/KlicperaBG19}: 
    \begin{equation*}
        \bm{\pi}_{\text{PPR}}^{(k+1)}(\bm{i}) = \alpha\mD^{-1/2}\mW\mD^{-1/2}\bm{\pi}_{\text{PPR}}^{(k)}(\bm{i}) + \beta\bm{i},
    \end{equation*}
    where $k$ is the iteration step, $0 < \alpha < 1$ and $\beta = (1 - \alpha)$ represents the restart probability. Without loss of generality, suppose $\bm{\pi}_{\text{PPR}}^{(0)}(\bm{i}) = \bm{i}$. Then we have,
	\begin{equation*}
		\begin{aligned}
			\bm{\pi}_{\text{PPR}}^{(k)}(\bm{i}) = {} & \alpha\mD^{-1/2}\mW\mD^{-1/2}\bm{\pi}_{\text{PPR}}^{(k-1)}(\bm{i}) + \beta\bm{i} \\
			= {} & \alpha\mD^{-1/2}\mW\mD^{-1/2}\left(\alpha\bm{D}^{-1/2}\bm{W}\bm{D}^{-1/2}\bm{\pi}_{\text{PPR}}^{(k-2)}(\bm{i}) + \beta\bm{i}\right) + \beta\bm{i} \\
			= {} & \left(\alpha\mD^{-1/2}\mW\mD^{-1/2}\right)^2\bm{\pi}_{\text{PPR}}^{(t-2)}(\bm{i}) + \beta\alpha\mD^{-1/2}\mW\mD^{-1/2}\bm{i} + \beta\bm{i} \\
			= {} & \left(\alpha\mD^{-1/2}\mW\mD^{-1/2}\right)^k\bm{\pi}_{\text{PPR}}^{(0)}(\bm{i}) + \beta\sum_{t=0}^{k-1}\left(\alpha\mD^{-1/2}\mW\mD^{-1/2}\right)^t\bm{i} \\
			= {} & \left(\alpha\mD^{-1/2}\mW\mD^{-1/2}\right)^k\bm{i} + \beta\sum_{t=0}^{k-1}\left(\alpha\mD^{-1/2}\mW\mD^{-1/2}\right)^t\bm{i} \\
		\end{aligned}
	\end{equation*}
	Since $0 < \alpha < 1$ and the eigenvalues of $\mD^{-1/2}\mW\mD^{-1/2}$ in $[-1, 1]$, we have
	\begin{equation*}
		\lim_{k \rightarrow \infty}\left(\alpha\mD^{-1/2}\mW\mD^{-1/2}\right)^k = 0,
	\end{equation*}
	and we also have
	\begin{equation*}
		\lim_{k \rightarrow \infty}\sum_{t=0}^{k-1}\left(\alpha\mD^{-1/2}\mW\mD^{-1/2}\right)^t = \left(\mI_N - \alpha\mD^{-1/2}\mW\mD^{-1/2}\right)^{-1}.
	\end{equation*}
	Therefore,
	\begin{equation*}
		\begin{aligned}
			\bm{\pi}_{\text{PPR}}(\bm{i}) = \lim_{k \rightarrow \infty}\bm{\pi}_{\text{PPR}}^{(k)}(\bm{i}) = {} & \beta\left(\mI_N - \alpha\mD^{-1/2}\mW\mD^{-1/2}\right)^{-1}\bm{i} \\
			= {} & \beta\left(\alpha\Delta_2 + (1-\alpha)\mI_N\right)^{-1}\bm{i} \\
			= {} & \mu(\Delta_2 + \mu\mI_N)^{-1}\bm{i},
		\end{aligned}
	\end{equation*}
    where we let $\alpha = \frac{1}{1 + \mu}$ and $\beta = \frac{\mu}{1 + \mu}$, $\mu > 0$. Then the fully personalized PageRank matrix can be obtained by substituting $\bm{i}$ with $\mI_N$:
    \begin{equation*}
        \bm{\Pi}_{\text{PPR}} = \mu(\Delta_2 + \mu\mI_N)^{-1}.
    \end{equation*}
\end{proof}

\subsection{Proof of Theorem 2}\label{App:the2}
\begin{proof}
	By the definition of $\gL_p(f)$ in \cref{eq:eq8}, we have for some positive real value $\mu, \mu > 0$
	\begin{equation*}
		\gL_p(\mF) = \frac{1}{2}\sum_{i=1}^N\sum_{j=1}^N\left\|\sqrt{\frac{\emW_{i,j}}{\emD_{i,i}}}\mF_{i,:} - \sqrt{\frac{\emW_{i,j}}{\emD_{j,j}}}\mF_{j,:}\right\|^p + \mu\sum_{i=1}^N\|\mF_{i,:} - \mX_{i,:}\|^2.
	\end{equation*}
	and by \cref{eq:eq12},
	\begin{equation*}
		\emM_{i,j}^{(k)} := \emW_{i,j}\left\|\sqrt{\frac{\emW_{i,j}}{\emD_{i,i}}}\mF_{i,:}^{(k)} - \sqrt{\frac{\emW_{i,j}}{\emD_{j,j}}}\mF_{j,:}^{(k)}\right\|^{p-2}
	\end{equation*}
	Then, we have
	\begin{align*}
		\frac{\partial\gL_p(\mF^{(k)})}{\partial\mF_{i,:}^{(k)}} = {} & p\sum_{j=1}^N\sqrt{\frac{\emW_{i,j}}{\emD_{i,i}}}\left\|\sqrt{\frac{\emW_{i,j}}{\emD_{i,i}}}\mF_{i,:}^{(k)} - \sqrt{\frac{\emW_{i,j}}{\emD_{j,j}}}\mF_{j,:}^{(k)}\right\|^{p-2}\left(\sqrt{\frac{\emW_{i,j}}{\emD_{i,i}}}\mF_{i,:}^{(k)} - \sqrt{\frac{\emW_{i,j}}{\emD_{j,j}}}\mF_{j,:}^{(k)}\right) + 2\mu(\mF_{i,:}^{(k)} - \mX_{i,:}) \\
		= {} & p\left(\sum_{j=1}^N\frac{\emM_{i,j}^{(k)}}{\emD_{i,i}}\mF_{i,:}^{(k)} - \sum_{j=1}^N\frac{\emM_{i,j}^{(k)}}{\sqrt{\emD_{i,i}\emD_{j,j}}}\mF_{j,:}^{(k)}\right) + 2\mu(\mF_{i,:}^{(k)} - \mX_{i,:}) \\
		= {} & p\left(\left(\sum_{j=1}^N\frac{\emM_{i,j}^{(k)}}{\emD_{i,i}} + \frac{2\mu}{p}\right)\mF_{i,:}^{(k)} - \left(\sum_{j=1}^N\frac{\emM_{i,j}^{(k)}}{\sqrt{\emD_{i,i}\emD_{j,j}}}\mF_{j,:}^{(k)} + \frac{2\mu}{p}\mX_{i,:}\right)\right) \\
		= {} & \frac{p}{\alpha_{i,i}^{(k)}}\left(\mF_{i,:}^{(k)} - \left(\alpha_{i,i}^{(k)}\sum_{j=1}^N\frac{\emM_{i,j}^{(k)}}{\sqrt{\emD_{i,i}\emD_{j,j}}}\mF_{j,:}^{(k)} + \beta_{i,i}^{(k)}\mX_{i,:}\right)\right) \\
		= {} & \frac{p}{\alpha_{i,i}^{(k)}}\left(\mF_{i,:}^{(k)} - \mF_{i,:}^{(k+1)}\right),
	\end{align*}
	which indicates that
	\begin{equation*}
		\mF_{i,:}^{(k)} - \mF_{i,:}^{(k+1)} = \frac{\alpha_{i,i}^{(k)}}{p} \cdot \frac{\partial\gL_p(\mF^{(k)})}{\partial\mF_{i,:}^{(k)}}.
	\end{equation*}
	
	For all $i, j \in [N], \vv \in \R^{1 \times c}$, denote by
	\begin{align*}
	    \partial\gL_p(\mF_{i,:}^{(k)}) := {} & \frac{\partial\gL_p(\mF^{(k)})}{\partial\mF_{i,:}^{(k)}}, \\
	    \emM_{i,j}'^{(k)} := {} & \emW_{i,j}\left\|\sqrt{\frac{\emW_{i,j}}{\emD_{i,i}}}(\mF_{i,:}^{(k)} + \vv) - \sqrt{\frac{\emW_{i,j}}{\emD_{j,j}}}\mF_{j,:}^{(k)}\right\|^{p-2}, \\
	     \alpha_{i,i}'^{(k)} := {} & 1 \left/ \left(\sum_{j=1}^N\frac{\emM_{i,j}'^{(k)}}{\emD_{i,i}} + \frac{2\mu}{p}\right)\right., \\
	     \beta_{i,i}'^{(k)} := {} & \frac{2\mu}{p}\alpha_{i,i}'^{(k)} \\
	    \mF_{i,:}'^{(k+1)} := {} & \alpha_{i,i}'^{(k)}\sum_{j=1}^N\frac{\emM_{i,j}'^{(k)}}{\sqrt{\emD_{i,i}\emD_{j,j}}}\mF_{j,:}^{(k)} + \beta_{i,i}'\mX_{i,:}.
	\end{align*}
    Then
	\begin{align*}
	    {} & \left\|\partial\gL_p(\mF_{i,:}^{(k)} + \vv) - \partial\gL_p(\mF_{i,:}^{(k)})\right\| \\
	    = {} & \left\|\frac{p}{\alpha_{i,i}'^{(k)}}\left(\mF_{i,:}^{(k)} + \vv - \mF_{i,:}'^{(k+1)}\right) - \frac{p}{\alpha_{i,i}^{(k)}}\left(\mF_{i,:}^{(k)} - \mF_{i,:}^{(k+1)}\right)\right\| \\
	    \leq {} & \frac{p}{\alpha_{i,i}'^{(k)}}\|\vv\| + \left\|\frac{p}{\alpha_{i,i}'^{(k)}}\left(\mF_{i,:}^{(k)} - \mF_{i,:}'^{(k)}\right) - \frac{p}{\alpha_{i,i}^{(k)}}\left(\mF_{i,:}^{(k)} - \mF_{i,:}^{(k+1)}\right)\right\| \\
	    = {} & \frac{p}{\alpha_{i,i}'^{(k)}}\|\vv\| + \left\|\left(\frac{p}{\alpha_{i,i}'^{(k)}} - \frac{p}{\alpha_{i,i}^{(k)}}\right)\mF_{i,:}^{(k)} -  \frac{p}{\alpha_{i,i}'^{(k)}}\mF_{i,:}'^{(k+1)} +  \frac{p}{\alpha_{i,i}^{(k)}}\mF_{i,:}^{(k+1)}\right\| \\
	    = {} & \frac{p}{\alpha_{i,i}'^{(k)}}\|\vv\| + p\left\|\left(\sum_{j=1}^N\frac{\emM_{i,j}'^{(k)}}{\emD_{i,i}} - \sum_{j=1}^N\frac{\emM_{i,j}^{(k)}}{\emD_{i,i}}\right)\mF_{i,:}^{(k)} -  \sum_{j=1}^N\frac{\emM_{i,j}'^{(k)}}{\sqrt{\emD_{i,i}\emD_{j,j}}}\mF_{j,:}^{(k)} - \frac{2\mu}{p}\mX_{i,:} +  \sum_{j=1}^N\frac{\emM_{i,j}^{(k)}}{\sqrt{\emD_{i,i}\emD_{j,j}}}\mF_{j,:}^{(k)} + \frac{2\mu}{p}\mX_{i,:}\right\| \\
	    = {} & \frac{p}{\alpha_{i,i}'^{(k)}}\|\vv\| + p\left\|\left(\sum_{j=1}^N\frac{\emM_{i,j}'^{(k)}}{\emD_{i,i}} - \sum_{j=1}^N\frac{\emM_{i,j}^{(k)}}{\emD_{i,i}}\right)\mF_{i,:}^{(k)} -  \sum_{j=1}^N\frac{\emM_{i,j}'^{(k)}}{\sqrt{\emD_{i,i}\emD_{j,j}}}\mF_{j,:}^{(k)} +  \sum_{j=1}^N\frac{\emM_{i,j}^{(k)}}{\sqrt{\emD_{i,i}\emD_{j,j}}}\mF_{j,:}^{(k)}\right\| \\
	    = {} & \left(p\sum_{j=1}^N\frac{\emM_{i,j}^{(k)}}{\emD_{i,i}} + 2\mu\right)\|\vv\| + p\sum_{j=1}^N\frac{\emM_{i,j}'^{(k)} - \emM_{i,j}^{(k)}}{\emD_{i,i}}\|\vv\| + p\left\|\sum_{j=1}^N\frac{\emM_{i,j}'^{(k)} - \emM_{i,j}^{(k)}}{\emD_{i,i}}\mF_{i,:}^{(k)} -  \sum_{j=1}^N\frac{\emM_{i,j}'^{(k)} - \emM_{i,j}^{(k)}}{\sqrt{\emD_{i,i}\emD_{j,j}}}\mF_{j,:}^{(k)}\right\| \\
	    = {} & p\left(\sum_{j=1}^N\frac{\emM_{i,j}^{(k)}}{\emD_{i,i}} + \frac{2\mu}{p} + o\left(p, \vv, \mX, \gG\right)\right)\|\vv\|. \\
	\end{align*}
    Therefore, there exists some real positive value $\mu \in o\left(p, \vv, \mX, \gG\right) > 0$ such that
    \begin{equation}\label{App:eq:eq_lip}
        \left\|\partial\gL_p(\mF_{i,:}^{(k)} + \vv) - \partial\gL_p(\mF_{i,:}^{(k)})\right\| \leq p\big(\sum_{j=1}^N\frac{\emM_{i,j}^{(k)}}{\emD_{i,i}} + \frac{2\mu}{p}\big)\|\vv\| = \frac{p}{\alpha_{i,i}^{(k)}}\|\vv\|.
    \end{equation}
    Let $\bm{\gamma} = (\gamma_1, \dots, \gamma_N)^\top \in \R^{N}$ and $\bm{\eta} \in \R^{N \times c}$. By Taylor's theorem, we have:
    \begin{align*}
        {} & \gL_p(\mF_{i,:}^{(k)} + \gamma_i\bm{\eta}_{i,:}) \\
        = {} & \gL_p(\mF_{i,:}^{(k)}) + \gamma_i\int_0^1\langle\partial\gL_p(\mF_{i,:}^{(k)} + \epsilon\gamma_i\bm{\eta}_{i,:}), \bm{\eta}_{i,:}\rangle\mathrm{d}\epsilon \\
        = {} & \gL_p(\mF_{i,:}^{(k)}) + \gamma_i\langle\bm{\eta}_{i,:}, \partial\gL_p(\mF_{i,:}^{(k)})\rangle + \gamma_i\int_0^1\langle\partial\gL_p(\mF_{i,:}^{(k)} + \epsilon\gamma_i\bm{\eta}_{i,:}) - \partial\gL_p(\mF_{i,:}^{(k)}), \bm{\eta}_{i,:}\rangle\mathrm{d}\epsilon \\
        \leq {} & \gL_p(\mF_{i,:}^{(k)}) + \gamma_i\langle\bm{\eta}_{i,:}, \partial\gL_p(\mF_{i,:}^{(k)})\rangle + \gamma_i\int_0^1\|\partial\gL_p(\mF_{i,:}^{(k)} + \epsilon\gamma_i\bm{\eta}_{i,:}) - \partial\gL_p(\mF_{i,:}^{(k)})\|\|\bm{\eta}_{i,:}\|\mathrm{d}\epsilon \\
        \leq {} & \gL_p(\mF_{i,:}^{(k)}) + \gamma_i\langle\bm{\eta}_{i,:}, \partial\gL_p(\mF_{i,:}^{(k)})\rangle + \frac{p}{2\alpha_{i,i}^{(k)}}\gamma_i^2\|\bm{\eta}_{i,:}\|^2 \\
    \end{align*}
    Let $\bm{\eta} = -\nabla\gL_p(\mF^{(k)})$ and choose some positive real value $\mu$ which depends on $\mX, \gG, p$ and $p > 1$, i.e. $\mu \in o\left(p, \mX, \gG\right)$. By \cref{App:eq:eq_lip}, we have for all $i \in [N]$,
    \begin{align*}
        \gL_p(\mF_{i,:}^{(k)} - \gamma\partial\gL_p(\mF_{i,:}^{(k)})) \leq {} & \gL_p(\mF_{i,:}^{(k)}) - \langle\gamma_i\partial\gL_p(\mF_{i,:}^{(k)}), \partial\gL_p(\mF_{i,:}^{(k)})\rangle + \frac{p}{2\alpha_{i,i}^{(k)}}\gamma_i^2\|\partial\gL_p(\mF_{i,:}^{(k)})\|^2 \\
        = {} & \gL_p(\mF_{i,:}^{(k)}) - \frac{p}{2\alpha_{i,i}^{(k)}}\left(\frac{2\alpha_{i,i}^{(k)}\gamma_i}{p} - \gamma_i^2\right)\|\partial\gL_p(\mF_{i,:})\|^2 \\
        = {} & \gL_p(\mF_{i,:}^{(k)}) - \frac{p}{2\alpha_{i,i}^{(k)}}\left(\frac{\left(\alpha_{i,i}^{(k)}\right)^2}{p^2} - \left(\gamma_i - \frac{\alpha_{i,i}^{(k)}}{p}\right)^2\right)\|\partial\gL_p(\mF_{i,:}^{(k)})\|^2.
    \end{align*}
    Then for all $i \in [N]$, when $0 \leq \gamma_i \leq \frac{2\alpha_{i,i}^{(k)}}{p}$, we have $\gS_p(\mF_{i,:}^{(k)} - \gamma_i\partial\gS_p(\mF_{i,:}^{(k)})) \leq \gS_p(\mF_{i,:}^{(k)})$ and $\gamma_i = \frac{\alpha_{i,i}^{(k)}}{p}$ minimizes $\gS_p(\mF_{i,:}^{(k)} - \gamma_i\partial\gS_p(\mF_{i,:}^{(k)}))$. Therefore,
    \begin{equation*}
        \gL_p(\mF^{(k+1)}) = \gL_p(\mF^{(k)} - \frac{1}{p} \cdot \bm{\alpha}^{(k)}\nabla\gL_p(\mF^{(k)})) \leq \gL_p(\mF^{(k)}).
    \end{equation*}
\end{proof}

\subsection{Proof of Theorem 3}\label{App:the6}
\begin{proof}
	Without loss of generality, suppose $\mF^{(0)} = \mX$. Denote $\tilde{\mM}^{(k)} = \mD^{-1/2}\rmM^{(k)}\mD^{-1/2}$, by \cref{eq:eq15}, we have for $K \geq 2$,
	\begin{align}
		\mF^{(K)} = {} & \bm{\alpha}^{(K-1)}\mD^{-1/2}\mM^{(K-1)}\mD^{-1/2}\mF^{(K-1)} + \bm{\beta}^{(K-1)}\mX \notag \\
		= {} & \bm{\alpha}^{(K-1)}\tilde{\mM}^{K-1}\mF^{(K-1)} + \bm{\beta}^{(K-1)}\mX \notag \\
		= {} & \bm{\alpha}^{(K-1)}\tilde{\mM}^{K-1}\left(\bm{\alpha}^{(K-2)}\tilde{\mM}^{K-2}\mF^{(K-2)} + \bm{\beta}^{(K-2)}\mX\right) + \bm{\beta}^{(K-1)}\mX \notag \\
		= {} & \bm{\alpha}^{(K-1)}\bm{\alpha}^{(K-2)}\tilde{\mM}^{K-1}\tilde{\mM}^{K-2}\mF^{(K-2)} + \bm{\alpha}^{(K-1)}\tilde{\mM}^{K-1}\bm{\beta}^{(K-2)}\mX + \bm{\beta}^{(K-1)}\mX \notag \\
		= {} & \left(\prod_{k=0}^{K-1}\bm{\alpha}^{(k)}\right)\left(\prod_{k=0}^{K-1}\tilde{\mM}^{(k)}\right)\mF^{(0)} + \sum_{k=1}^{K-1}\left(\prod_{l=K-k}^{K-1}\bm{\alpha}^{(l)}\tilde{\mM}^{(l)}\right)\bm{\beta}^{(K-1-k)}\mX + \bm{\beta}^{(K-1)}\mX \notag \\
		= {} & \left(\prod_{k=0}^{K-1}\bm{\alpha}^{(k)}\right)\left(\prod_{k=0}^{K-1}\tilde{\mM}^{(k)}\right)\mX + \sum_{k=1}^{K-1}\left(\prod_{l=K-k}^{K-1}\bm{\alpha}^{(l)}\tilde{\mM}^{(l)}\right)\bm{\beta}^{(K-1-k)}\mX + \bm{\beta}^{(K-1)}\mX. \label{eq:eq23}
	\end{align}
	Recall \cref{eq:eq12,eq:eq13}, we have 
	\begin{equation*}
		\tilde{\emM}_{i,j}^{(k)} = \frac{\emW_{i, j}}{\sqrt{\emD_{i,i}\emD_{j,j}}}\left\|\sqrt{\frac{\emW_{i,j}}{\emD_{i,i}}}\mF_{i,:}^{(k)} - \sqrt{\frac{\emW_{i,j}}{\emD_{j,j}}}\mF_{j,:}^{(k)}\right\|^{p-2}, \text{ for all } i, j = 1, 2, \dots, N,
	\end{equation*}
	and 
	\begin{equation*}
		\alpha_{i,i}^{(k)} = \frac{1}{{\sum_{j=1}^N\frac{\emM_{i,j}^{(k)}}{\emD_{i,i}} + \frac{2\mu}{p}}}, \text{ for all } i = 1, 2, \dots, N.
	\end{equation*}
	Note that the eigenvalues of $\tilde{\mM}$ are not infinity and $0 < \alpha_{i,i} < 1$ for all $i = 1, \dots, N$. Then we have
	\begin{equation*}
		\lim_{K \rightarrow \infty}\prod_{k=0}^{K-1}\bm{\alpha}^{(k)} = 0,
	\end{equation*}
	and
	\begin{equation*}
		\lim_{K \rightarrow \infty}\left(\prod_{k=0}^{K-1}\bm{\alpha}^{(k)}\right)\left(\prod_{k=0}^{K-1}\tilde{\mM}^{(k)}\right) = 0.
	\end{equation*}
	Therefore,
	\begin{equation}\label{eq:eq24}
		\lim_{K \rightarrow \infty}\mF^{(K)} = \lim_{K \rightarrow \infty}\left(\sum_{k=1}^{K-1}\left(\prod_{l=K-k}^{K-1}\bm{\alpha}^{(l)}\tilde{\mM}^{(l)}\right)\bm{\beta}^{(K-1-k)}\mX + \bm{\beta}^{(K-1)}\mX\right).
	\end{equation}
	By \cref{eq:eq6,eq:eq12}, we have
	\begin{align}
		\Delta_pf(i) = {} & \sum_{j=1}^N\frac{\emW_{i,j}}{\emD_{i,i}}\|(\nabla f)([j, i])\|^{p-2}f(i) - \sum_{j=1}^N\frac{\emW_{i,j}}{\sqrt{\emD_{i,i}\emD_{j,j}}}\|(\nabla f)([j, i])\|^{p-2}f(j) \notag \\
		= {} & \sum_{j=1}^N\frac{\emM_{i,j}}{\emD_{i,i}}f(i) - \sum_{j=1}^N\frac{\emM_{i,j}}{\sqrt{\emD_{i,i}\emD_{j,j}}}f(j). \label{eq:eq25}
	\end{align}
	By \cref{eq:eq13}, we have
	\begin{equation}\label{eq:eq26}
		\sum_{j=1}^N\frac{\emM_{i,j}^{(k)}}{\emD_{i,i}} = \frac{1}{\alpha_{i,i}^{(k)}} - \frac{2\mu}{p}.
	\end{equation}
	\cref{eq:eq25,eq:eq26} show that
	\begin{equation}\label{eq:eq27}
		\Delta_p^{(k)} = \left(\left(\bm{\alpha}^{(k)}\right)^{-1} - \frac{2\mu}{p}\mI_N\right) - \tilde{\mM}^{(k)},
	\end{equation}
	which indicates
	\begin{equation}\label{eq:eq28}
		\bm{\alpha}^{(k)}\tilde{\mM}^{(k)} = \mI_N - \frac{2\mu}{p}\bm{\alpha}^{(k)} - \bm{\alpha}^{(k)}\Delta_p^{(k)}.
	\end{equation}
	\cref{eq:eq28} shows that $\bm{\alpha}^{(k)}\tilde{\mM}^{(k)}$ is linear w.r.t $\Delta_p$ and therefore can be expressed by a linear combination in terms of $\Delta_p$:
	\begin{equation}\label{eq:eq29}
		\bm{\alpha}^{(k)}\tilde{\mM}^{(k)} = \bm{\theta}'^{(k)}\Delta_p,
	\end{equation}
	where $\bm{\theta}' = \mathrm{diag}(\theta_0', \theta_1', \dots, \theta_{N-1}')$ are the parameters. Therefore, we have
	\begin{align*}
		\lim_{K \rightarrow \infty}\mF^{(K)} = {} & \lim_{K \rightarrow \infty}\left(\sum_{k=1}^{K-1}\left(\prod_{l=K-k}^{K-1}\bm{\alpha}^{(l)}\tilde{\mM}^{(l)}\right)\bm{\beta}^{(K-1-k)}\mX + \bm{\beta}^{(K-1)}\mX\right) \notag \\
		= {} & \lim_{K \rightarrow \infty}\left(\sum_{k=1}^{K-1}\left(\prod_{l=K-k}^{K-1}\bm{\theta}'^{(l)}\Delta_p\right)\bm{\beta}^{(K-1-k)}\mX + \bm{\beta}^{(K-1)}\mX\right) \notag \\
		= {} & \lim_{K \rightarrow \infty}\left(\sum_{k=1}^{K-1}\bm{\beta}^{(K-1-k)}\left(\prod_{l=K-k}^{K-1}\bm{\theta}'^{(l)}\right)\Delta_p^k\mX + \bm{\beta}^{(K-1)}\mX\right) \notag \\
		= {} & \lim_{K \rightarrow \infty}\sum_{k=0}^{K-1}\bm{\theta}''^{(k)}\Delta_p^k\mX,
	\end{align*}
	where $\bm{\theta}''^{(k)} = \mathrm{diag}(\theta_1''^{(k)}, \theta_2''^{(k)}, \dots, \theta_N''^{(k)})$ defined as $\bm{\theta}''^{(0)} = \bm{\beta}^{(K-1)}$ and
	\begin{equation*}
		\theta_i''^{(k)} = \beta_{i,i}^{(K-1-k)}\prod_{l=K-k}^{K-1}\theta_i'^{(l)}, \text{ for } k = 1, 2, \dots, K-1.
	\end{equation*}
	Let $\bm{\theta} = (\theta_0, \theta_1, \dots, \theta_{K-1})$ defined as $\theta_k = \sum_{i=1}^N\theta_i''^{(k)}$ for all $k = 0, 1, \dots, K-1$, then
	\begin{align*}
		\lim_{K \rightarrow \infty}\mF^{(K)} = {} & \lim_{K \rightarrow \infty}\left(\sum_{k=1}^{K-1}\theta_k\Delta_p^k\mX + \theta_0\mX\right) \\
		= {} & \lim_{K \rightarrow \infty}\sum_{k=0}^{K-1}\theta_k\Delta_p^k\mX.
	\end{align*}
	Therefore complete the proof.
\end{proof}

\subsection{Proof of Theorem 4}\label{App:proof:the3}
\begin{proof}
    The first-order Taylor expansion  with Peano’s form of remainder for $\sigma$ at $\mX_{i,:}^*$ is given by:
	\begin{equation*}
		\sigma(\mF_{j,:}^{(K-1)}) = \sigma(\mX_{i,:}^*) + \frac{\partial \sigma(\mX_{i,:}^*)}{\partial\mX}\left(\mF_{j,:}^{(K-1)} - \mX_{i,:}^*\right)^\top + o(\|\mF_{j,:}^{(K-1)} - \mX_{i,:}^*\|).
	\end{equation*}
	Note that in general the output non-linear layer $\sigma(\cdot)$ is simple. Here we assume that it can be well approximated by the first-order Taylor expansion and we can ignore the Peano’s form of remainder. For all $i = 1, \dots, N$, $\emD_{i,i} = \emD_{j,j} = d$, we have $\alpha_{i,i}\sum_{j=1}^N\emD_{i,i}^{-1/2}\emM_{i,j}^{(K-1)}\emD_{j,j}^{-1/2} + \beta_{i,i}^{(K-1)} = 1$. Then
    \begin{align*}
		{} & \left|\evy_i - \tilde{\evy}_i^{(K)}\right| \\
		= {} & \left|\evy_i - \alpha_{i,i}^{(K-1)}\sum_{j=1}^N\frac{\emM_{i,j}^{(K-1)}}{\sqrt{\emD_{i,i}\emD_{j,j}}}\sigma\left(\mF_{j,:}^{K-1}\right) - \beta_{i,i}^{(K-1)}\sigma\left(\mX_{i,:}\right)\right| \\
		= {} & \left|\evy_i - \beta_{i,i}^{(K-1)}\sigma\left(\mX_{i,:}\right) - \alpha_{i,i}^{(K-1)}\sum_{j=1}^N\frac{\emM_{i,j}^{(K-1)}}{\sqrt{\emD_{i,i}\emD_{j,j}}}\left(\sigma\left(\mX_{i,:}^*\right) + \frac{\partial\sigma\left(\mX_{i,:}^*\right)}{\partial\mX}\left(\mF_{j,:}^{(K-1)} - \mX_{i,:}^*\right)^\top\right)\right| \\
		= {} & \left|\evy_i - \alpha_{i,i}^{(K-1)}\sum_{j=1}^N\frac{\emM_{i,j}^{(K-1)}}{\sqrt{\emD_{i,i}\emD_{j,j}}}\evy_i - \beta_{i,i}^{(K-1)}\sigma(\mX_{i,:}) - \alpha_{i,i}^{(K-1)}\sum_{j=1}^N\frac{\emM_{i,j}^{(K-1)}}{\sqrt{\emD_{i,i}\emD_{j,j}}}\left(\frac{\partial \sigma(\mX_{i,:}^*)}{\partial\mX}\left(\mF_{j,:}^{(K-1)} - \mX_{i,:}^*\right)^\top\right)\right| \\
		= {} & \left|\beta_{i,i}^{(K-1)}(\evy_i - \sigma(\mX_{i,:})) - \alpha_{i,i}^{(K-1)}\sum_{j=1}^N\frac{\emM_{i,j}^{(K-1)}}{\sqrt{\emD_{i,i}\emD_{j,j}}}\left(\frac{\partial \sigma(\mX_{i,:}^*)}{\partial\mX}\left(\mF_{j,:}^{(K-1)} - \mX_{i,:} - \bm{\epsilon}_{i,:}\right)^\top\right)\right| \\
		\leq {} & \beta_{i,i}^{(K-1)}\left|\evy_i - \sigma(\mX_{i,:})\right| + \alpha_{i,i}^{(K-1)}\left|\sum_{j=1}^N\frac{\emM_{i,j}^{(K-1)}}{\sqrt{\emD_{i,i}\emD_{j,j}}}\frac{\partial \sigma(\mX_{i,:}^*)}{\partial\mX}\left(\mF_{j,:}^{(K-1)} - \mX_{i,:}\right)^\top\right| + (1 - \beta_{i,i}^{(K-1)})\left|\frac{\partial \sigma(\mX_{i,:}^*)}{\partial\mX}\bm{\epsilon}_{i,:}^\top\right| \\
		\leq {} & \beta_{i,i}^{(K-1)}\left|\evy_i - \sigma(\mX_{i,:})\right| + \alpha_{i,i}^{(K-1)}\left\|\frac{\partial \sigma(\mX_{i,:}^*)}{\partial\mX}\right\|\left\|\sum_{j=1}^N\frac{\emM_{i,j}^{(K-1)}}{\sqrt{\emD_{i,i}\emD_{j,j}}}\left(\mF_{j,:}^{(K-1)} - \mX_{i,:}\right)\right\| + (1 - \beta_{i,i}^{(K-1)})\left\|\frac{\partial \sigma(\mX_{i,:}^*)}{\partial\mX}\right\|\left\|\bm{\epsilon}_{i,:}\right\| \\
		\leq {} & \beta_{i,i}^{(K-1)}\left|\evy_i - \sigma(\mX_{i,:})\right| + \alpha_{i,i}^{(K-1)}L\left\|\sum_{j=1}^N\frac{\emM_{i,j}^{(K-1)}}{d}\left(\mF_{j,:}^{(K-1)} - \mX_{i,:}\right)\right\| + (1 - \beta_{i,i}^{(K-1)})L\left\|\bm{\epsilon}_{i,:}\right\| \\
		= {} & \beta_{i,i}^{(K-1)}\left|\evy_i - \sigma(\mX_{i,:})\right| + \alpha_{i,i}^{(K-1)}L\left\|\sum_{j=1}^N\frac{\emM_{i,j}^{(K-1)}}{d}\left(\mF_{j,:}^{(K-1)} - \mF_{i,:}^{(K-1)} + \mF_{i,:}^{(K-1)} - \mX_{i,:}\right)\right\| \\ {} & \qquad\qquad\qquad\qquad\qquad + (1 - \beta_{i,i}^{(K-1)})L\left\|\bm{\epsilon}_{i,:}\right\| \\
		= {} & \beta_{i,i}^{(K-1)}\left|\evy_i - \sigma(\mX_{i,:})\right| + \alpha_{i,i}^{(K-1)}L\left\|\Delta_p\mF_{i,:}^{(K-1)} + \sum_{j=1}^N\frac{\emM_{i,j}^{(K-1)}}{d} \cdot \sum_{j=1}^N\frac{\emM_{i,j}^{(K-2)}}{d}\left(\mF_{j,:}^{(K-2)} - \mX_{i,:}\right)\right\| \\ {} & \qquad\qquad\qquad\qquad\qquad + (1 - \beta_{i,i}^{(K-1)})L\left\|\bm{\epsilon}_{i,:}\right\| \\
		= {} & \beta_{i,i}^{(K-1)}\left|\evy_i - \sigma(\mX_{i,:})\right| + \alpha_{i,i}^{(K-1)}L\left\|\Delta_p^{(K-1)}\mF_{i,:}^{(K-1)} + \sum_{k=0}^{K-2}\prod_{l=k}^{K-2}\left(\sum_{j=1}^N\frac{\emM_{i,j}^{(l)}}{d}\right)\Delta_p^{(k)}\mX_{i,:}\right\| + (1 -\beta_{i,i}^{(K-1)})L\left\|\bm{\epsilon}_{i,:}\right\| \\
	    = {} & \beta_{i,i}^{(K-1)}\left|\evy_i - \sigma(\mX_{i,:})\right| + \alpha_{i,i}^{(K-1)}L\left\|\Delta_p^{(K-1)}\mF_{i,:}^{(K-1)} + \sum_{k=0}^{K-2}\prod_{l=k}^{K-2}\left(\sum_{j=1}^N\frac{\emM_{i,j}^{(l)}}{d}\right)\Delta_p^{(k)}\mX_{i,:}\right\| + (1 -\beta_{i,i}^{(K-1)})L\left\|\bm{\epsilon}_{i,:}\right\|
	\end{align*}
	Therefore,
	\begin{align*}
		\frac{1}{N}\sum_{i=1}^N\left|\evy_i - \tilde{\evy}_i\right| \leq {} & \frac{1}{N}\sum_{i=1}^N\beta_{i,i}^{(K-1)}\left|\evy_i - \sigma(\mX_{i,:})\right| \\ {} & \qquad + \frac{L}{N}\sum_{i=1}^N\alpha_{i,i}^{(K-1)}\left\|\Delta_p^{(K-1)}\mF_{i,:}^{(K-1)} + \sum_{k=0}^{K-2}\prod_{l=k}^{K-2}\left(\sum_{j=1}^N\frac{\emM_{i,j}^{(l)}}{d}\right)\Delta_p^{(k)}\mX_{i,:}\right\| \\ {} & \qquad + \frac{L}{N}\sum_{i=1}^N(1 -\beta_{i,i}^{(K-1)})\left\|\bm{\epsilon}_{i,:}\right\|
	\end{align*}
\end{proof}

\subsection{Proof of Theorem 6}\label{App:proof:the5}
\begin{proof}
	Note that
	\begin{equation*}
		\phi_p(\vu)^\top\vu = \sum_{i=1}^N\phi_p(\evu_i)\evu_i = \sum_{i=1}^N\|\evu_i\|^{p-2}\evu_i^2 = \sum_{i=1}^N\|\evu_i\|^p = \sum_{i=1}^N|\evu_i|^p = \|\vu\|_p^p = 1,
	\end{equation*}
	then we have
	\begin{equation*}
		\Delta_p\mU = \Phi_p(\mU)\mLambda = \Phi_p(\mU)\mLambda\Phi(\mU)^\top\mU.
	\end{equation*}
	Therefore, $\Delta_p = \Phi_p(\mU)\mLambda\Phi_p(\mU)^\top$.
	
	When $p = 2$, by $\Phi_2(\mU) = \mU$, we get $\Delta_2 = \Phi_2(\mU)\mLambda\Phi_2(\mU)^\top = \mU\mLambda\mU^\top$.
\end{proof}

\subsection{Proof of Theorem 7}\label{App:proof:the7}
\begin{proof}
	By the definition of graph $p$-Laplacian, we have for all $i = 1, 2, \dots, N$,
	\begin{equation*}
		\left(\Delta_p\vu\right)_i = \sum_{j=1}^N\sqrt{\frac{\emW_{i,j}}{\emD_{i,i}}}\left\|\sqrt{\frac{\emW_{i,j}}{\emD_{i,i}}}\evu_i - \sqrt{\frac{\emW_{i,j}}{\emD_{j,j}}}\evu_j\right\|^{p-2}\left(\sqrt{\frac{\emW_{i,j}}{\emD_{i,i}}}\evu_i - \sqrt{\frac{\emW_{i,j}}{\emD_{j,j}}}\evu_j\right) = \lambda\phi_p(\evu_i).
	\end{equation*}
	Then, for all $i = 1, 2, \dots, N$,
	\begin{equation*}
		\begin{aligned}
			\lambda = {} & \frac{1}{\phi_p(\evu_i)}\sum_{j=1}^N\sqrt{\frac{\emW_{i,j}}{\emD_{i,i}}}\left\|\sqrt{\frac{\emW_{i,j}}{\emD_{i,i}}}\evu_i - \sqrt{\frac{\emW_{i,j}}{\emD_{j,j}}}\evu_j\right\|^{p-2}\left(\sqrt{\frac{\emW_{i,j}}{\emD_{i,i}}}\evu_i - \sqrt{\frac{\emW_{i,j}}{\emD_{j,j}}}\evu_j\right) \\
			= {} & \frac{1}{\|\evu_i\|^{p-2}\evu_i}\sum_{j=1}^N\sqrt{\frac{\emW_{i,j}}{\emD_{i,i}}}\left\|\sqrt{\frac{\emW_{i,j}}{\emD_{i,i}}}\evu_i - \sqrt{\frac{\emW_{i,j}}{\emD_{j,j}}}\evu_j\right\|^{p-2}\left(\sqrt{\frac{\emW_{i,j}}{\emD_{i,i}}}\evu_i - \sqrt{\frac{\emW_{i,j}}{\emD_{j,j}}}\evu_j\right) \\
			= {} & \sum_{j=1}^N\sqrt{\frac{\emW_{i,j}}{\emD_{i,i}}}\frac{\left\|\sqrt{\frac{\emW_{i,j}}{\emD_{i,i}}}\evu_i - \sqrt{\frac{\emW_{i,j}}{\emD_{j,j}}}\evu_j\right\|^{p-2}}{\|\evu_i\|^{p-2}}\left(\sqrt{\frac{\emW_{i,j}}{\emD_{i,i}}} - \sqrt{\frac{\emW_{i,j}}{\emD_{j,j}}}\frac{\evu_j}{\evu_i}\right) \\
			= {} & \sum_{j=1}^N\left(\frac{\emW_{i,j}}{\emD_{i,i}} - \frac{\emW_{i,j}}{\sqrt{\emD_{i,i}\emD_{j,j}}}\frac{\evu_j}{\evu_i}\right)\left(\frac{\left\|\sqrt{\frac{\emW_{i,j}}{\emD_{i,i}}}\evu_i - \sqrt{\frac{\emW_{i,j}}{\emD_{j,j}}}\evu_j\right\|}{\|\evu_i\|}\right)^{p-2} \\
			= {} & \sum_{j=1}^N\left(\frac{\emW_{i,j}}{\emD_{i,i}} - \frac{\emW_{i,j}}{\sqrt{\emD_{i,i}\emD_{j,j}}}\frac{\evu_j}{\evu_i}\right)\left\|\frac{\sqrt{\frac{\emW_{i,j}}{\emD_{i,i}}}\evu_i - \sqrt{\frac{\emW_{i,j}}{\emD_{j,j}}}\evu_j}{\evu_i}\right\|^{p-2} \\
			= {} & \sum_{j=1}^N\left(\frac{\emW_{i,j}}{\emD_{i,i}} - \frac{\emW_{i,j}}{\sqrt{\emD_{i,i}\emD_{j,j}}}\frac{\evu_j}{\evu_i}\right)\left\|\sqrt{\frac{\emW_{i,j}}{\emD_{i,i}}} - \sqrt{\frac{\emW_{i,j}}{\emD_{j,j}}}\frac{\evu_j}{\evu_i}\right\|^{p-2}			
		\end{aligned}
	\end{equation*}
	Let $l=\argmax\{\|u_i\|\}_{i=1,2,\dots,N}$, the above equation holds for all $i = 1, 2, \dots, N$, then
	\begin{equation*}
		\begin{aligned}
			\lambda = {} & \sum_{j=1}^N\left(\frac{\emW_{l,j}}{\emD_{l,l}} - \frac{\emW_{l,j}}{\sqrt{\emD_{l,l}\emD_{j,j}}}\frac{\evu_j}{\evu_l}\right)\left\|\sqrt{\frac{\emW_{l,j}}{\emD_{l,l}}} - \sqrt{\frac{\emW_{l,j}}{\emD_{j,j}}}\frac{\evu_j}{\evu_l}\right\|^{p-2}	\\
			\geq {} & \sum_{j=1}^N\left(\frac{\emW_{l,j}}{\emD_{l,l}} - \frac{\emW_{l,j}}{\sqrt{\emD_{l,l}\emD_{j,j}}}\left|\frac{\evu_j}{\evu_l}\right|\right)\left\|\sqrt{\frac{\emW_{l,j}}{\emD_{l,l}}} - \sqrt{\frac{\emW_{l,j}}{\emD_{j,j}}}\frac{\evu_j}{\evu_l}\right\|^{p-2} \\
			\geq {} & \sum_{j=1}^N\left(\frac{\emW_{l,j}}{\emD_{l,l}} - \frac{\emW_{l,j}}{\sqrt{\emD_{l,l}\emD_{j,j}}}\right)\left\|\sqrt{\frac{\emW_{l,j}}{\emD_{l,l}}} - \sqrt{\frac{\emW_{l,j}}{\emD_{j,j}}}\frac{\evu_j}{\evu_l}\right\|^{p-2} \\
			\geq {} & 0.
		\end{aligned}
	\end{equation*}
	When $p=1$, 
	\begin{equation*}
		\begin{aligned}
			\lambda = {} & \sum_{j=1}^N\left(\frac{W_{i,j}}{D_{i,i}} - \frac{W_{i,j}}{\sqrt{D_{i,i}D_{j,j}}}\frac{u_j}{u_i}\right)\left\|\sqrt{\frac{W_{i,j}}{D_{i,i}}} - \sqrt{\frac{W_{i,j}}{D_{j,j}}}\frac{u_j}{u_i}\right\|^{-1} \\
			\leq {} & \sum_{j=1}^N\sqrt{\frac{W_{i,j}}{D_{i,i}}} \leq \sqrt{N_i\sum_{j=1}^N\frac{W_{i,j}}{D_{i,i}}} = \sqrt{N_i},
		\end{aligned}
	\end{equation*}
	where the last inequality holds by using the Cauchy-Schwarz inequality. The above inequality holds for all $i = 1, 2, \dots, N$, therefore,
	\begin{equation*}
		\lambda \leq \sum_{j=1}^N\sqrt{\frac{W_{i,j}}{D_{i,i}}} \leq \sqrt{N_{min}}.
	\end{equation*}
	When $p > 1$, we have for $i = 1, 2, \dots, N$,
	\begin{equation*}
		\begin{aligned}
			\lambda = {} & \sum_{j=1}^N\left(\frac{\emW_{i,j}}{\emD_{i,i}} - \frac{\emW_{i,j}}{\sqrt{\emD_{i,i}\emD_{j,j}}}\frac{\evu_j}{u_i}\right)\left\|\sqrt{\frac{\emW_{i,j}}{\emD_{i,i}}} - \sqrt{\frac{\emW_{i,j}}{\emD_{j,j}}}\frac{\evu_j}{u_i}\right\|^{p-2} \\
			\leq {} & \sum_{j=1}^N\sqrt{\frac{\emW_{i,j}}{\emD_{i,i}}}\left\|\sqrt{\frac{\emW_{i,j}}{\emD_{i,i}}} - \sqrt{\frac{\emW_{i,j}}{\emD_{j,j}}}\frac{\evu_j}{u_i}\right\|^{p-1} \\
			\leq {} & \sum_{j=1}^N\sqrt{\frac{\emW_{i,j}}{\emD_{i,i}}}\left\|\sqrt{\frac{\emW_{i,j}}{\emD_{i,i}}} + \sqrt{\frac{\emW_{i,j}}{\emD_{j,j}}}\left|\frac{\evu_j}{u_i}\right|\right\|^{p-1} \\
			= {} & \sum_{j=1}^N\left(\sqrt{\frac{\emW_{i,j}}{\emD_{i,i}}}\right)^p\left\|1 + \sqrt{\frac{\emD_{i,i}}{\emD_{j,j}}}\left|\frac{\evu_j}{u_i}\right|\right\|^{p-1}.
		\end{aligned}
	\end{equation*}
	Without loss of generality, let $k = \argmax(\{|\evu_i| / \sqrt{\emD_{i,i}}\}_{i=1, 2, \dots, N})$. Because the above inequality holds for all $i = 1, 2, \dots, N$, then we have
	\begin{equation*}
		\begin{aligned}
			\lambda \leq {} & \sum_{j=1}^N\left(\sqrt{\frac{\emW_{k,j}}{\emD_{k,k}}}\right)^p\left(1 + \sqrt{\frac{\emD_{k,k}}{\emD_{j,j}}}\left|\frac{\evu_j}{u_k}\right|\right)^{p-1} \\
			\leq {} & 2^{p-1}\sum_{j=1}^N\left(\sqrt{\frac{\emW_{k,j}}{\emD_{k,k}}}\right)^p.
		\end{aligned}
	\end{equation*}
	For $p \geq 2$,
	\begin{equation*}
		\lambda \leq 2^{p-1}\sum_{j=1}^N\left(\sqrt{\frac{\emW_{k,j}}{\emD_{k,k}}}\right)^p \leq 2^{p-1}\sum_{j=1}^N\left(\sqrt{\frac{\emW_{k,j}}{\emD_{k,k}}}\right)^2 = 2^{p-1}.
	\end{equation*}
	For $1 < p < 2$,
	\begin{equation*}
		\lambda \leq 2^{p-1}\sum_{j=1}^N\left(\sqrt{\frac{\emW_{k,j}}{\emD_{k,k}}}\right)^p \leq 2^{p-1}\sum_{j=1}^N\sqrt{\frac{\emW_{k,j}}{\emD_{k,k}}} \leq 2^{p-1}\sqrt{N_k\sum_{j=1}^N\frac{W_{k,j}}{D_{k,k}}} = 2^{p-1}\sqrt{N_k}.
	\end{equation*}
\end{proof}

\subsection{Proof of Proposition 1}\label{App:prop1}

\begin{proof}
    We proof \cref{the:prop1} based on the bounds of $p$-eigenvalues as demonstrated in \cref{the:the7}.
    
	By \cref{eq:eq6} and \cref{eq:eq12}, we have
	\begin{align}
		\Delta_pf(i) = {} & \sum_{j=1}^N\frac{\emW_{i,j}}{\emD_{i,i}}\|(\nabla f)([j, i])\|^{p-2}f(i) - \sum_{j=1}^N\frac{\emW_{i,j}}{\sqrt{\emD_{i,i}\emD_{j,j}}}\|(\nabla f)([j, i])\|^{p-2}f(j) \notag \\
		= {} & \sum_{j=1}^N\frac{\emM_{i,j}}{\emD_{i,i}}f(i) - \sum_{j=1}^N\frac{\emM_{i,j}}{\sqrt{\emD_{i,i}\emD_{j,j}}}f(j). \label{eq:eq30}
	\end{align}
	By \cref{eq:eq13}, we have
	\begin{equation}\label{eq:eq31}
		\sum_{j=1}^N\frac{\emM_{i,j}^{(k)}}{\emD_{i,i}} = \frac{1}{\alpha_{i,i}^{(k)}} - \frac{2\mu}{p}.
	\end{equation}
	\cref{eq:eq30,eq:eq31} show that
	\begin{equation}\label{eq:eq32}
		\Delta_p^{(k)} = \left(\left(\bm{\alpha}^{(k)}\right)^{-1} - \frac{2\mu}{p}\mI_N\right) - \mD^{-1/2}\rmM^{(k)}\mD^{-1/2},
	\end{equation}
	which indicates
	\begin{equation}\label{eq:eq33}
		\bm{\alpha}^{(k)}\mD^{-1/2}\rmM^{(k)}\mD^{-1/2} = \mI_N - \frac{2\mu}{p}\bm{\alpha}^{(k)} - \bm{\alpha}^{(k)}\Delta_p^{(k)}.
	\end{equation}
	For $i = 1, 2, \dots, N$, let $\tilde{\bm{\alpha}} := (\tilde{\alpha}_1, \dots, \tilde{\alpha}_N)$, $\tilde{\alpha}_i := 1 / \sum_{j=1}^N\frac{\emM_{i,j}}{\emD_{i,i}}$, then
	\begin{align}
		\alpha_{i,i}\sum_{j=1}^N\frac{\emM_{i,j}}{\sqrt{\emD_{i,i}\emD_{j,j}}} = {} & (1 - \frac{2\mu}{p}\alpha_{i,i}) - \alpha_{i,i}\lambda_i \notag \\
		= {} & \frac{\sum_{j=1}^N\frac{\emM_{i,j}}{\emD_{i,i}}}{\sum_{j=1}^N\frac{\emM_{i,j}}{\emD_{i,i}} + \frac{2\mu}{p}}\left(1 - \frac{1}{\sum_{j=1}^N\frac{\emM_{i,j}}{\emD_{i,i}}}\lambda_i\right) \notag \\
		= {} & \frac{1}{1 + \frac{2\mu\tilde{\alpha}_i}{p}}\left(1 - \tilde{\alpha}_i\lambda_i\right), \label{eq:eq34}
	\end{align}
	Recall the \cref{eq:eq12} that
	\begin{equation*}
		\emM_{i,j} = \emW_{i,j}\left\|\sqrt{\frac{\emW_{i,j}}{\emD_{i,i}}}\mF_{i,:} - \sqrt{\frac{\emW_{i,j}}{\emD_{j,j}}}\mF_{j,:}\right\|^{p-2} = \emW_{i,j}\|(\nabla f)([i, j])\|^{p-2},
	\end{equation*}
	\begin{enumerate}
		\item When $p = 2$, for all $i = 1, \dots, N$, $\tilde{\alpha}_i = 1$ and $0 \leq \lambda_{i-1} \leq 2$, $g_2(\lambda_{i-1})$ works as low-high-pass filters.
		
		\item When $p > 2$, by \cref{the:the7} we have for all $i = 1, \dots, N$, $0 \leq \lambda_{i-1} \leq 2^{p-1}$. If $0 \leq \tilde{\alpha}_i \leq 2^{1-p}$, then $0 \leq 1 - \tilde{\alpha}_i\lambda_i \leq 1$, which indicates that $g_p(\lambda_{i-1})$ works as a low-pass filter; If $\tilde{\alpha}_i > 2^{1-p}$, then $g_p(\lambda_{i-1})$ works as low-high-pass filters. Since
		\begin{align*}
			\sum_{j=1}^N\frac{\emM_{i,j}}{\emD_{i,i}} = {} & \sum_{j=1}^N\frac{\emW_{i,j}\|(\nabla f)([i, j])\|^{p-2}}{\emD_{i,i}} \\
			\leq {} & \sqrt{\sum_{j=1}^N\left(\frac{\emW_{i,j}}{\emD_{i,i}}\right)^2\sum_{j=1}^N\|(\nabla f)([i, j])\|^{2(p-2)}} \\
			\leq {} & \sqrt{\sum_{j=1}^N\|(\nabla f)([i, j])\|^{2(p-2)}} \\
			\leq {} & \|\nabla f(i)\|^{p-2},
		\end{align*}
		which indicates that $\tilde{\alpha}_i \geq \|\nabla f(i)\|^{2-p}$. $0 \leq \tilde{\alpha}_i \leq 2^{1-p}$ directly implies that $0 \leq \|\nabla f(i)\|^{2-p} \leq 2^{1-p}$, i.e. $\|\nabla f(i)\| \geq 2^{(p-1)/(p-2)}$ and when $\|\nabla f(i)\|^{2-p} \geq 2^{1-p}$, i.e. $\|\nabla f(i)\| \leq 2^{(p-1)/(p-2)}$, $\tilde{\alpha}_i \geq 2^{1-p}$ always holds. Therefore, if $\|\nabla f(i)\| \leq 2^{(p-1)/(p-2)}$, $g_p(\lambda_{i-1})$ works as low-high-pass filters on node $i$; If $g_p(\lambda_{i-1})$ works as a low-pass filter, $\|\nabla f(i)\| \geq 2^{(p-1)/(p-2)}$.
		
		\item When $1 \leq p < 2$, by \cref{the:the7} we have for all $i = 1, \dots, N$, $0 \leq \lambda_{i-1} \leq 2^{p-1}\sqrt{N_k}$. If $0 \leq \tilde{\alpha}_i \leq 2^{1-p}/\sqrt{N_k}$, $0 \leq 1 - \tilde{\alpha}_i\lambda_i \leq 1$, which indicates that $g_p(\lambda_{i-1})$ work as low-pass filters; If $\tilde{\alpha}_i \geq 2^{1-p}/\sqrt{N_k}$, $g_p(\lambda_{i-1})$ work as low-high-pass filters. By
		\begin{align*}
			{} & \sum_{j=1}^N\frac{\emW_{i,j}}{\emD_{i,i}}\frac{1}{\left\|(\nabla f)([i, j])\right\|^{p-2}} \left/ \frac{1}{\sum_{j=1}^N\frac{\emW_{i,j}}{\emD_{i,i}}\left\|(\nabla f)([i, j])\right\|^{p-2}}\right. \\
			= {} & \sum_{j=1}^N\frac{\emW_{i,j}}{\emD_{i,i}}\frac{1}{\left\|(\nabla f)([i, j])\right\|^{p-2}} \cdot \sum_{j=1}^N\frac{\emW_{i,j}}{\emD_{i,i}}\left\|(\nabla f)([i, j])\right\|^{p-2} \\
			\geq {} & \sum_{j=1}^N\frac{\emW_{i,j}}{\emD_{i,i}}\left(\frac{1}{\left\|(\nabla f)([i, j])\right\|^{p-2}} \cdot \left\|(\nabla f)([i, j])\right\|^{p-2}\right) \\
			= {} & 1,
		\end{align*}
		we have
		\begin{align*}
			\tilde{\alpha}_i = {} & \frac{1}{\sum_{i=1}^N\frac{\emM_{i,j}}{\emD_{i,i}}} = \frac{1}{\sum_{j=1}^N\frac{\emW_{i,j}}{\emD_{i,i}}\left\|(\nabla f)([i, j])\right\|^{p-2}} \\
			\leq {} & \sum_{j=1}^N\frac{\emW_{i,j}}{\emD_{i,i}}\frac{1}{\left\|(\nabla f)([i, j])\right\|^{p-2}} \\
			= {} & \sum_{j=1}^N\frac{\emW_{i,j}}{\emD_{i,i}}\left\|(\nabla f)([i, j])\right\|^{2-p} \\
			\leq {} & \|\nabla f(i)\|^{2-p}.
		\end{align*}
		$\tilde{\alpha}_i \geq 2^{1-p}/\sqrt{N_k}$ directly implies that $\|\nabla f(i)\|^{2-p} \geq 2^{1-p}/\sqrt{N_k}$, i.e. $\|\nabla f(i)\| \geq 2(2\sqrt{N_k})^{1/(p-2)}$ and when $0 \leq \|\nabla f(i)\|^{2-p} \leq 2^{1-p}/\sqrt{N_k}$, i.e. $0 \leq \|\nabla f(i)\| \leq 2(2\sqrt{N_k})^{1/(p-2)}$, $0 \leq \tilde{\alpha}_i \leq 2^{1-p}/\sqrt{N_k}$ always holds. Therefore, if $0 \leq \|\nabla f(i)\| \leq 2(2\sqrt{N_k})^{1/(p-2)}$, $g_p(\lambda_{i-1})$ work as low-pass filters; If $g_p(\lambda_{i-1})$ work as low-high-pass filters, $\|\nabla f(i)\| \geq 2\left(2\sqrt{N_k}\right)^{1/(p-2)}$. 
		
		Specifically, when $p = 1$, by \cref{the:the7} we have for all $i = 1, \dots, N$, $0 \leq \lambda_{i-1} \leq 2^{p-1}\sqrt{N_{\text{min}}}$. Following the same derivation above we attain if $0 \leq \|\nabla f(i)\| \leq 2(2\sqrt{N_{\text{min}}})^{1/(p-2)}$, $g_p(\lambda_{i-1})$ work as low-pass filters; If $g_p(\lambda_{i-1})$ work as both low-high-pass filters, $\|\nabla f(i)\| \geq 2\left(2\sqrt{N_{\text{min}}}\right)^{1/(p-2)}$. 
	\end{enumerate}
\end{proof}

%% file: sections/10-App_exp.tex
\section{Dataset Statistics and Hyperparameters}\label{App:exp_setup}

\subsection{Illustration of Graph Gradient and Graph Divergence}\label{App:subsec:example}
\begin{figure}[htp]
	\centering
	\begin{subfigure}{0.48\textwidth}
		\centering
		\includegraphics[width=\textwidth]{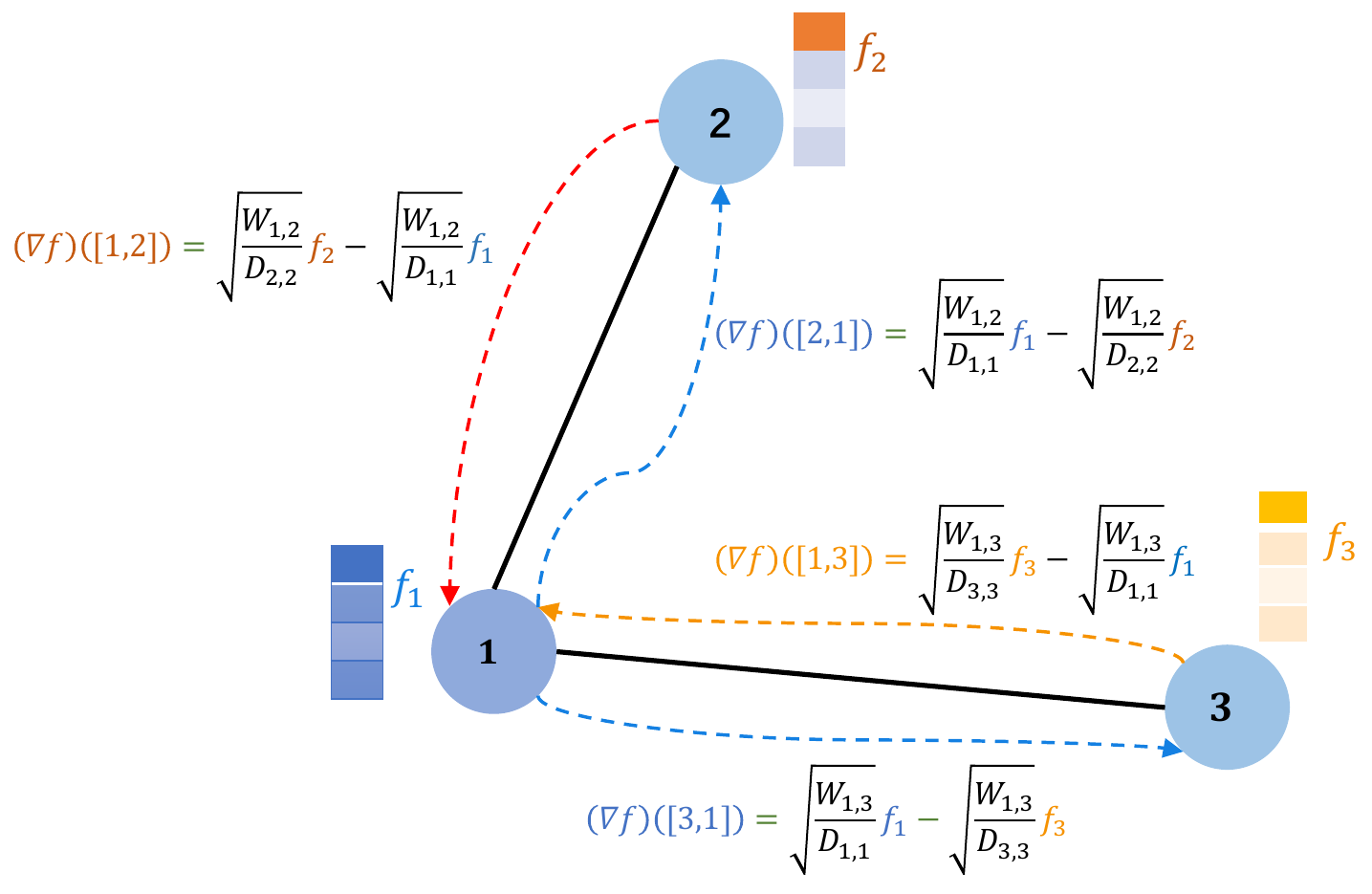}
		\caption{Graph gradient.}
	\end{subfigure}
	\begin{subfigure}{0.48\textwidth}
		\centering
		\includegraphics[width=\textwidth]{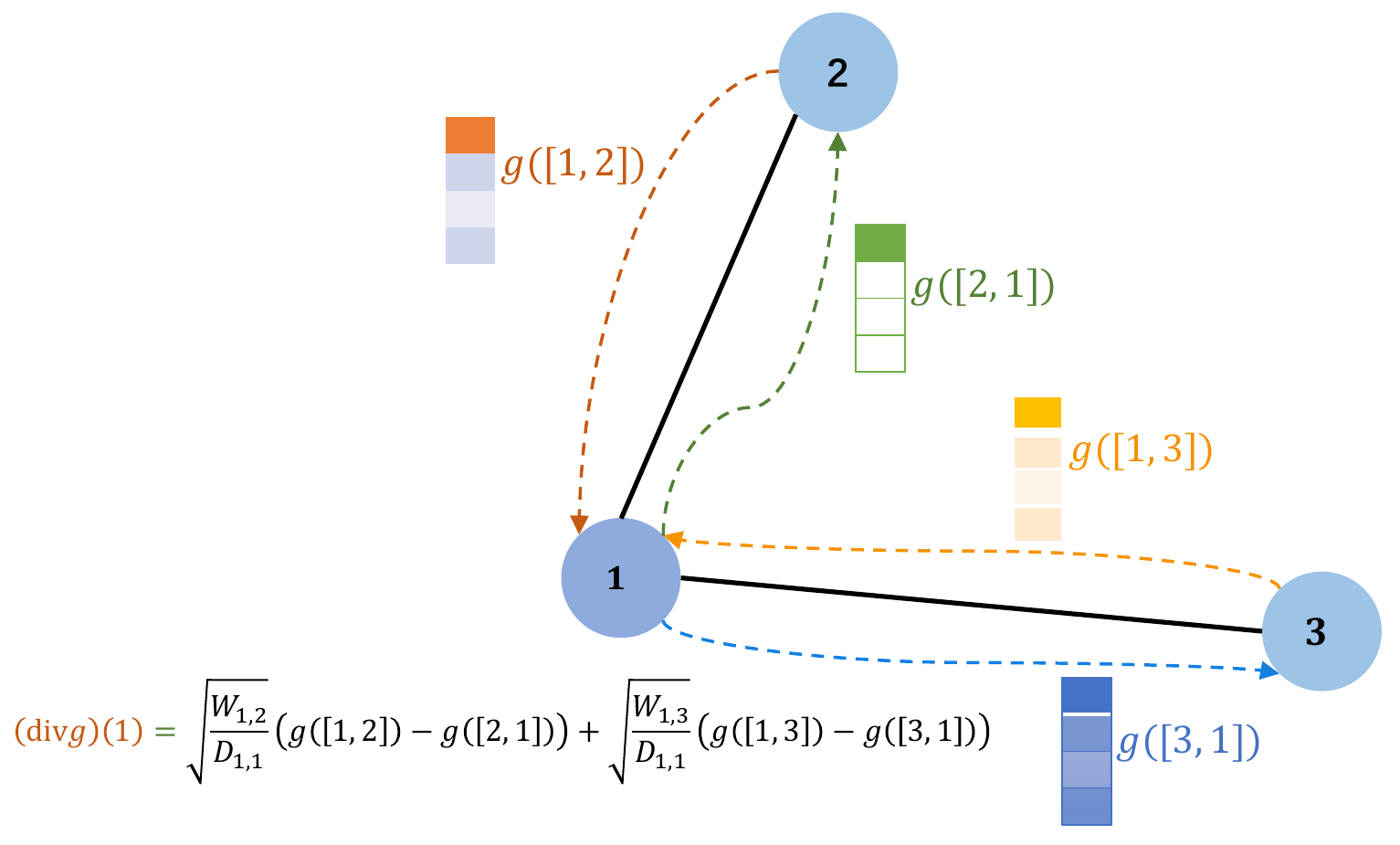}
		\caption{Graph divergence.}
	\end{subfigure}
	\caption{A tiny example of illustration of graph gradient and graph divergence. Best view in colors.}\label{App:fig:example}
\end{figure}

\subsection{Dataset Statistics}\label{App:subsec:datasets}
\cref{App:tab:datasets} summarizes the dataset statistics and the levels of homophily $\gH(\gG)$ of all benchmark datasets. Note that the homophily scores here is different with the scores reported by \citet{DBLP:conf/iclr/ChienP0M21}. There is a bug in their code when computing the homophily scores (doing division with torch integers) which caused their homophily scores to be smaller. Additionally, We directly used the data from Pytorch Geometric library~\citep{DBLP:journals/corr/abs-1903-02428} where they did \textbf{not} transform Chameleon and Squirrel to undirected graphs, which is different from \citet{DBLP:conf/iclr/ChienP0M21} where they did so.
\begin{table}[htp]
	\centering
	\caption{Statistics of datasets.}\label{App:tab:datasets}
	\vspace{-5pt}
	\footnotesize
	\setlength\tabcolsep{2pt}
	\begin{tabular}{ccccccccc}
		\toprule
		Dataset & \#Class & \#Feature & \#Node & \#Edge & Training & Validation & Testing & $\gH(\gG)$ \\
		\midrule
		Cora & 7 & 1433 & 2708 & 5278 & 2.5\% & 2.5\% & 95\% & 0.825 \\
		CiteSeer & 6 & 3703 & 3327 & 4552 & 2.5\% & 2.5\% & 95\% & 0.717 \\
		PubMed & 3 & 500 & 19717 & 44324 & 2.5\% & 2.5\% & 95\% & 0.792 \\
		Computers & 10 & 767 & 13381 & 245778 & 2.5\% & 2.5\% & 95\% & 0.802 \\
		Photo & 8 & 745 & 7487 & 119043 & 2.5\% & 2.5\% & 95\% & 0.849 \\
		CS & 15 & 6805 & 18333 & 81894 & 2.5\% & 2.5\% & 95\% & 0.832 \\
		Physics & 5 & 8415 & 34493 & 247962 & 2.5\% & 2.5\% & 95\% & 0.915 \\
		\midrule
		Chameleon & 5 & 2325 & 2277 & 31371 & 60\% & 20\% & 20\% & 0.247 \\
		Squirrel & 5 & 2089 & 5201 & 198353 & 60\% & 20\% & 20\% & 0.216 \\
		Actor & 5 & 932 & 7600 & 26659 & 60\% & 20\% & 20\% & 0.221 \\
		Wisconsin & 5 & 251 & 499 & 1703 & 60\% & 20\% & 20\% & 0.150 \\
		Texas & 5 & 1703 & 183 & 279 & 60\% & 20\% & 20\% & 0.097 \\
		Cornell & 5 & 1703 & 183 & 277 & 60\% & 20\% & 20\% & 0.386 \\
		\bottomrule
	\end{tabular}
\end{table}

\subsection{Hyperparameter Settings}\label{App:subsec:param}
We set the number of layers as 2, the maximum number of epochs as 1000, the number for early stopping as 200, the weight decay as 0 or 0.0005 for all models. The other hyperparameters for each model are listed as below:
\begin{itemize}
    \item $^{1.0}$GNN, $^{1.5}$GNN, $^{2.0}$GNN, $^{2.5}$GNN:
        \begin{itemize}
            \item Number of hidden units: $16$
            
            \item Learning rate: $\{0.001, 0.01, 0.05\}$
            
            \item Dropout rate: $\{0, 0.5\}$
            
            \item $\mu$: $\{0.01, 0.1, 0.2, 1, 10\}$
            
            \item $K$: $4, 6, 8$
        \end{itemize}
    \item MLP:
        \begin{itemize}
            \item Number of hidden units: $16$
            
            \item Learning rate: $\{0.001, 0.01\}$
            
            \item Dropout rate: $\{0, 0.5\}$
        \end{itemize}
    \item GCN:
        \begin{itemize}
            \item Number of hidden units: $16$
            
            \item Learning rate: $\{0.001, 0.01\}$
            
            \item Dropout rate: $\{0, 0.5\}$
        \end{itemize}
    \item SGC:
        \begin{itemize}
            \item Number of hidden units: $16$
            
            \item Learning rate: $\{0.2, 0.01\}$
            
            \item Dropout rate: $\{0, 0.5\}$
            
            \item $K$: $2$
        \end{itemize}
    \item GAT:
        \begin{itemize}
            \item Number of hidden units: $8$
            
            \item Number of attention heads: $8$
            
            \item Learning rate: $\{0.001, 0.005\}$
            
            \item Dropout rate: $\{0, 0.6\}$
        \end{itemize}
    \item JKNet:
        \begin{itemize}
            \item Number of hidden units: $16$
            
            \item Learning rate: $\{0.001, 0.01\}$
            
            \item Dropout rate: $\{0, 0.5\}$
            
            \item $K$: $10$
            
            \item $\alpha$: $\{0.1, 0.5, 0.7, 1\}$
            
            \item The number of GCN based layers: $2$
            
            \item The layer aggregation: LSTM with $16$ channels and $4$ layers
        \end{itemize}
    \item APPNP:
        \begin{itemize}
            \item Number of hidden units: $16$
            
            \item Learning rate: $\{0.001, 0.01\}$
            
            \item Dropout rate: $\{0, 0.5\}$
            
            \item $K$: $10$
            
            \item $\alpha$: $\{0.1, 0.5, 0.7, 1\}$
        \end{itemize}
    \item GPRGNN:
        \begin{itemize}
            \item Number of hidden units: $16$
            
           \item Learning rate: $\{0.001, 0.01, 0.05\}$
            
            \item Dropout rate: $\{0, 0.5\}$
            
            \item $K$: $10$
            
            \item $\alpha$: $\{0, 0.1, 0.2, 0.5, 0.7, 0.9, 1\}$
            
            \item dprate: $\{0, 0.5, 0.7\}$
        \end{itemize}
\end{itemize}

\newpage

\section{Additional Experiments}\label{App:more_exp}

\subsection{Experimental Results on Homophilic Benchmark Datasets}\label{App:homo_exp}

\textbf{Competitive Performance on Real-World  Homophilic  Datasets.} \cref{App:tab:tab_homo_exp} summarizes the averaged accuracy (the micro-F1 score) and standard deviation of semi-supervised node classification on homophilic benchmark datasets. \cref{App:tab:tab_homo_exp} shows that the performance of \pgnn is very close to APPNP, JKNet, GCN on Cora, CiteSeer, PubMed datasets and slightly outperforms all baselines on Computers, Photo, CS, Physics datasets. Moreover, we observe that \pgnns outperform GPRGNN on all homophilic datasets, which confirms that \pgnns work better under weak supervised information (2.5\% training rate) as discussed in \cref{remk:remk2}. We also see that all GNN models work significantly better than MLP on all homophilic datasets. It illustrates that the graph topological information is  helpful for the label prediction tasks. Notably,  $^{1.0}$GNN is slightly worse than the other \pgnns with larger $p$, which suggests to use $p \approx 2$ for homophilic graphs. Overall, the results of \cref{App:tab:tab_homo_exp} indicates  that \pgnns obtain competitive performance against all baselines on homophilic datasets.

\begin{table}[htp]
	\centering
	\caption{Results on homophilic benchmark datasets. Averaged accuracy (\%) for 100 runs. Best results are outlined in bold and the results within $95\%$ confidence interval of the best results are outlined in underlined bold. OOM denotes out of memory.}\label{App:tab:tab_homo_exp}
	\vspace{-5pt}
	\footnotesize
	\setlength\tabcolsep{2pt}
	\begin{tabular}{cccccccc}
		\toprule
		Method & Cora & CiteSeer & PubMed & Computers & Photo & CS & Physics \\
		\midrule
		MLP & $43.47_{\pm 3.82}$ & $46.95_{\pm 2.15}$ & $78.95_{\pm 0.49}$ & $66.11_{\pm 2.70}$ & $76.44_{\pm 2.83}$ & $86.24_{\pm 1.43}$ & $92.58_{\pm 0.83}$ \\
		GCN & $76.23_{\pm 0.79}$ & $62.43_{\pm 0.81}$ & $83.72_{\pm 0.27}$ & $84.17_{\pm 0.59}$ & $90.46_{\pm 0.48}$ & $90.33_{\pm 0.36}$ & $94.46_{\pm 0.08}$ \\
		SGC & $77.19_{\pm 1.47}$ & $\bm{64.10}_{\pm 1.36}$ & $79.26_{\pm 0.69}$ & $84.32_{\pm 0.59}$ & $89.81_{\pm 0.57}$ & $91.06_{\pm 0.05}$ & \text{OOM} \\
		GAT & $75.62_{\pm 1.01}$ & $61.28_{\pm 1.09}$ & $83.60_{\pm 0.22}$ & $82.72_{\pm 1.29}$ & $90.48_{\pm 0.57}$ & $89.96_{\pm 0.27}$ & $93.96_{\pm 0.21}$ \\
		JKNet & $77.19_{\pm 0.98}$ & $63.32_{\pm 0.95}$ & $82.54_{\pm 0.43}$ & $79.94_{\pm 2.47}$ & $88.29_{\pm 1.64}$ & $89.69_{\pm 0.66}$ & $93.92_{\pm 0.32}$ \\
		APPNP & $\bm{79.58}_{\pm 0.59}$ & $63.02_{\pm 1.10}$ & $\bm{84.80}_{\pm 0.22}$ & $83.32_{\pm 1.11}$ & $90.42_{\pm 0.53}$ & $91.54_{\pm 0.24}$ & $\bm{94.93}_{\pm 0.06}$ \\
		GPRGNN & $76.10_{\pm 1.30}$ & $61.60_{\pm 1.69}$ & $83.16_{\pm 0.84}$ & $82.78_{\pm 1.87}$ & $89.81_{\pm 0.66}$ & $90.59_{\pm 0.38}$ & $\bm{\underline{94.72}}_{\pm 0.16}$ \\
		\midrule
		$^{1.0}$GNN & $77.59_{\pm 0.69}$ & $63.19_{\pm 0.98}$ & $83.21_{\pm 0.30}$ & $84.46_{\pm 0.89}$ & $\bm{\underline{90.69}}_{\pm 0.66}$ & $91.46_{\pm 0.50}$ & $\bm{\underline{94.72}}_{\pm 0.37}$ \\
		$^{1.5}$GNN & $78.86_{\pm 0.75}$ & $\bm{\underline{63.80}}_{\pm 0.79}$ & $83.65_{\pm 0.17}$ & $\bm{85.03}_{\pm 0.90}$ & $\bm{90.91}_{\pm 0.50}$ & $\bm{\underline{92.12}}_{\pm 0.40}$ & $\bm{\underline{94.90}}_{\pm 0.16}$ \\
		$^{2.0}$GNN & $78.93_{\pm 0.60}$ & $\bm{\underline{63.65}}_{\pm 1.08}$ & $84.19_{\pm 0.22}$ & $84.39_{\pm 0.85}$ & $90.40_{\pm 0.63}$ & $\bm{92.28}_{\pm 0.47}$ & $\bm{94.93}_{\pm 0.14}$ \\
		$^{2.5}$GNN & $78.87_{\pm 0.57}$ & $63.28_{\pm 0.97}$ & $\bm{\underline{84.45}}_{\pm 0.18}$ & $83.85_{\pm 0.87}$ & $89.82_{\pm 0.64}$ & $91.94_{\pm 0.40}$ & $\bm{\underline{94.87}}_{\pm 0.11}$ \\	
		\bottomrule
	\end{tabular}
\end{table}

\subsection{Experimental Results of Aggregation Weight Entropy Distribution}\label{App:aggr}
\begin{figure}[htp]
	\centering
	\begin{subfigure}[b]{0.8\textwidth}
		\centering
		\includegraphics[width=\textwidth]{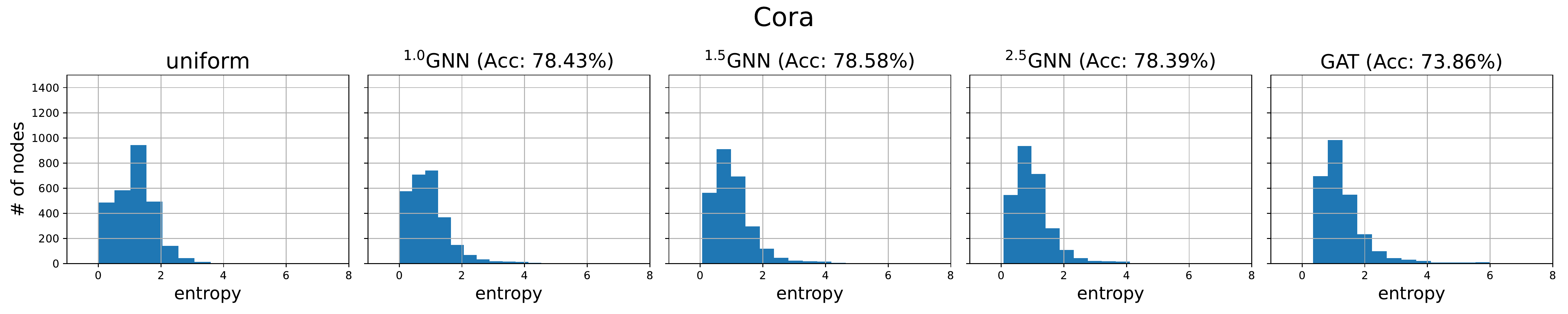}
	\end{subfigure}
	\begin{subfigure}[b]{0.8\textwidth}
		\centering
		\includegraphics[width=\textwidth]{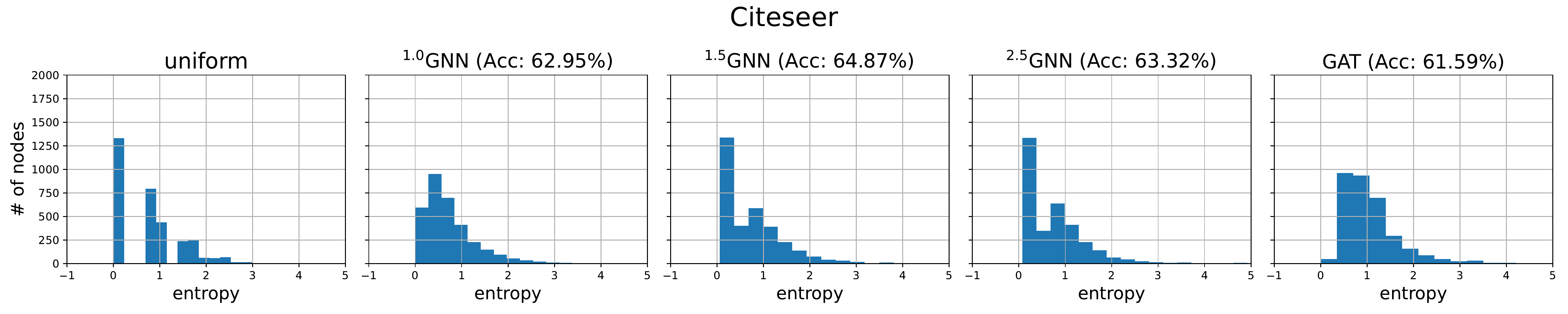}
	\end{subfigure}
	\begin{subfigure}[b]{0.8\textwidth}
		\centering
		\includegraphics[width=\textwidth]{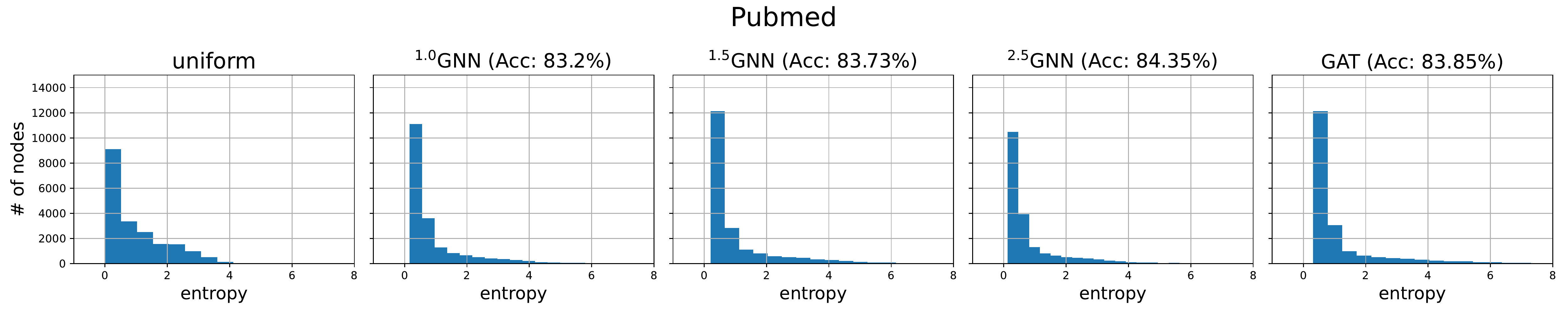}
	\end{subfigure}
	\begin{subfigure}[b]{0.8\textwidth}
		\centering
		\includegraphics[width=\textwidth]{figures/Computers.pdf}
	\end{subfigure}
	\begin{subfigure}[b]{0.8\textwidth}
		\centering
		\includegraphics[width=\textwidth]{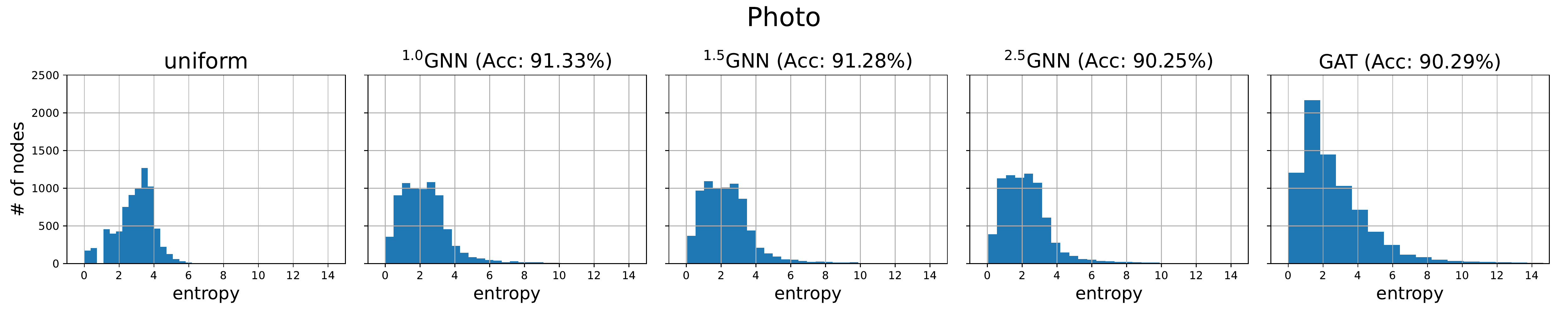}
	\end{subfigure}
	\begin{subfigure}[b]{0.8\textwidth}
		\centering
		\includegraphics[width=\textwidth]{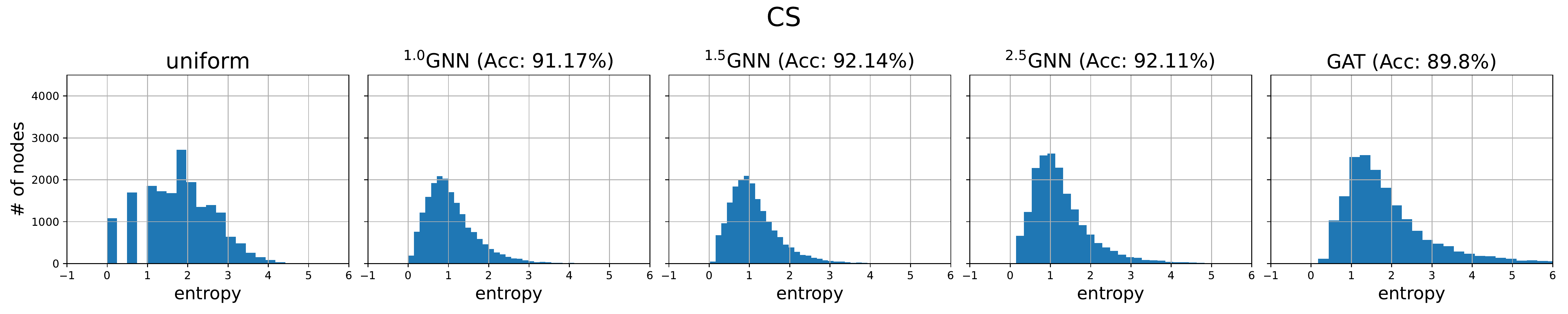}
	\end{subfigure}
	\begin{subfigure}[b]{0.8\textwidth}
		\centering
		\includegraphics[width=\textwidth]{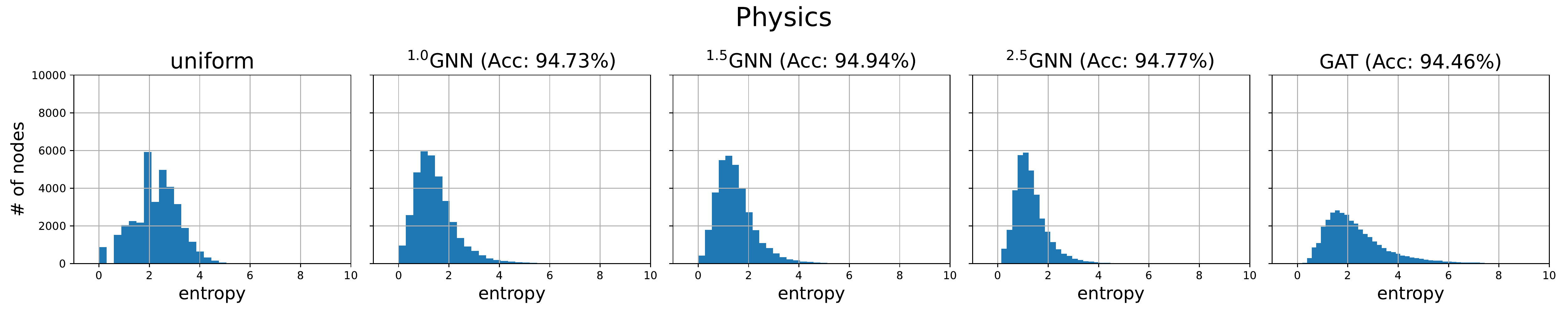}
	\end{subfigure}
	\vspace{-5pt}
	\caption{Aggregation weight entropy distribution of homophilic benchmark graphs. Low entropy means high degree of concentration, vice versa.  An entropy of zero means all aggregation weights are on one source node.}\label{App:fig:aggr_homo}
\end{figure}

\begin{figure}[htp]
	\centering
	\begin{subfigure}[b]{0.8\textwidth}
		\centering
		\includegraphics[width=\textwidth]{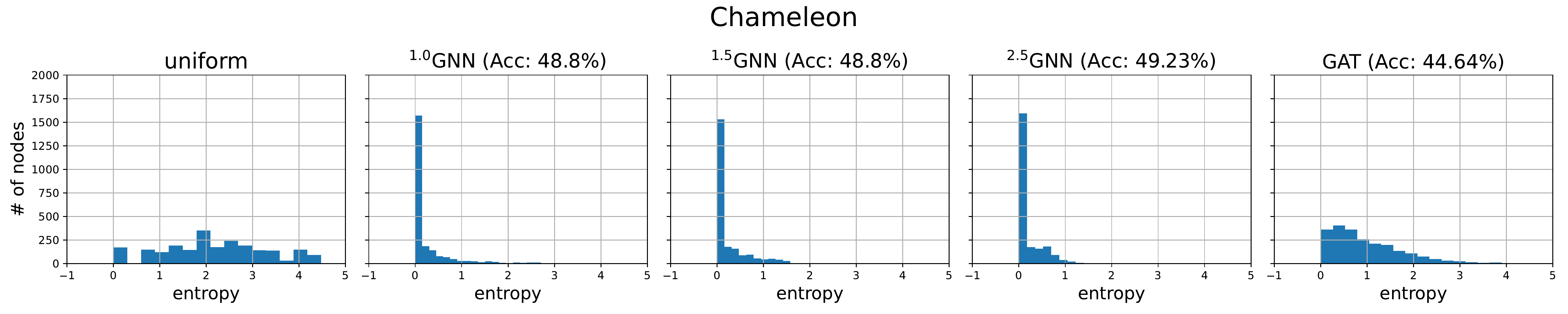}
	\end{subfigure}
	\begin{subfigure}[b]{0.8\textwidth}
		\centering
		\includegraphics[width=\textwidth]{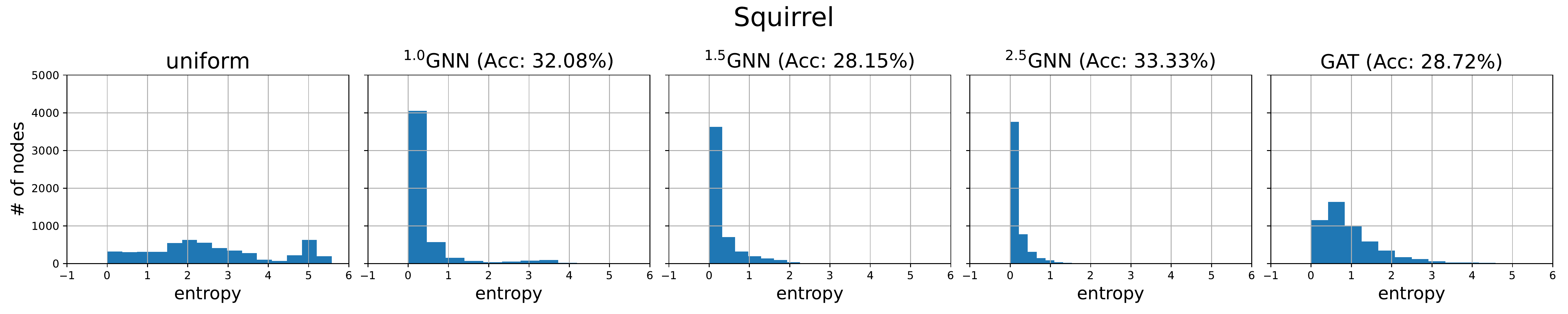}
	\end{subfigure}
	\begin{subfigure}[b]{0.8\textwidth}
		\centering
		\includegraphics[width=\textwidth]{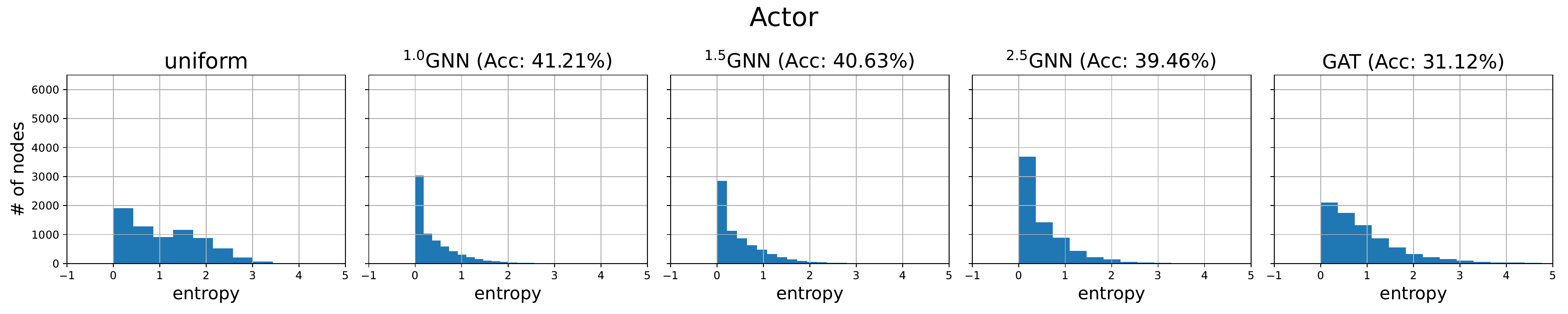}
	\end{subfigure}
	\begin{subfigure}[b]{0.8\textwidth}
		\centering
		\includegraphics[width=\textwidth]{figures/Wisconsin.pdf}
	\end{subfigure}
	\begin{subfigure}[b]{0.8\textwidth}
		\centering
		\includegraphics[width=\textwidth]{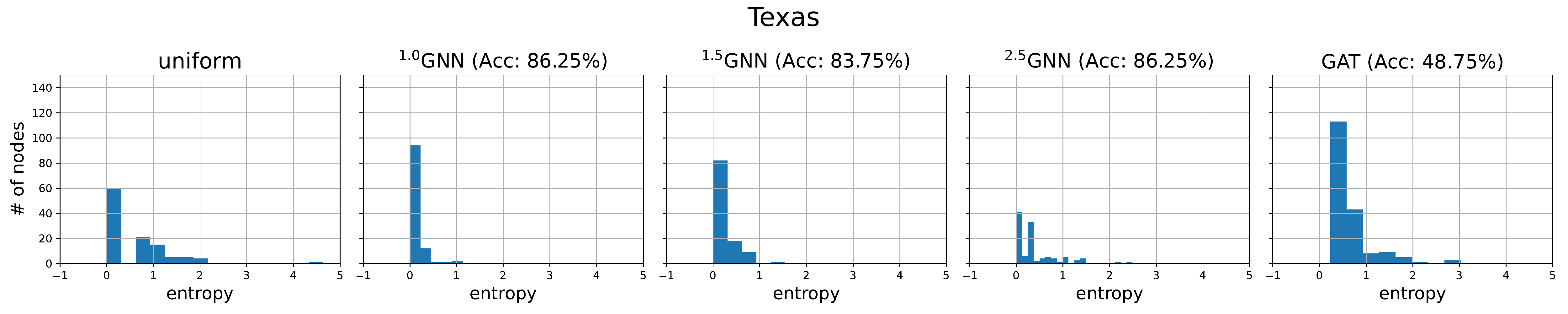}
	\end{subfigure}
	\begin{subfigure}[b]{0.8\textwidth}
		\centering
		\includegraphics[width=\textwidth]{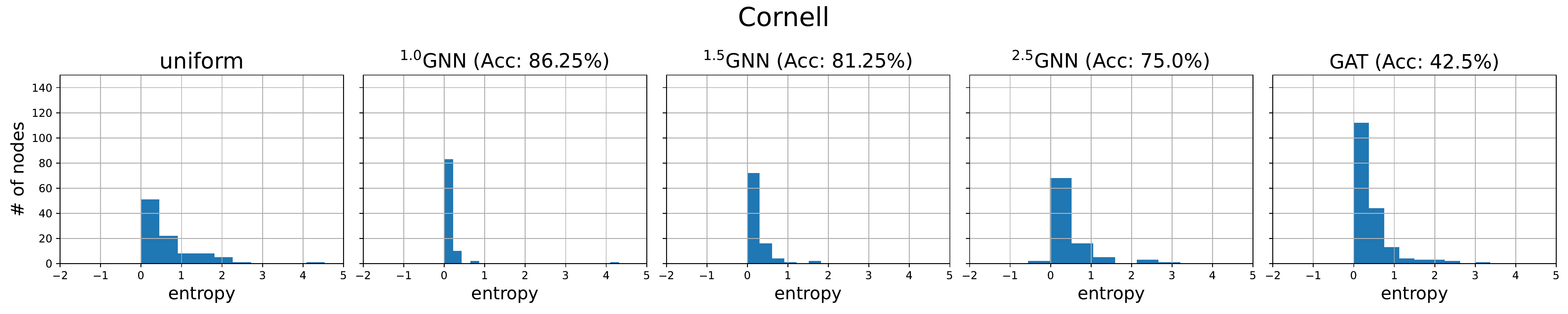}
	\end{subfigure}
	\vspace{-5pt}
	\caption{Aggregation weight entropy distribution of heterophilic benchmark graphs. Low entropy means high degree of concentration and vice versa.  An entropy of zero means all aggregation weights are on one source node.}\label{App:fig:aggr_heter}
\end{figure}

Here we present the visualization results of the learned aggregation weight entropy distribution of \pgnns and GAT on all benchmark datasets. \cref{App:fig:aggr_homo} and \cref{App:fig:aggr_heter} show the results obtained on homophilic and heterophilic benchmark datasets, respectively.

We observe from \cref{App:fig:aggr_homo} that the aggregation weight entropy distributions learned by \pgnns and GAT on homophilic benchmark datasets are similar to the uniform cases, which indicates that aggregating and transforming node features over the original graph topology is very helpful for label prediction. It explains why \pgnns and GNN baselines obtained similar performance on homophilic benchmark datasets and all GNN models significantly outperform MLP.

Contradict to the results on homophilic graphs shown in \cref{App:fig:aggr_homo}, \cref{App:fig:aggr_heter} shows that the aggregation weight entropy distributions of \pgnns on heterophilic benchmark datasets are very different from that of GAT and the uniform cases. We observe from \cref{App:fig:aggr_heter} that the entropy of most of the aggregation weights learned by \pgnns are around zero, which means that most aggregation weights are on one source node. It indicates that the graph topological information in these heterophilic benchmark graphs is not helpful for label prediction. Therefore, propagating and transforming node features over the graph topology could lead to worse performance than MLPs, which validates the results in \cref{App:tab:tab_homo_exp} that the performance of MLP is significantly better most GNN baselines on all heterophilic graphs and closed to \pgnns.

\subsection{Experimental Results on cSBM}\label{App:csbm}
In this section we present the experimental results on cSBM using sparse splitting and dense splitting, respectively. We used the same settings in \citet{DBLP:conf/iclr/ChienP0M21} in which the number of nodes $n = 5000$, the number of features $f = 2000$, $\epsilon = 3.25$ for all experiments. \cref{App:tab:tab_csbm_sparse} reports the results on cSBM with sparse splitting setting, which also are presented in \cref{fig:fig_csbm} and discussed in \cref{sec:sec5}. \cref{App:tab:tab_csbm_dense} reports the results on cSBM with dense splitting settings. 

\begin{table}[htp]
	\centering
	\caption{Results on cSBM with sparse splitting setting. Average accuracy (\%) for 20 runs. Best results are outlined in bold and the results within $95\%$ confidence interval of the best results are outlined in underlined bold.}\label{App:tab:tab_csbm_sparse}
	\vspace{-5pt}
	\tiny
	\setlength\tabcolsep{2pt}
	\begin{tabular}{cccccccccc}
		\toprule
		Method & $\phi = -1$ & $\phi = -0.75$ & $\phi = -0.5$ & $\phi = -0.25$ & $\phi = 0$ & $\phi = 0.25$ & $\phi = 0.5$ & $\phi = 0.75$ & $\phi = 1$ \\
		\midrule
		MLP & $49.72_{\pm 0.36}$ & $51.42_{\pm 1.83}$ & $59.21_{\pm 1.01}$ & $\bm{61.57}_{\pm 0.38}$ & $\bm{\underline{61.70}}_{\pm 0.30}$ & $59.92_{\pm 1.88}$ & $57.20_{\pm 0.62}$ & $54.48_{\pm 0.48}$ & $50.09_{\pm 0.51}$ \\
		GCN & $57.24_{\pm 1.15}$ & $58.19_{\pm 1.46}$ & $57.30_{\pm 1.30}$ & $51.97_{\pm 0.44}$ & $54.45_{\pm 1.38}$ & $64.70_{\pm 2.38}$ & $82.45_{\pm 1.35}$ & $91.31_{\pm 0.54}$ & $76.07_{\pm 3.30}$ \\
		SGC & $55.98_{\pm 1.48}$ & $58.56_{\pm 1.40}$ & $56.97_{\pm 0.54}$ & $51.54_{\pm 0.22}$ & $52.69_{\pm 2.36}$ & $64.14_{\pm 1.05}$ & $79.88_{\pm 1.57}$ & $90.37_{\pm 0.09}$ & $75.94_{\pm 0.92}$ \\
		GAT & $59.72_{\pm 2.23}$ & $60.20_{\pm 2.14}$ & $55.38_{\pm 1.96}$ & $50.15_{\pm 0.55}$ & $53.05_{\pm 1.40}$ & $64.00_{\pm 2.03}$ & $81.04_{\pm 1.71}$ & $90.37_{\pm 1.33}$ & $78.24_{\pm 1.95}$ \\
		JKNet & $49.70_{\pm 0.39}$ & $49.75_{\pm 0.79}$ & $49.65_{\pm 0.52}$ & $48.93_{\pm 0.48}$ & $52.36_{\pm 2.09}$ & $62.76_{\pm 2.54}$ & $87.10_{\pm 1.52}$ & $\bm{\underline{97.43}}_{\pm 0.36}$ & $\bm{\underline{97.69}}_{\pm 0.52}$ \\
		APPNP & $48.45_{\pm 0.98}$ & $49.65_{\pm 0.46}$ & $53.31_{\pm 0.89}$ & $56.58_{\pm 0.58}$ & $60.10_{\pm 0.65}$ & $65.02_{\pm 2.23}$ & $82.95_{\pm 1.38}$ & $95.49_{\pm 0.43}$ & $89.85_{\pm 0.60}$ \\
		GPRGNN & $97.26_{\pm 0.66}$ & $94.81_{\pm 0.91}$ & $82.14_{\pm 0.47}$ & $61.15_{\pm 2.55}$ & $60.20_{\pm 0.76}$ & $62.90_{\pm 2.22}$ & $83.61_{\pm 1.28}$ & $96.96_{\pm 0.41}$ & $\bm{98.01}_{\pm 0.71}$ \\
		\midrule
		$^{1.0}$GNN & $95.75_{\pm 1.21}$ & $93.06_{\pm 1.13}$ & $77.39_{\pm 4.21}$ & $\bm{\underline{61.38}}_{\pm 0.39}$ & $\bm{61.80}_{\pm 0.29}$ & $\bm{\underline{65.73}}_{\pm 2.11}$ & $85.85_{\pm 3.24}$ & $96.80_{\pm 0.87}$ & $97.40_{\pm 1.10}$ \\
		$^{1.5}$GNN & $95.90_{\pm 3.01}$ & $94.10_{\pm 4.57}$ & $73.08_{\pm 2.59}$ & $\bm{\underline{61.44}}_{\pm 0.30}$ & $\bm{\underline{61.77}}_{\pm 0.35}$ & $\bm{66.01}_{\pm 1.88}$ & $\bm{90.57}_{\pm 0.71}$ & $\bm{\underline{97.38}}_{\pm 0.43}$ & $\bm{\underline{97.76}}_{\pm 0.86}$ \\
		$^{2.0}$GNN & $\bm{98.37}_{\pm 0.78}$ & $\bm{\underline{96.32}}_{\pm 1.50}$ & $\bm{84.93}_{\pm 0.39}$ & $61.13_{\pm 0.51}$ & $\bm{\underline{61.79}}_{\pm 0.34}$ & $63.55_{\pm 1.73}$ & $88.55_{\pm 1.05}$ & $\bm{97.56}_{\pm 0.16}$ & $\bm{\underline{97.94}}_{\pm 0.39}$ \\
		$^{2.5}$GNN & $97.74_{\pm 0.99}$ & $\bm{96.78}_{\pm 0.44}$ & $83.21_{\pm 2.12}$ & $\bm{\underline{61.30}}_{\pm 0.41}$ & $\bm{\underline{61.74}}_{\pm 0.34}$ & $62.88_{\pm 2.31}$ & $79.64_{\pm 2.15}$ & $95.71_{\pm 0.34}$ & $97.25_{\pm 0.58}$ \\
		\bottomrule
	\end{tabular}
\end{table}

\begin{table}[htp]
	\centering
	\caption{Results on cSBM with dense splitting setting. Average accuracy (\%) for 20 runs. Best results are outlined in bold and the results within $95\%$ confidence interval of the best results are outlined in underlined bold.}\label{App:tab:tab_csbm_dense}
	\vspace{-5pt}
	\tiny
	\setlength\tabcolsep{2pt}
	\begin{tabular}{cccccccccc}
		\toprule
		Method & $\phi = -1$ & $\phi = -0.75$ & $\phi = -0.5$ & $\phi = -0.25$ & $\phi = 0$ & $\phi = 0.25$ & $\phi = 0.5$ & $\phi = 0.75$ & $\phi = 1$ \\
		\midrule
		MLP & $50.37_{\pm 0.60}$ & $65.22_{\pm 0.92}$ & $75.82_{\pm 0.65}$ & $81.18_{\pm 0.55}$ & $79.86_{\pm 0.69}$ & $79.97_{\pm 0.57}$ & $75.03_{\pm 0.89}$ & $67.53_{\pm 0.68}$ & $51.96_{\pm 0.69}$ \\
		GCN & $83.14_{\pm 0.49}$ & $82.59_{\pm 0.48}$ & $77.17_{\pm 0.59}$ & $58.58_{\pm 0.41}$ & $61.18_{\pm 1.06}$ & $82.59_{\pm 0.50}$ & $92.20_{\pm 0.27}$ & $97.21_{\pm 0.27}$ & $97.10_{\pm 0.12}$ \\
		SGC & $78.35_{\pm 0.36}$ & $82.13_{\pm 0.09}$ & $77.76_{\pm 0.12}$ & $59.14_{\pm 0.57}$ & $60.31_{\pm 0.63}$ & $81.96_{\pm 0.34}$ & $91.68_{\pm 0.13}$ & $96.56_{\pm 0.09}$ & $96.87_{\pm 0.05}$ \\
		GAT & $92.99_{\pm 0.86}$ & $90.89_{\pm 0.60}$ & $87.02_{\pm 0.80}$ & $68.40_{\pm 1.60}$ & $61.98_{\pm 1.16}$ & $82.92_{\pm 0.51}$ & $92.05_{\pm 0.73}$ & $97.28_{\pm 0.25}$ & $98.04_{\pm 0.46}$ \\
		JKNet & $68.95_{\pm 9.05}$ & $79.21_{\pm 7.67}$ & $67.97_{\pm 5.22}$ & $56.12_{\pm 4.10}$ & $58.33_{\pm 1.70}$ & $80.15_{\pm 0.80}$ & $91.21_{\pm 0.50}$ & $\bm{\underline{97.62}}_{\pm 0.25}$ & $\bm{\underline{98.32}}_{\pm 0.21}$ \\
		APPNP & $49.86_{\pm 0.39}$ & $50.47_{\pm 0.89}$ & $65.28_{\pm 0.68}$ & $73.98_{\pm 0.64}$ & $79.37_{\pm 0.66}$ & $86.60_{\pm 0.73}$ & $\bm{92.45}_{\pm 0.39}$ & $\bm{97.67}_{\pm 0.14}$ & $97.65_{\pm 0.49}$ \\
		GPRGNN & $\bm{\underline{99.06}}_{\pm 0.25}$ & $\bm{97.14}_{\pm 0.31}$ & $\bm{94.59}_{\pm 0.32}$ & $\bm{83.84}_{\pm 0.69}$ & $78.81_{\pm 1.30}$ & $85.85_{\pm 1.01}$ & $92.08_{\pm 0.81}$ & $\bm{\underline{97.49}}_{\pm 0.22}$ & $\bm{98.46}_{\pm 0.15}$ \\
		\midrule
		$^{1.0}$GNN & $98.19_{\pm 0.28}$ & $94.38_{\pm 0.44}$ & $86.40_{\pm 1.00}$ & $80.57_{\pm 0.43}$ & $\bm{\underline{80.21}}_{\pm 0.42}$ & $\bm{87.32}_{\pm 0.47}$ & $\bm{\underline{92.42}}_{\pm 0.62}$ & $\bm{\underline{97.52}}_{\pm 0.33}$ & $\bm{\underline{98.37}}_{\pm 0.26}$ \\
		$^{1.5}$GNN & $\bm{\underline{98.88}}_{\pm 0.16}$ & $95.62_{\pm 0.21}$ & $86.87_{\pm 1.22}$ & $80.70_{\pm 0.71}$ & $\bm{80.28}_{\pm 0.31}$ & $86.29_{\pm 0.43}$ & $\bm{\underline{92.40}}_{\pm 0.24}$ & $\bm{\underline{97.56}}_{\pm 0.25}$ & $\bm{\underline{98.24}}_{\pm 0.32}$ \\
		$^{2.0}$GNN & $\bm{99.21}_{\pm 0.09}$ & $\bm{\underline{96.91}}_{\pm 0.16}$ & $92.96_{\pm 0.31}$ & $80.83_{\pm 0.61}$ & $\bm{\underline{80.04}}_{\pm 0.49}$ & $84.96_{\pm 0.60}$ & $91.18_{\pm 0.27}$ & $\bm{\underline{97.41}}_{\pm 0.14}$ & $\bm{\underline{98.45}}_{\pm 0.14}$ \\
		$^{2.5}$GNN & $\bm{99.21}_{\pm 0.14}$ & $\bm{\underline{96.94}}_{\pm 0.16}$ & $93.28_{\pm 0.37}$ & $80.93_{\pm 0.44}$ & $\bm{80.28}_{\pm 0.38}$ & $83.83_{\pm 0.70}$ & $86.10_{\pm 0.39}$ & $96.28_{\pm 0.43}$ & $97.76_{\pm 0.18}$ \\
		\bottomrule
	\end{tabular}
	\vspace{-10pt}
\end{table}

\cref{App:tab:tab_csbm_dense} shows that \pgnns obtain the best performance on weak homophilic graphs ($\phi = 0, 0.25$) while competitive performance against GPRGNN on strong heterophilic graphs ($\phi = -0.75, -1$) and competitive performance with state-of-the-art GNNs on strong homophilic graphs ($\phi = 0.75, 1$). We also observe that GPRGNN is slightly better than \pgnns on weak heterophilic graphs ($\phi = -0.25, -0.5$), which suggests that GPRGNN could work very well using strong supervised information ($60\%$ training rate and $20\%$ validation rate). However, as shown in \cref{App:tab:tab_csbm_sparse}, \pgnns work better than GPRGNN under weak supervised information ($2.5\%$ training rate and $2.5\%$) on all heterophilic graphs. The result is reasonable, as discussed in \cref{remk:remk2} in \cref{subsec:pgnn}, GPRGNN can adaptively learn the generalized PageRank (GPR) weights and it works similarly to $^{2.0}$GNN on both homophilic and heterophilic graphs. However, it needs more supervised information in order to learn optimal GPR weights. On the contrary, \pgnns need less supervised information to obtain similar results because $\Theta^{(2)}$ acts like a hyperplane for classification. Therefore, \pgnns can work better under weak supervised information.

\subsection{Experimental Results on Graphs with Noisy Edges}\label{App:noisy_exp}
Here we present more experimental results on graph with noisy edges. \cref{App:tab:noisy_exp_homo} reports the results on homophilic graphs (Computers, Photo, CS, Physics) and \cref{App:tab:noisy_exp_heter} reports the results on heterophilic graphs (Wisconsin Texas). We observe from \cref{App:tab:noisy_exp_homo,App:tab:noisy_exp_heter} that \pgnns dominate all baselines. Moreover, \pgnns even slightly better than MLP when the graph topologies are completely random, i.e. the noisy edge rate $r = 1$. We also observe that the performance of GCN, SGC, JKNet on homophilic graphs dramatically degrades as the noisy edge rate $r$ increases while they do not change a lot for the cases on heterophilic graphs. It is reasonable since the original graph topological information is very helpful for label prediction on these homophilic graphs. Adding noisy edges and remove the same number of original edges could significantly degrade the performance of ordinary GNNs. On the other hand, since we find that the original graph topological information in Wisconsin and Texas is not helpful for label prediction. Therefore, adding noisy edges and removing original edges on these heterophilic graphs would not affect too much their performance.

\begin{table}[htp]
	\centering
	\caption{Results on homophilic graphs with random edges. Average accuracy (\%) for 20 runs. Best results are outlined in bold and the results within $95\%$ confidence interval of the best results are outlined in underlined bold. OOM denotes out of memory.}\label{App:tab:noisy_exp_homo}
	\vspace{-5pt}
	\footnotesize
	\setlength\tabcolsep{2pt}
	\begin{tabular}{ccccccc}
		\toprule
		\multirow{2}{*}{Method} & \multicolumn{3}{c}{Computers} & \multicolumn{3}{c}{Photo} \\
		\cmidrule{2-7}
		& $r=0.25$ & $r=0.5$ & $r=1$ & $r=0.25$ & $r=0.5$ & $r=1$ \\
		\midrule
		MLP & $66.11_{\pm 2.70}$ & $66.11_{\pm 2.70}$ & $66.11_{\pm 2.70}$ & $76.44_{\pm 2.83}$ & $76.44_{\pm 2.83}$ & $76.44_{\pm 2.83}$ \\
		GCN & $74.70_{\pm 1.72}$ & $62.16_{\pm 2.76}$ & $8.95_{\pm 6.90}$ & $81.43_{\pm 0.76}$ & $75.52_{\pm 3.59}$ & $12.78_{\pm 5.20}$ \\ 
		SGC & $75.15_{\pm 1.08}$ & $66.96_{\pm 1.05}$ & $15.79_{\pm 7.47}$ & $82.22_{\pm 0.36}$ & $77.80_{\pm 0.49}$ & $13.57_{\pm 3.63}$ \\ 
		GAT & $76.44_{\pm 1.81}$ & $68.34_{\pm 2.61}$ & $11.58_{\pm 7.70}$ & $82.70_{\pm 1.31}$ & $77.20_{\pm 2.10}$ & $13.74_{\pm 5.14}$ \\ 
		JKNet & $56.74_{\pm 6.48}$ & $46.11_{\pm 8.43}$ & $12.50_{\pm 6.56}$ & $73.46_{\pm 6.74}$ & $64.18_{\pm 4.06}$ & $15.66_{\pm 6.10}$ \\ 
		APPNP & $78.23_{\pm 1.84}$ & $74.57_{\pm 2.25}$ & $66.67_{\pm 2.68}$ & $87.63_{\pm 1.05}$ & $86.22_{\pm 1.73}$ & $75.55_{\pm 1.72}$ \\ 
		GPRGNN & $77.30_{\pm 2.24}$ & $77.11_{\pm 1.80}$ & $66.85_{\pm 1.65}$ & $85.95_{\pm 1.05}$ & $85.64_{\pm 1.22}$ & $\bm{77.46}_{\pm 1.44}$ \\ 
		\midrule 
		$^{1.0}$GNN & $75.14_{\pm 14.95}$ & $63.26_{\pm 20.67}$ & $41.60_{\pm 16.17}$ & $\bm{\underline{87.97}}_{\pm 0.70}$ & $84.47_{\pm 3.05}$ & $41.17_{\pm 18.15}$ \\
		$^{1.5}$GNN & $\bm{81.79}_{\pm 1.33}$ & $\bm{78.12}_{\pm 2.08}$ & $66.04_{\pm 2.73}$ & $\bm{88.09}_{\pm 1.18}$ & $86.20_{\pm 1.61}$ & $68.78_{\pm 8.97}$ \\
		$^{2.0}$GNN & $80.34_{\pm 1.07}$ & $76.90_{\pm 1.93}$ & $\bm{67.17}_{\pm 1.63}$ & $87.65_{\pm 0.94}$ & $\bm{87.06}_{\pm 1.50}$ & $\bm{\underline{77.07}}_{\pm 1.83}$ \\
		$^{2.5}$GNN & $79.14_{\pm 1.51}$ & $75.49_{\pm 1.25}$ & $64.95_{\pm 2.27}$ & $87.38_{\pm 0.85}$ & $86.11_{\pm 1.10}$ & $76.65_{\pm 1.46}$ \\
		\bottomrule
		\toprule
		\multirow{2}{*}{Method} & \multicolumn{3}{c}{CS} & \multicolumn{3}{c}{Physics} \\
		\cmidrule{2-7}
		& $r=0.25$ & $r=0.5$ & $r=1$ & $r=0.25$ & $r=0.5$ & $r=1$ \\
		\midrule
		MLP & $86.24_{\pm 1.43}$ & $86.24_{\pm 1.43}$ & $\bm{\underline{86.24}}_{\pm 1.43}$ & $92.58_{\pm 0.83}$ & $92.58_{\pm 0.83}$ & $92.58_{\pm 0.83}$ \\
        GCN & $81.05_{\pm 0.59}$ & $68.37_{\pm 0.85}$ & $7.72_{\pm 2.39}$ & $89.02_{\pm 0.16}$ & $80.45_{\pm 0.34}$ & $19.78_{\pm 3.94}$ \\
        SGC & $83.41_{\pm 0.01}$ & $71.98_{\pm 0.12}$ & $8.00_{\pm 1.43}$ & \text{OOM} & \text{OOM} & \text{OOM} \\
        GAT & $80.11_{\pm 0.67}$ & $68.66_{\pm 1.42}$ & $8.49_{\pm 2.39}$ & $88.72_{\pm 0.61}$ & $82.05_{\pm 1.86}$ & $22.39_{\pm 5.04}$ \\
        JKNet & $81.35_{\pm 0.74}$ & $71.30_{\pm 2.14}$ & $11.43_{\pm 1.18}$ & $87.98_{\pm 0.97}$ & $81.90_{\pm 2.27}$ & $26.38_{\pm 5.80}$ \\
        APPNP & $88.63_{\pm 0.68}$ & $87.56_{\pm 0.51}$ & $76.90_{\pm 0.96}$ & $93.46_{\pm 0.12}$ & $92.81_{\pm 0.24}$ & $90.49_{\pm 0.33}$ \\
        GPRGNN & $85.77_{\pm 0.81}$ & $83.89_{\pm 1.54}$ & $72.79_{\pm 2.24}$ & $92.18_{\pm 0.29}$ & $90.96_{\pm 0.48}$ & $91.77_{\pm 0.41}$ \\
        \midrule
        $^{1.0}$GNN & $90.27_{\pm 0.86}$ & $89.56_{\pm 0.81}$ & $\bm{86.60}_{\pm 1.22}$ & $\bm{94.35}_{\pm 0.39}$ & $\bm{94.23}_{\pm 0.27}$ & $\bm{92.97}_{\pm 0.36}$ \\
        $^{1.5}$GNN & $\bm{91.27}_{\pm 0.40}$ & $\bm{90.50}_{\pm 0.71}$ & $84.40_{\pm 1.84}$ & $\bm{\underline{94.34}}_{\pm 0.21}$ & $93.77_{\pm 0.29}$ & $92.51_{\pm 0.35}$ \\
        $^{2.0}$GNN & $90.97_{\pm 0.49}$ & $89.98_{\pm 0.50}$ & $80.84_{\pm 1.48}$ & $\bm{\underline{94.14}}_{\pm 0.18}$ & $93.30_{\pm 0.31}$ & $91.72_{\pm 0.44}$ \\
        $^{2.5}$GNN & $89.90_{\pm 0.45}$ & $89.00_{\pm 0.59}$ & $76.82_{\pm 2.11}$ & $93.61_{\pm 0.30}$ & $92.77_{\pm 0.26}$ & $91.16_{\pm 0.47}$ \\
		\bottomrule
	\end{tabular}
\end{table}

\begin{table}[htp]
	\centering
	\caption{Results on heterophilic graphs with random edges. Average accuracy (\%) for 20 runs. Best results are outlined in bold and the results within $95\%$ confidence interval of the best results are outlined in underlined bold.}\label{App:tab:noisy_exp_heter}
	\vspace{-5pt}
	\footnotesize
	\setlength\tabcolsep{2pt}
	\begin{tabular}{ccccccc}
		\toprule
		\multirow{2}{*}{Method} & \multicolumn{3}{c}{Wisconsin} & \multicolumn{3}{c}{Texas} \\
		\cmidrule{2-7}
		& $r=0.25$ & $r=0.5$ & $r=1$ & $r=0.25$ & $r=0.5$ & $r=1$ \\
		\midrule
		MLP & $93.56_{\pm 3.14}$ & $93.56_{\pm 3.14}$ & $93.56_{\pm 3.14}$ & $79.50_{\pm 10.62}$ & $79.50_{\pm 10.62}$ & $79.50_{\pm 10.62}$ \\
		GCN & $62.31_{\pm 8.12}$ & $59.44_{\pm 5.76}$ & $64.21_{\pm 4.49}$ & $41.56_{\pm 8.89}$ & $44.69_{\pm 23.05}$ & $40.31_{\pm 18.26}$ \\
		SGC & $64.68_{\pm 7.34}$ & $62.36_{\pm 2.64}$ & $51.81_{\pm 2.63}$ & $42.50_{\pm 5.49}$ & $40.94_{\pm 18.34}$ & $23.81_{\pm 14.54}$ \\
		GAT & $65.37_{\pm 9.04}$ & $60.05_{\pm 9.12}$ & $60.05_{\pm 7.46}$ & $39.50_{\pm 9.29}$ & $34.88_{\pm 21.59}$ & $29.38_{\pm 11.53}$ \\
		JKNet & $64.91_{\pm 13.07}$ & $51.39_{\pm 10.36}$ & $57.41_{\pm 2.57}$ & $47.75_{\pm 7.30}$ & $46.62_{\pm 23.23}$ & $40.69_{\pm 13.57}$ \\
		APPNP & $70.19_{\pm 9.04}$ & $60.32_{\pm 4.70}$ & $72.64_{\pm 4.73}$ & $66.69_{\pm 13.46}$ & $63.25_{\pm 9.87}$ & $69.81_{\pm 7.76}$ \\
		GPRGNN & $90.97_{\pm 3.83}$ & $87.50_{\pm 3.86}$ & $87.55_{\pm 2.97}$ & $74.25_{\pm 7.25}$ & $76.75_{\pm 14.05}$ & $80.69_{\pm 5.87}$ \\
		\midrule
		$^{1.0}$GNN & $\bm{94.91}_{\pm 2.73}$ & $\bm{95.97}_{\pm 2.00}$ & $\bm{95.97}_{\pm 2.27}$ & $81.50_{\pm 9.24}$ & $\bm{82.12}_{\pm 11.09}$ & $\bm{81.81}_{\pm 5.67}$ \\
		$^{1.5}$GNN & $\bm{\underline{94.58}}_{\pm 1.25}$ & $95.19_{\pm 2.18}$ & $94.95_{\pm 2.79}$ & $82.50_{\pm 6.39}$ & $78.12_{\pm 5.30}$ & $78.50_{\pm 7.98}$ \\
		$^{2.0}$GNN & $90.46_{\pm 2.79}$ & $90.97_{\pm 4.22}$ & $91.44_{\pm 2.27}$ & $\bm{86.06}_{\pm 5.17}$ & $69.38_{\pm 11.47}$ & $63.50_{\pm 8.90}$ \\
		$^{2.5}$GNN & $82.45_{\pm 3.93}$ & $88.24_{\pm 2.79}$ & $84.40_{\pm 1.98}$ & $80.00_{\pm 10.83}$ & $56.62_{\pm 10.01}$ & $52.31_{\pm 10.58}$ \\
		\bottomrule
	\end{tabular}
\end{table}

\subsection{Experimental Results of Intergrating \pgnns with GCN and JKNet}\label{App:plug_exp}
Here we further conduct experiments to study whether \pgnns can be intergrated into existing GNN architectures and improve their performance on heterophilic graphs. We use two popular GNN architectures: GCN~\citep{DBLP:conf/iclr/KipfW17} and JKNet~\citep{DBLP:conf/icml/XuLTSKJ18}. 

To incorporate \pgnns with GCN, we use the \pgnn layers as the first layer of the combined models, termed as \pgnn + GCN, and GCN layer as the second layer. Specifically, we use the aggregation weights $\bm{\alpha}\mD^{-1/2}\mM\mD^{-1/2}$ learned by the \pgnn in the first layer as the input edge weights of GCN layer in the second layer. To combine \pgnn with JKNet, we use the \pgnn layer as the GNN layers in the JKNet framework, termed as \pgnn + JKNet. \cref{App:tab:plug_exp_homo,App:tab:plug_exp_heter} report the experimental results on homophilic and heterophilic benchmark datasets, respectively.

\begin{table}[htp]
	\centering
	\caption{The results of \pgnns + GCN and \pgnns + JKNet on homophilic benchmark dataset. Averaged accuracy (\%) for 20 runs. Best results are outlined in bold and the results within $95\%$ confidence interval of the best results are outlined in underlined bold.}\label{App:tab:plug_exp_homo}
	\vspace{-5pt}
	\footnotesize
	\setlength\tabcolsep{2pt}
	\begin{tabular}{cccccccc}
		\toprule
		Method & Cora & CiteSeer & PubMed & Computers & Photo & CS & Physics \\
		\midrule
		GCN & $\bm{76.23}_{\pm 0.79}$ & $\bm{62.43}_{\pm 0.81}$ & $\bm{83.72}_{\pm 0.27}$ & $\bm{84.17}_{\pm 0.59}$ & $\bm{90.46}_{\pm 0.48}$ & $\bm{90.33}_{\pm 0.36}$ & $94.46_{\pm 0.08}$ \\
		$^{1.0}$GNN + GCN & $72.37_{\pm 1.35}$ & $60.56_{\pm 1.59}$ & $82.14_{\pm 0.31}$ & $83.75_{\pm 1.05}$ & $\bm{\underline{90.24}}_{\pm 1.12}$ & $89.60_{\pm 0.46}$ & $\bm{\underline{94.59}}_{\pm 0.33}$ \\
		$^{1.5}$GNN + GCN & $72.72_{\pm 1.39}$ & $60.23_{\pm 1.80}$ & $82.21_{\pm 0.22}$ & $\bm{\underline{83.89}}_{\pm 0.74}$ & $90.00_{\pm 0.68}$ & $89.48_{\pm 0.45}$ & $\bm{94.70}_{\pm 0.18}$ \\
		$^{2.0}$GNN + GCN & $72.39_{\pm 1.55}$ & $60.19_{\pm 1.60}$ & $82.24_{\pm 0.23}$ & $\bm{\underline{83.92}}_{\pm 1.09}$ & $\bm{\underline{90.17}}_{\pm 0.83}$ & $89.60_{\pm 0.71}$ & $94.51_{\pm 0.39}$ \\
		$^{2.5}$GNN + GCN & $72.85_{\pm 1.19}$ & $59.68_{\pm 1.85}$ & $82.23_{\pm 0.34}$ & $83.69_{\pm 0.92}$ & $90.02_{\pm 1.09}$ & $89.53_{\pm 0.68}$ & $\bm{\underline{94.58}}_{\pm 0.31}$ \\
		\bottomrule
		\toprule
		JKNet & $\bm{77.19}_{\pm 0.98}$ & $\bm{63.32}_{\pm 0.95}$ & $\bm{\underline{82.54}}_{\pm 0.43}$ & $79.94_{\pm 2.47}$ & $88.29_{\pm 1.64}$ & $89.69_{\pm 0.66}$ & $93.92_{\pm 0.32}$ \\
		$^{1.0}$GNN+JKNet & $75.67_{\pm 1.54}$ & $60.38_{\pm 1.65}$ & $81.68_{\pm 0.44}$ & $\bm{83.19}_{\pm 1.36}$ & $89.71_{\pm 1.05}$ & $90.26_{\pm 0.72}$ & $94.27_{\pm 0.69}$ \\
		$^{1.5}$GNN+JKNet & $76.40_{\pm 1.59}$ & $60.67_{\pm 1.93}$ & $\bm{\underline{82.42}}_{\pm 0.35}$ & $82.78_{\pm 2.09}$ & $\bm{90.25}_{\pm 1.03}$ & $\bm{90.76}_{\pm 0.75}$ & $\bm{94.82}_{\pm 0.34}$ \\
		$^{2.0}$GNN+JKNet & $76.75_{\pm 1.26}$ & $61.05_{\pm 1.48}$ & $\bm{\underline{82.50}}_{\pm 0.53}$ & $82.36_{\pm 2.39}$ & $89.31_{\pm 1.39}$ & $90.33_{\pm 0.63}$ & $\bm{\underline{94.70}}_{\pm 0.33}$ \\
		$^{2.5}$GNN+JKNet & $76.48_{\pm 1.28}$ & $60.97_{\pm 0.97}$ & $82.56_{\pm 1.04}$ & $81.45_{\pm 1.55}$ & $89.21_{\pm 1.10}$ & $89.66_{\pm 0.68}$ & $94.29_{\pm 0.59}$ \\
		\bottomrule
	\end{tabular}
\end{table}

\begin{table}[htp]
	\centering
	\caption{The results of \pgnns + GCN and \pgnns + JKNet on heterophilic benchmark dataset. Averaged accuracy (\%) for 20 runs. Best results are outlined in bold and the results within $95\%$ confidence interval of the best results are outlined in underlined bold.}\label{App:tab:plug_exp_heter}
	\vspace{-5pt}
	\footnotesize
	\setlength\tabcolsep{2pt}
	\begin{tabular}{ccccccc}
		\toprule
		Method & Chameleon & Squirrel & Actor & Wisconsin & Texas & Cornell \\
		\midrule
		GCN & $34.54_{\pm 2.78}$ & $25.28_{\pm 1.55}$ & $31.28_{\pm 2.04}$ & $61.93_{\pm 3.00}$ & $56.54_{\pm 17.02}$ & $51.36_{\pm 4.59}$ \\
		$^{1.0}$GNN + GCN & $48.52_{\pm 1.89}$ & $\bm{\underline{34.78}}_{\pm 1.11}$ & $32.37_{\pm 3.12}$ & $\bm{68.52}_{\pm 3.75}$ & $\bm{\underline{67.94}}_{\pm 12.60}$ & $\bm{\underline{67.81}}_{\pm 7.61}$ \\
		$^{1.5}$GNN + GCN & $48.85_{\pm 2.13}$ & $34.61_{\pm 1.11}$ & $32.37_{\pm 2.48}$ & $66.25_{\pm 3.95}$ & $65.62_{\pm 11.99}$ & $64.88_{\pm 9.19}$ \\
		$^{2.0}$GNN + GCN & $48.71_{\pm 2.24}$ & $\bm{35.06}_{\pm 1.18}$ & $\bm{32.72}_{\pm 2.02}$ & $66.34_{\pm 4.51}$ & $65.94_{\pm 7.63}$ & $\bm{68.62}_{\pm 6.55}$ \\
		$^{2.5}$GNN + GCN & $\bm{49.53}_{\pm 2.19}$ & $34.40_{\pm 1.60}$ & $32.40_{\pm 3.23}$ & $67.18_{\pm 3.50}$ & $\bm{68.31}_{\pm 9.18}$ & $66.06_{\pm 9.56}$ \\
		\bottomrule
		\toprule
		JKNet & $33.28_{\pm 3.59}$ & $25.82_{\pm 1.58}$ & $29.77_{\pm 2.61}$ & $61.08_{\pm 3.71}$ & $59.65_{\pm 12.62}$ & $55.34_{\pm 4.43}$ \\
		$^{1.0}$GNN + JKNet & $\bm{\underline{49.00}}_{\pm 2.09}$ & $35.56_{\pm 1.34}$ & $\bm{40.74}_{\pm 0.98}$ & $\bm{95.23}_{\pm 2.43}$ & $80.25_{\pm 6.87}$ & $\bm{78.38}_{\pm 8.14}$ \\
		$^{1.5}$GNN + JKNet & $48.77_{\pm 2.22}$ & $\bm{35.98}_{\pm 0.93}$ & $40.22_{\pm 1.27}$ & $94.86_{\pm 2.00}$ & $80.38_{\pm 9.79}$ & $72.25_{\pm 9.83}$ \\
		$^{2.0}$GNN + JKNet & $48.88_{\pm 1.63}$ & $35.77_{\pm 1.73}$ & $40.16_{\pm 1.31}$ & $88.84_{\pm 2.78}$ & $\bm{86.12}_{\pm 5.59}$ & $74.75_{\pm 7.81}$ \\
		$^{2.5}$GNN + JKNet & $\bm{49.04}_{\pm 1.95}$ & $35.78_{\pm 1.87}$ & $40.00_{\pm 1.12}$ & $85.42_{\pm 3.86}$ & $79.06_{\pm 7.60}$ & $76.81_{\pm 7.66}$ \\
		\bottomrule
	\end{tabular}
\end{table}

We observe from \cref{App:tab:plug_exp_homo} that intergrating \pgnns with GCN and JKNet does not improve their performance on homophilic graphs. The performance of GCN slightly degrade after incorporating \pgnns. The performance of JKNet also slightly degrade on Cora, CiteSeer, and PubMed but is improved on Computers, Photo, CS, Physics. It is reasonable since GCN and JKNet can predict well on these homophilic benchmark datasets based on their original graph topology.

However, for heterophilic benchmark datasets, \cref{App:tab:plug_exp_heter} shows that there are significant improvements over GCN, and JKNet after intergrating with \pgnns. Moreover, \pgnns + JKNet obtain advanced performance on all heterophilic benchmark datasets and even better than \pgnns on Squirrel. The results of \cref{App:tab:plug_exp_heter} demonstrate that intergrating \pgnns with  GCN and JKNet can sigificantly improve their performance on heterophilic graphs.

\subsection{Experimental Results of \pgnns on PPI Dataset for Inductive Learning}\label{App:ppi_exp}
Additionally, we conduct comparison experiments of \pgnns against GAT on PPI dataset~\citep{DBLP:journals/bioinformatics/ZitnikL17} using the inductive learning settings as in \citet{DBLP:conf/iclr/VelickovicCCRLB18} (20 graphs for training, 2 graphs for validation, 2 graphs for testing). We use three layers of GAT architecture with 256 hidden units, use 1 attention head for GAT (1 head) and 4 attention heads for GAT (4 heads). We use three \pgnn layers and a MLP layer as the first layer for \pgnns, set $\mu = 0.01$, $K = 1$, and use 256 hidden units for \pgnn-256 and 512 hidden units for \pgnn-512. The experimental results are reported in \cref{App:tab:ppi_exp}.

\begin{table}[htp]
	\centering
	\caption{Results on PPI datasets. Averaged micro-F1 scores for 10 runs. Best results are outlined in bold.}\label{App:tab:ppi_exp}
	\vspace{-5pt}
	\footnotesize
	\begin{tabular}{ll}
		\toprule
		Method & PPI \\
		\midrule
		GAT (1 head) & $0.917 \pm 0.041$ \\
		GAT (4 heads) & $0.972 \pm 0.002$ \\
		\midrule
		$^{1.0}$GNN-256 & $0.961 \pm 0.003$ \\
		$^{1.5}$GNN-256 & $0.967 \pm 0.008$ \\
		$^{2.0}$GNN-256 & $0.968 \pm 0.006$ \\
		$^{2.5}$GNN-256 & $0.973 \pm 0.002$ \\
		$^{1.0}$GNN-512 & $0.978 \pm 0.005$ \\
		$^{1.5}$GNN-512 & $0.977 \pm 0.008$ \\
		$^{2.0}$GNN-512 & $\bm{0.981} \pm 0.006$ \\
		$^{2.5}$GNN-512 & $0.978 \pm 0.005$ \\
		\bottomrule
	\end{tabular}
\end{table}

From \cref{App:ppi_exp} we observe that the results of $^{2.5}$GNN on PPI slightly better than GAT with 4 attention heads and other \pgnns are very close to it. Moreover, all results of \pgnns significantly outperform GAT with one attention head. The results of \pgnns on PPI is impressive. \pgnns have much less parameters than GAT with 4 attention heads while obtain very completitive performance on PPI. When we use more hidden units, 512 hidden units, \pgnns-512 significantly outperform GAT, while \pgnns-512 still have less parameters. It illustrates the superior potential of applying \pgnns to inducting learning on graphs.

\subsection{Experimental Results of \pgnns on OGBN arXiv Dataset}\label{App:obg_exp}
\begin{table}[htp]
	\centering
	\caption{Results on OGBN arXiv dataset. Average accuracy (\%) for 10 runs. Best results are outlined in bold.}\label{App:tab:ogbn_exp}
	\vspace{-5pt}
	\footnotesize
	\begin{tabular}{ll}
		\toprule
		Method & OGBN arXiv \\
		\midrule
		MLP & $55.50 \pm 0.23$ \\
		GCN & $71.74 \pm 0.29$ \\
		JKNet (GCN-based) & $72.19 \pm 0.21$ \\
		DeepGCN & $71.92 \pm 0.16$ \\
		GCN + residual (6 layers) & $72.86 \pm 0.16$ \\
		GCN + residual (8 layers) + C\&S & $72.97 \pm 0.22$ \\
		GCN + residual (8 layers) + C\&S v2 & $73.13 \pm 0.17$ \\
		\midrule
		$^{1}$GNN & $72.40 \pm 0.19$ \\
		$^{2}$GNN & $72.45 \pm 0.20$ \\
		$^{3}$GNN & $72.58 \pm 0.23$ \\
		$^{1}$GNN + residual (6 layers) + C\&S & $72.96 \pm 0.22$ \\
		$^{2}$GNN + residual (6 layers) + C\&S & $73.13 \pm 0.20$ \\
		$^{3}$GNN + residual (6 layers) + C\&S & $\bm{73.23} \pm 0.16$ \\
		\bottomrule
	\end{tabular}
\end{table}
Here we present the experimental of \pgnns on OGBN arXiv dataset~\citep{DBLP:conf/nips/HuFZDRLCL20}. We use the official data split setting of OGBN arXiv. We use three layers \pgnn architecture and 256 hidden units with $\mu = 0.5$, $K = 2$. We also combine \pgnns with correct and smooth model (C\&S)~\citep{DBLP:conf/iclr/HuangHSLB21} and introduce residual units. The results of MLP, GCN, JKNet, DeepGCN~\citep{DBLP:conf/iccv/Li0TG19}, GCN with residual units, C\&S model are extracted from the leaderboard for OGBN arXiv. dataset\footnote{https://ogb.stanford.edu/docs/leader\_nodeprop/\#ogbn-arxiv}. \cref{App:tab:ogbn_exp} summaries the results of \pgnns against the baselines.

We observe from \cref{App:tab:ogbn_exp} that \pgnns outperform MLP, GCN, JKNet, and DeepGCN. The performance of \pgnns can be further improved by combining it with C\&S model and residual units and $^{3}$GNN + residual (6 layers) + C\&S obtains the best performance against the baselines.

\subsection{Running Time of \pgnns}
\cref{App:tab:time_homo,App:tab:time_heter} report the averaged running time of \pgnns and baselines on homophilic and heterophilic benchmark datasets, respectively.

\begin{table}[htp]
	\centering
	\caption{Efficiency on homophilic benchmark datasests. Averaged running time per epoch (ms) / averaged total running time (s). OOM denotes out of memory.}\label{App:tab:time_homo}
	\vspace{-5pt}
	\tiny
	\setlength\tabcolsep{2pt}
	\begin{tabular}{ccccccccc}
		\toprule
		Method & Cora & CiteSeer & PubMed & Computers & Photo & CS & Physics \\
		\midrule
		MLP & $7.7$ ms / $5.27$s & $8.1$ ms / $5.37$s & $7.8$ ms / $5.52$s & $8.8$ ms / $5.45$s & $8.4$ ms / $5.34$s & $10.5$ ms / $8.18$s & $14.6$ ms / $12.78$s \\
        GCN & $82.2$ ms / $6.1$s & $84.2$ ms / $6.1$s & $85$ ms / $6.13$s & $85.2$ ms / $7.07$s & $83.6$ ms / $6.08$s & $85$ ms / $9.68$s & $90$ ms / $13.8$s \\
        SGC & $89.5$ ms / $4.96$s & $74.7$ ms / $4.86$s & $80.6$ ms / $5.28$s & $109$ ms / $5.21$s & $85.9$ ms / $4.96$s & $213.6$ ms / $8.01$s & OOM \\
        GAT & $534.8$ ms / $13.06$s & $313.6$ ms / $13.36$s & $314.6$ ms / $13.97$s & $441.3$ ms / $24.62$s & $309.8$ ms / $15.96$s & $454$ ms / $21.87$s & $436.9$ ms / $40.9$s \\
        JKNet & $95.4$ ms / $20.07$s & $101.1$ ms / $19.58$s & $105.4$ ms / $20.8$s & $106.1$ ms / $29.72$s & $97.9$ ms / $21.18$s & $102.7$ ms / $24.94$s & $119.2$ ms / $40.83$s \\
        APPNP & $86.7$ ms / $11.6$s & $86.3$ ms / $11.98$s & $85.5$ ms / $11.97$s & $92.1$ ms / $15.75$s & $86$ ms / $12.19$s & $90.5$ ms / $17.36$s & $99.6$ ms / $25.89$s \\
        GPRGNN & $86.5$ ms / $12.42$s & $195.8$ ms / $12.6$s & $88.6$ ms / $12.59$s & $93.3$ ms / $15.98$s & $86.7$ ms / $12.65$s & $92$ ms / $17.8$s & $217.1$ ms / $26.33$s \\
        $^{1.0}$GNN & $96$ ms / $20.12$s & $98.1$ ms / $19.81$s & $100.2$ ms / $21.74$s & $151.4$ ms / $64.08$s & $121.3$ ms / $34.07$s & $109.7$ ms / $25.03$s & $122.9$ ms / $49.59$s \\
        $^{1.5}$GNN & $98.2$ ms / $20.19$s & $97$ ms / $20.26$s & $100.2$ ms / $22.6$s & $140.3$ ms / $64.08$s & $120$ ms / $34.22$s & $112.3$ ms / $25.11$s & $127.9$ ms / $49.54$s \\
        $^{2.0}$GNN & $98.1$ ms / $20.11$s & $96.3$ ms / $19.97$s & $99.3$ ms / $22.17$s & $141$ ms / $64.04$s & $129.3$ ms / $34.14$s & $104.7$ ms / $24.93$s & $124.6$ ms / $49.35$s \\
        $^{2.5}$GNN & $96.6$ ms / $20.12$s & $92.9$ ms / $20.16$s & $103$ ms / $22.17$s & $141.6$ ms / $64.01$s & $128.1$ ms / $34.22$s & $110.8$ ms / $25.07$s & $124$ ms / $49.39$s \\
		\bottomrule
	\end{tabular}
\end{table}

\begin{table}[htp]
	\centering
	\caption{Efficiency on heterophilic benchmark datasests. Averaged running time per epoch (ms) / averaged total running time (s).}\label{App:tab:time_heter}
	\vspace{-5pt}
	\tiny
	\setlength\tabcolsep{3pt}
	\begin{tabular}{cccccccc}
		\toprule
		Method & Chameleon & Squirrel & Actor & Wisconsin & Texas & Cornell \\
		\midrule
		MLP & $7.7$ ms / $5.29$s & $8$ ms / $5.44$s & $8.6$ ms / $5.4$s & $7.7$ ms / $5.16$s & $7.9$ ms / $5.22$s & $7.6$ ms / $5.19$s \\
        GCN & $83.4$ ms / $6.1$s & $83.2$ ms / $6.2$s & $90.7$ ms / $6.07$s & $83.5$ ms / $5.94$s & $80.7$ ms / $5.96$s & $87.1$ ms / $5.92$s \\
        SGC & $78.1$ ms / $4.93$s & $110.9$ ms / $5.21$s & $77.1$ ms / $4.71$s & $73.2$ ms / $4.52$s & $74.2$ ms / $4.79$s & $71.3$ ms / $4.8$s \\
        GAT & $374.9$ ms / $13.49$s & $324.2$ ms / $17.15$s & $420$ ms / $13.82$s & $317.5$ ms / $12.68$s & $357.9$ ms / $12.38$s & $383.3$ ms / $12.45$s \\
        JKNet & $102.4$ ms / $21.15$s & $101$ ms / $22.84$s & $97.2$ ms / $21.24$s & $98.5$ ms / $21.07$s & $103.6$ ms / $20.92$s & $102.2$ ms / $20.79$s \\
        APPNP & $87.1$ ms / $12.12$s & $98.8$ ms / $12.41$s & $87.2$ ms / $11.81$s & $84.2$ ms / $11.83$s & $86$ ms / $11.9$s & $83.1$ ms / $11.94$s \\
        GPRGNN & $93$ ms / $12.98$s & $86.1$ ms / $13.01$s & $94.2$ ms / $13.01$s & $84.3$ ms / $12.66$s & $92$ ms / $12.64$s & $89.1$ ms / $12.6$s \\
        $^{1.0}$GNN & $107.3$ ms / $22.43$s & $116.3$ ms / $30.92$s & $117.8$ ms / $23.6$s & $94.5$ ms / $18.47$s & $92$ ms / $18.83$s & $92.7$ ms / $18.97$s \\
        $^{1.5}$GNN & $97.2$ ms / $22.54$s & $115$ ms / $31.04$s & $119.2$ ms / $23.47$s & $93.3$ ms / $18.64$s & $90.8$ ms / $19.09$s & $94.9$ ms / $18.88$s \\
        $^{2.0}$GNN & $98.7$ ms / $22.37$s & $114.8$ ms / $31.14$s & $100.8$ ms / $23.73$s & $92.2$ ms / $19.09$s & $92.5$ ms / $18.72$s & $98$ ms / $18.64$s \\
        $^{2.5}$GNN & $97.9$ ms / $22.38$s & $115.9$ ms / $31.09$s & $97.3$ ms / $23.77$s & $92.8$ ms / $19.03$s & $91$ ms / $18.84$s & $90.7$ ms / $18.83$s \\
		\bottomrule
	\end{tabular}
\end{table}

\subsection{Experimental Results on Benchmark Datasets for 64 Hidden Units}
\begin{table}[htp]
	\centering
	\caption{Results on heterophilic benchmark datasets for 64 hidden units. Averaged accuracy (\%) for 20 runs. Best results outlined in bold and the results within $95\%$ confidence interval of the best results are outlined in underlined bold.}
	\vspace{-5pt}
	\footnotesize
	\setlength\tabcolsep{2pt}
	\begin{tabular}{ccccccc}
		\toprule
		Method & Chameleon & Squirrel & Actor & Wisconsin & Texas & Cornell \\
		\midrule
		MLP & $46.55_{\pm 0.90}$ & $\bm{33.83}_{\pm 0.59}$ & $38.40_{\pm 0.76}$ & $93.91_{\pm 2.47}$ & $\bm{\underline{87.51}}_{\pm 8.53}$ & $\bm{\underline{86.75}}_{\pm 8.22}$ \\
		GCN & $34.74_{\pm 2.62}$ & $25.68_{\pm 1.17}$ & $30.86_{\pm 1.51}$ & $65.93_{\pm 5.47}$ & $58.56_{\pm 13.28}$ & $46.81_{\pm 4.28}$ \\
		SGC & $34.57_{\pm 4.71}$ & $24.39_{\pm 1.54}$ & $35.50_{\pm 2.09}$ & $62.87_{\pm 8.92}$ & $50.62_{\pm 5.60}$ & $29.44_{\pm 14.83}$ \\
		GAT & $43.33_{\pm 1.53}$ & $30.07_{\pm 0.99}$ & $33.44_{\pm 2.45}$ & $66.57_{\pm 4.69}$ & $50.69_{\pm 12.89}$ & $42.62_{\pm 13.37}$ \\
		JKNet & $32.69_{\pm 4.47}$ & $27.18_{\pm 0.76}$ & $25.72_{\pm 2.75}$ & $66.57_{\pm 10.53}$ & $43.88_{\pm 17.10}$ & $47.69_{\pm 3.25}$ \\
		APPNP & $35.09_{\pm 3.18}$ & $28.15_{\pm 0.93}$ & $32.28_{\pm 1.75}$ & $66.30_{\pm 1.60}$ & $69.00_{\pm 4.53}$ & $54.88_{\pm 3.85}$ \\
		GPRGNN & $34.65_{\pm 2.86}$ & $28.56_{\pm 1.35}$ & $34.58_{\pm 1.45}$ & $93.70_{\pm 3.12}$ & $86.50_{\pm 6.04}$ & $84.75_{\pm 8.38}$ \\
		\midrule
		$^{1.0}$GNN & $\bm{\underline{49.51}}_{\pm 1.32}$ & $32.67_{\pm 1.00}$ & $\bm{40.70}_{\pm 0.88}$ & $\bm{95.23}_{\pm 1.60}$ & $84.12_{\pm 7.39}$ & $82.56_{\pm 6.97}$ \\
		$^{1.5}$GNN & $\bm{49.52}_{\pm 1.15}$ & $33.14_{\pm 1.10}$ & $39.82_{\pm 1.54}$ & $94.03_{\pm 2.26}$ & $86.94_{\pm 6.99}$ & $\bm{86.89}_{\pm 6.63}$ \\
		$^{2.0}$GNN & $49.19_{\pm 0.81}$ & $\bm{\underline{33.78}}_{\pm 0.87}$ & $39.75_{\pm 1.26}$ & $94.49_{\pm 1.81}$ & $\bm{87.62}_{\pm 6.64}$ & $85.56_{\pm 7.25}$ \\
		$^{2.5}$GNN & $48.93_{\pm 0.74}$ & $33.31_{\pm 1.27}$ & $39.47_{\pm 1.20}$ & $92.13_{\pm 2.16}$ & $\bm{\underline{87.25}}_{\pm 5.57}$ & $80.56_{\pm 5.28}$ \\
		\bottomrule
	\end{tabular}
\end{table}
\newpage
\begin{table}[htp]
	\centering
	\caption{Results on homophilic benchmark datasets for 64 hidden units. Averaged accuracy (\%) for 20 runs. Best results are outlined in bold and the results within $95\%$ confidence interval of the best results are outlined in underlined bold. OOM denotes out of memory.}
	\vspace{-5pt}
	\footnotesize
	\setlength\tabcolsep{2pt}
	\begin{tabular}{cccccccc}
		\toprule
		Method & Cora & CiteSeer & PubMed & Computers & Photo & CS & Physics \\
		\midrule
		MLP & $49.05_{\pm 0.82}$ & $50.67_{\pm 1.25}$ & $80.32_{\pm 0.40}$ & $70.58_{\pm 0.82}$ & $79.44_{\pm 0.79}$ & $89.48_{\pm 0.50}$ & $92.84_{\pm 0.62}$ \\
		GCN & $77.65_{\pm 0.42}$ & $64.72_{\pm 0.52}$ & $84.13_{\pm 0.12}$ & $\bm{\underline{84.56}}_{\pm 0.79}$ & $90.16_{\pm 0.88}$ & $91.14_{\pm 0.10}$ & $94.75_{\pm 0.04}$ \\
		SGC & $70.32_{\pm 1.87}$ & $\bm{65.77}_{\pm 0.99}$ & $76.27_{\pm 0.94}$ & $83.24_{\pm 0.81}$ & $89.43_{\pm 1.03}$ & $91.11_{\pm 0.10}$ & OOM \\
		GAT & $76.97_{\pm 1.18}$ & $61.28_{\pm 1.62}$ & $83.57_{\pm 0.23}$ & $83.84_{\pm 1.93}$ & $\bm{\underline{90.54}}_{\pm 0.56}$ & $89.68_{\pm 0.42}$ & $93.91_{\pm 0.20}$ \\
		JKNet & $78.77_{\pm 0.79}$ & $64.62_{\pm 0.80}$ & $82.82_{\pm 0.16}$ & $82.22_{\pm 1.32}$ & $88.43_{\pm 0.53}$ & $90.48_{\pm 0.13}$ & $93.75_{\pm 0.32}$ \\
		APPNP & $\bm{79.95}_{\pm 0.72}$ & $\bm{\underline{65.56}}_{\pm 0.64}$ & $84.00_{\pm 0.22}$ & $83.83_{\pm 0.78}$ & $\bm{\underline{90.50}}_{\pm 0.59}$ & $91.90_{\pm 0.12}$ & $\bm{\underline{94.84}}_{\pm 0.08}$ \\
		GPRGNN & $78.17_{\pm 1.31}$ & $61.26_{\pm 2.14}$ & $84.54_{\pm 0.24}$ & $83.77_{\pm 1.06}$ & $89.86_{\pm 0.63}$ & $91.34_{\pm 0.25}$ & $94.63_{\pm 0.26}$ \\
		\midrule
		$^{1.0}$GNN & $77.11_{\pm 0.39}$ & $63.17_{\pm 0.89}$ & $83.14_{\pm 0.46}$ & $82.64_{\pm 0.98}$ & $89.60_{\pm 0.69}$ & $92.53_{\pm 0.22}$ & $\bm{\underline{94.86}}_{\pm 0.24}$ \\
		$^{1.5}$GNN & $78.69_{\pm 0.43}$ & $63.14_{\pm 0.93}$ & $83.97_{\pm 0.04}$ & $\bm{84.64}_{\pm 1.42}$ & $\bm{90.67}_{\pm 0.67}$ & $\bm{92.93}_{\pm 0.14}$ & $\bm{\underline{94.93}}_{\pm 0.12}$ \\
		$^{2.0}$GNN & $79.06_{\pm 0.41}$ & $63.92_{\pm 1.14}$ & $84.24_{\pm 0.27}$ & $\bm{\underline{84.57}}_{\pm 0.96}$ & $90.17_{\pm 0.88}$ & $\bm{\underline{92.74}}_{\pm 0.26}$ & $\bm{95.05}_{\pm 0.09}$ \\
		$^{2.5}$GNN & $79.15_{\pm 0.39}$ & $63.16_{\pm 1.25}$ & $\bm{84.88}_{\pm 0.09}$ & $83.84_{\pm 0.71}$ & $89.05_{\pm 0.85}$ & $92.31_{\pm 0.19}$ & $94.92_{\pm 0.10}$ \\
		\bottomrule
	\end{tabular}
\end{table}

\subsection{Training Curves for $p=1$}
\begin{figure}[htp]
	\centering
	\begin{subfigure}[b]{\textwidth}
		\centering
        \includegraphics[width=0.24\textwidth]{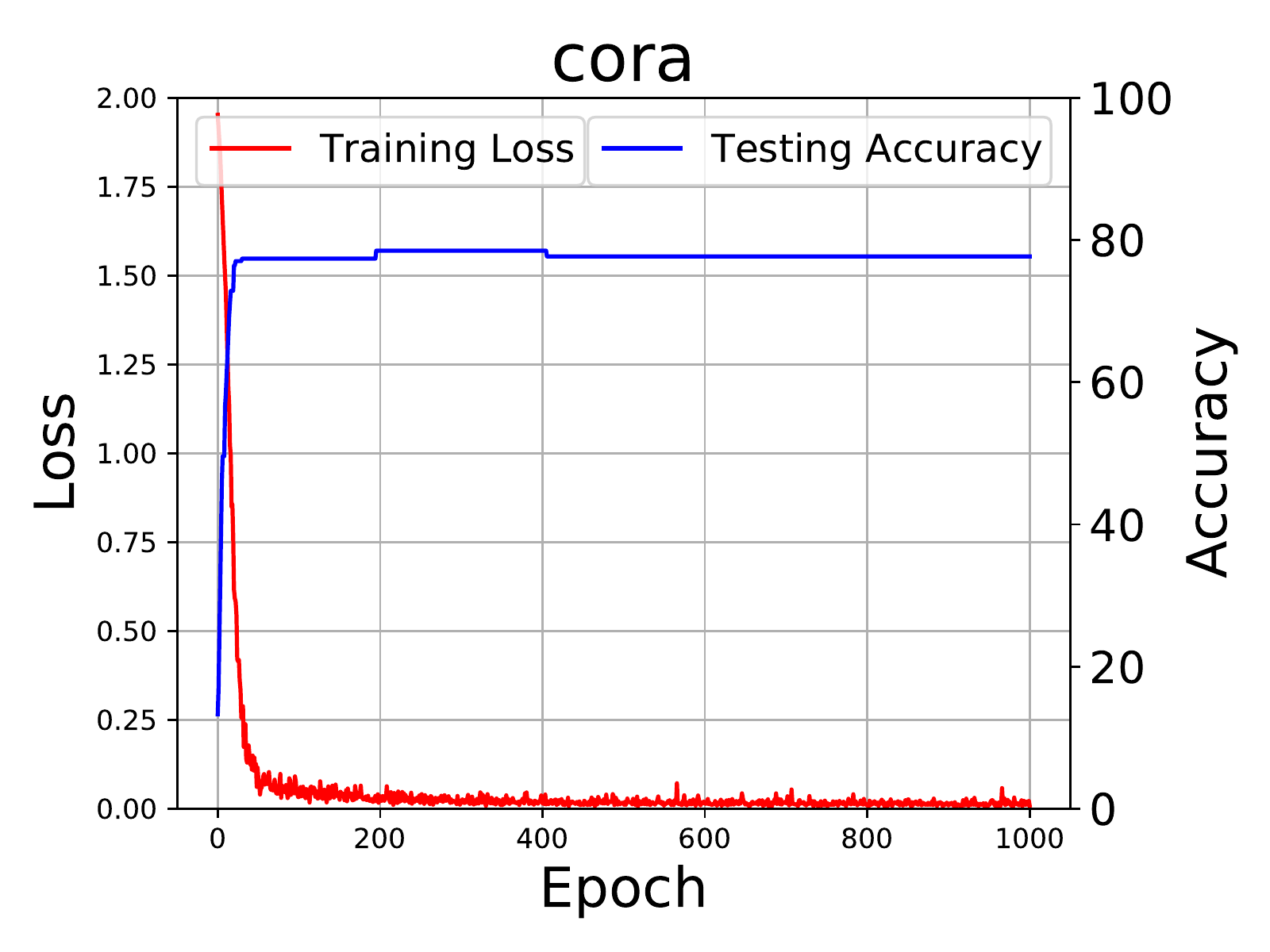}
        \includegraphics[width=0.24\textwidth]{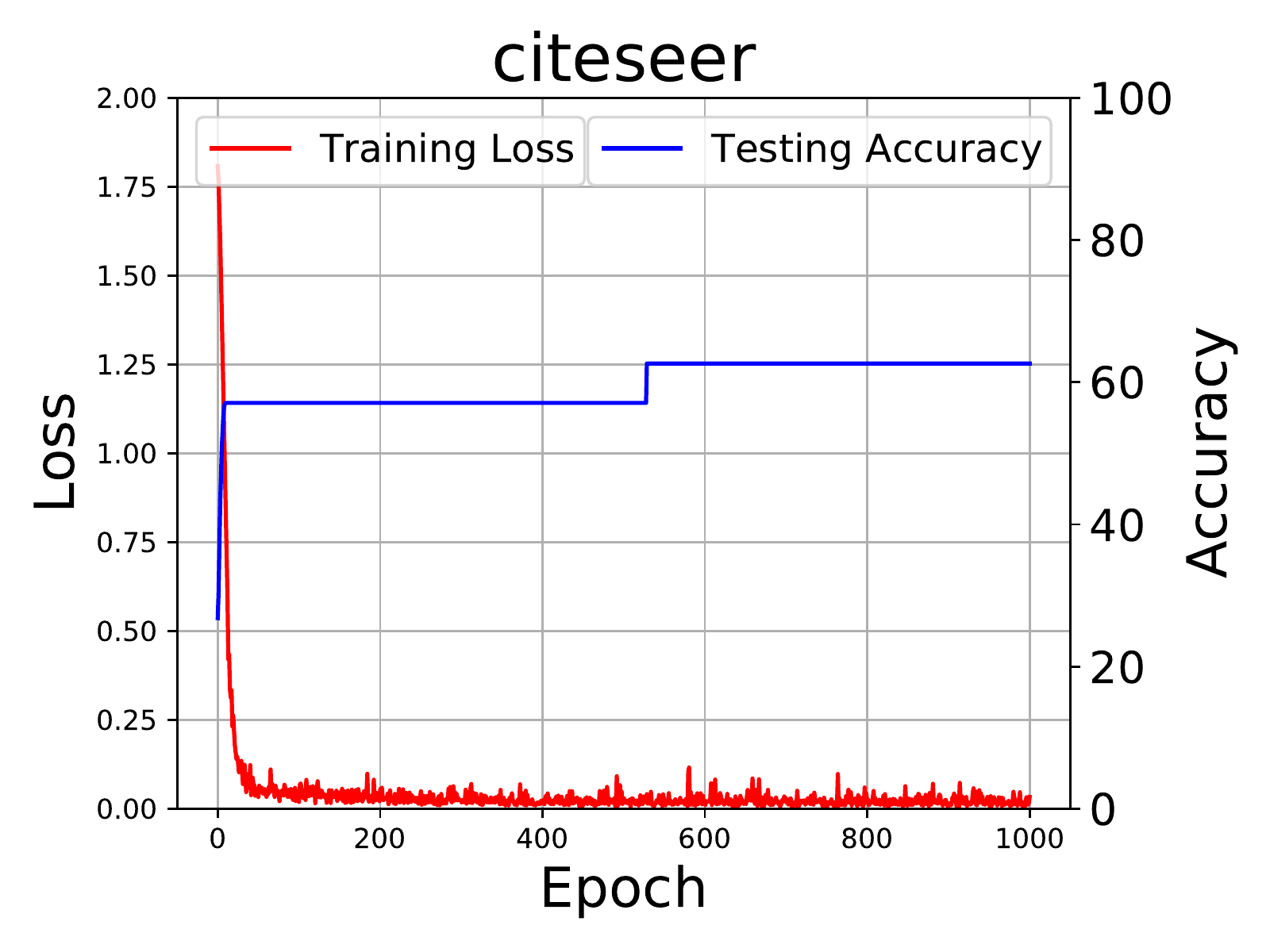}
        \includegraphics[width=0.24\textwidth]{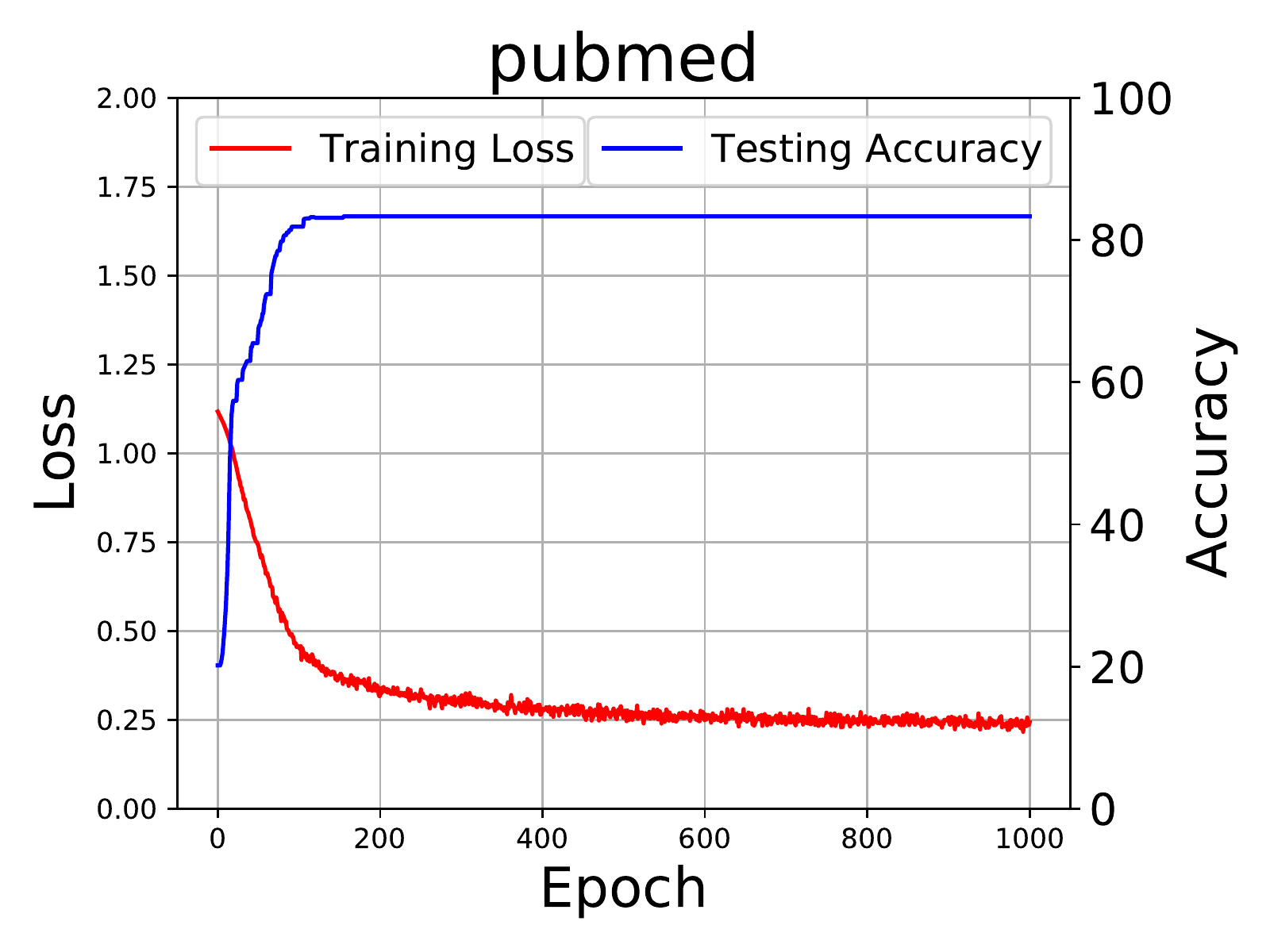}
        \includegraphics[width=0.24\textwidth]{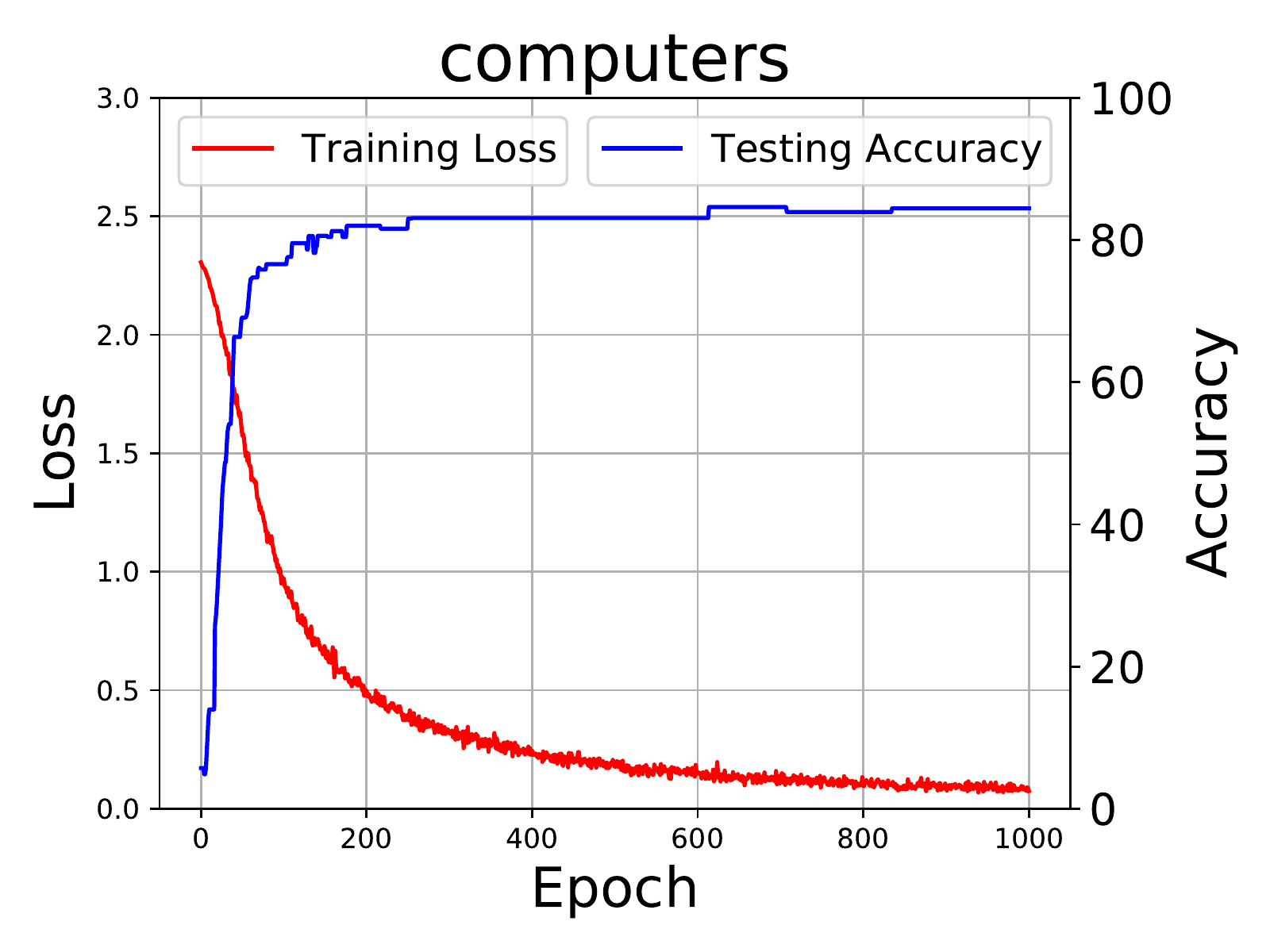}
    \end{subfigure}
    \begin{subfigure}[b]{\textwidth}
		\centering
        \includegraphics[width=0.24\textwidth]{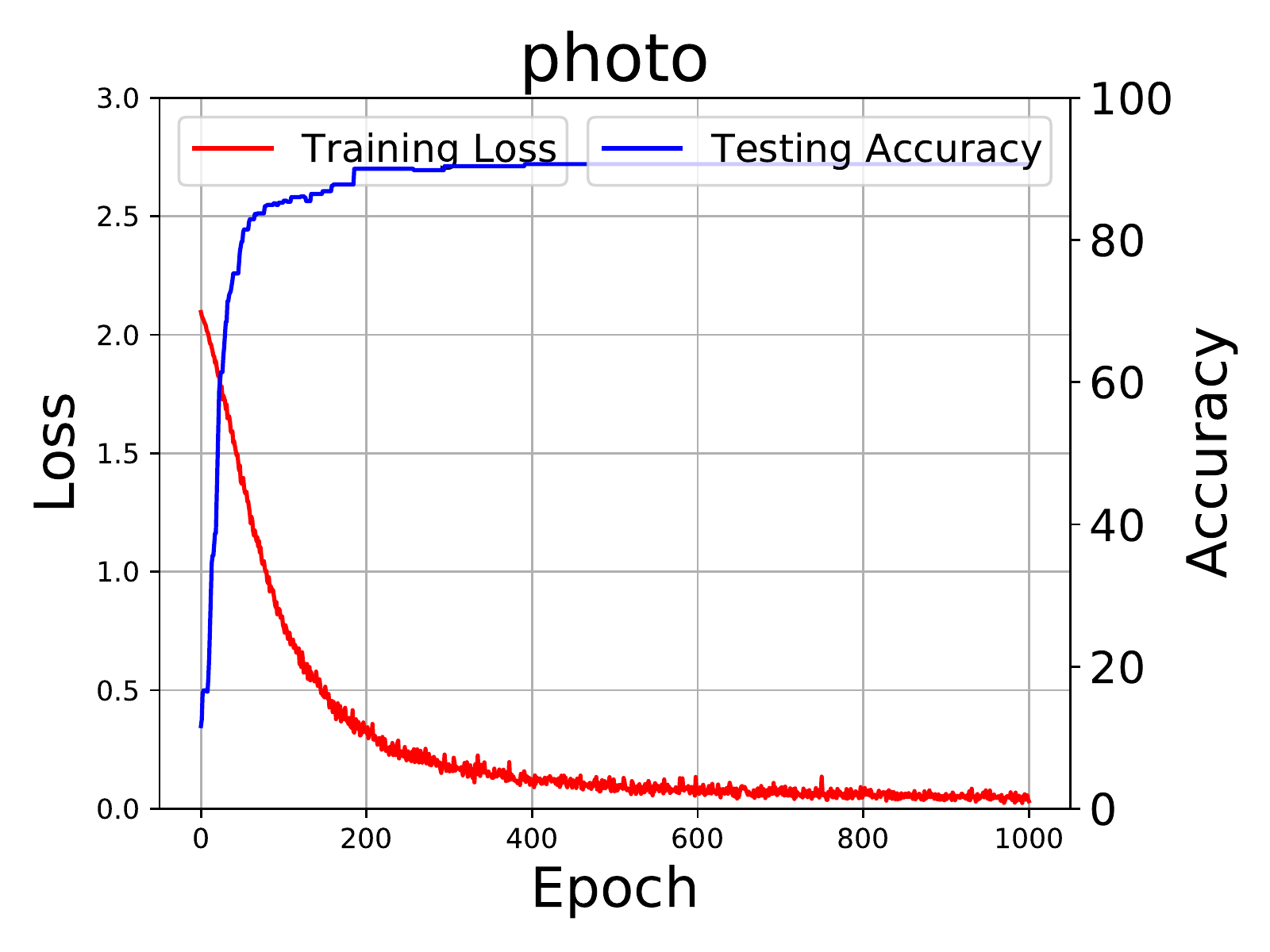}
        \includegraphics[width=0.24\textwidth]{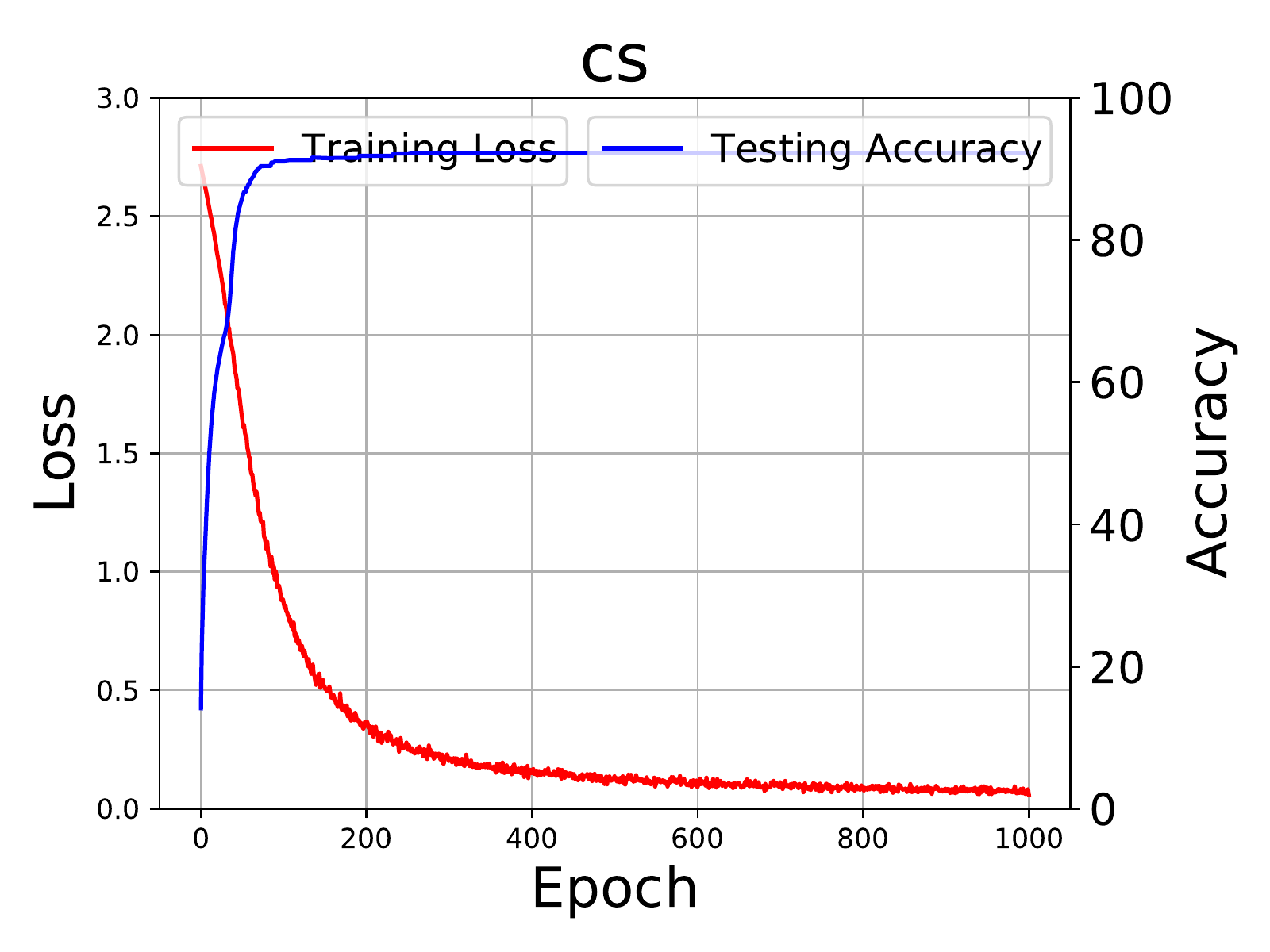}
        \includegraphics[width=0.24\textwidth]{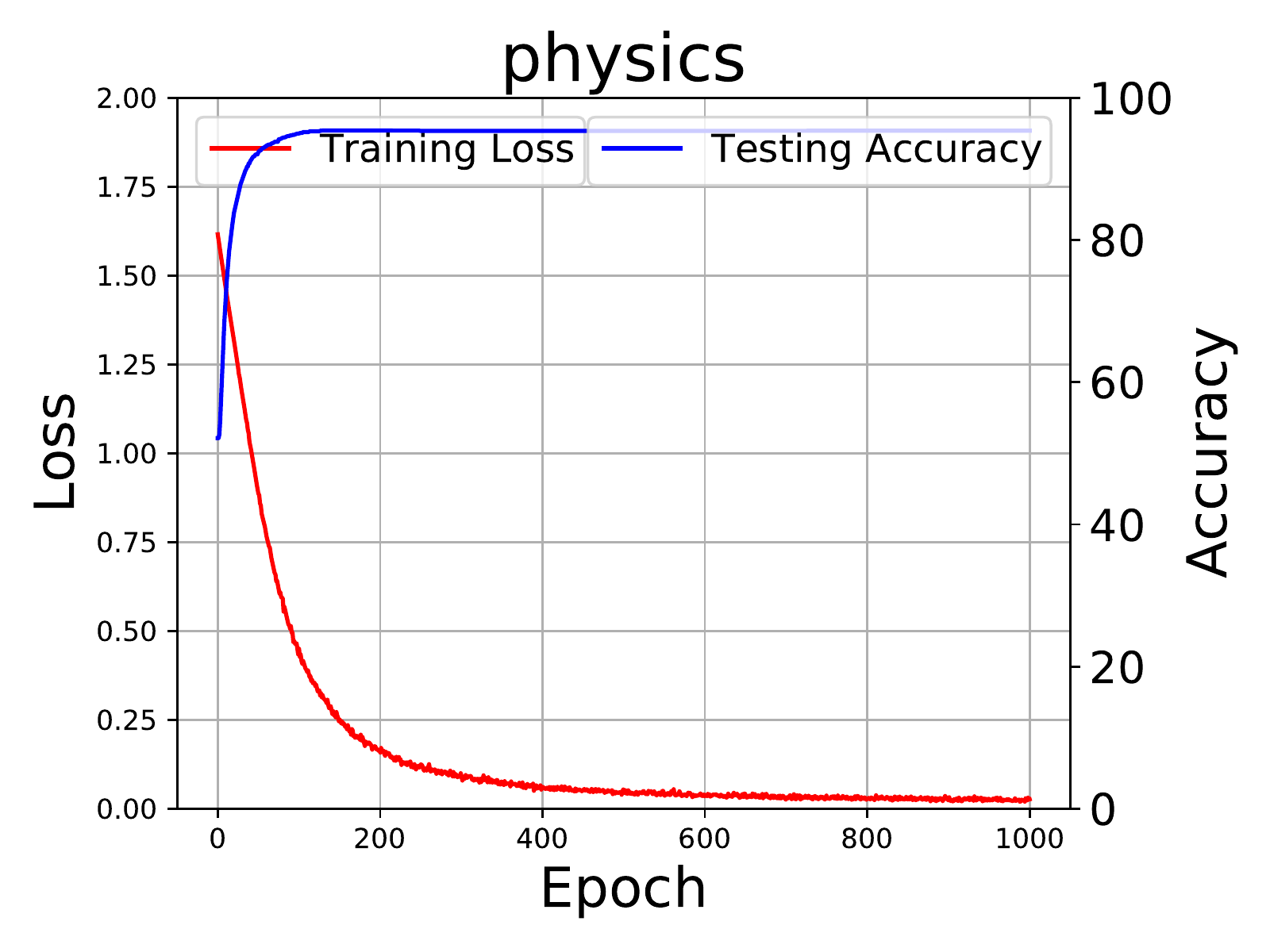}
        \includegraphics[width=0.24\textwidth]{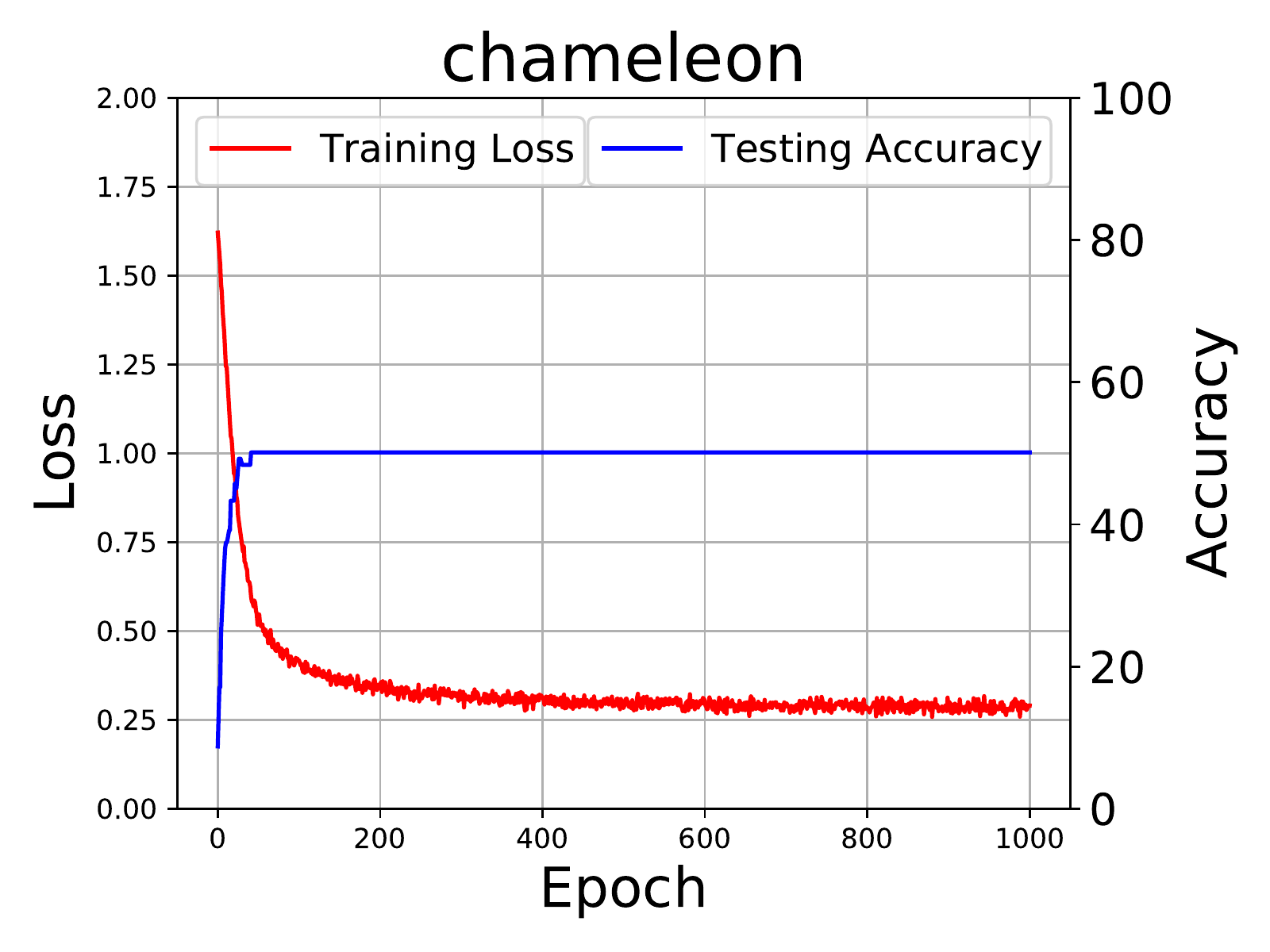}
        \end{subfigure}
    \begin{subfigure}[b]{\textwidth}
		\centering
        \includegraphics[width=0.24\textwidth]{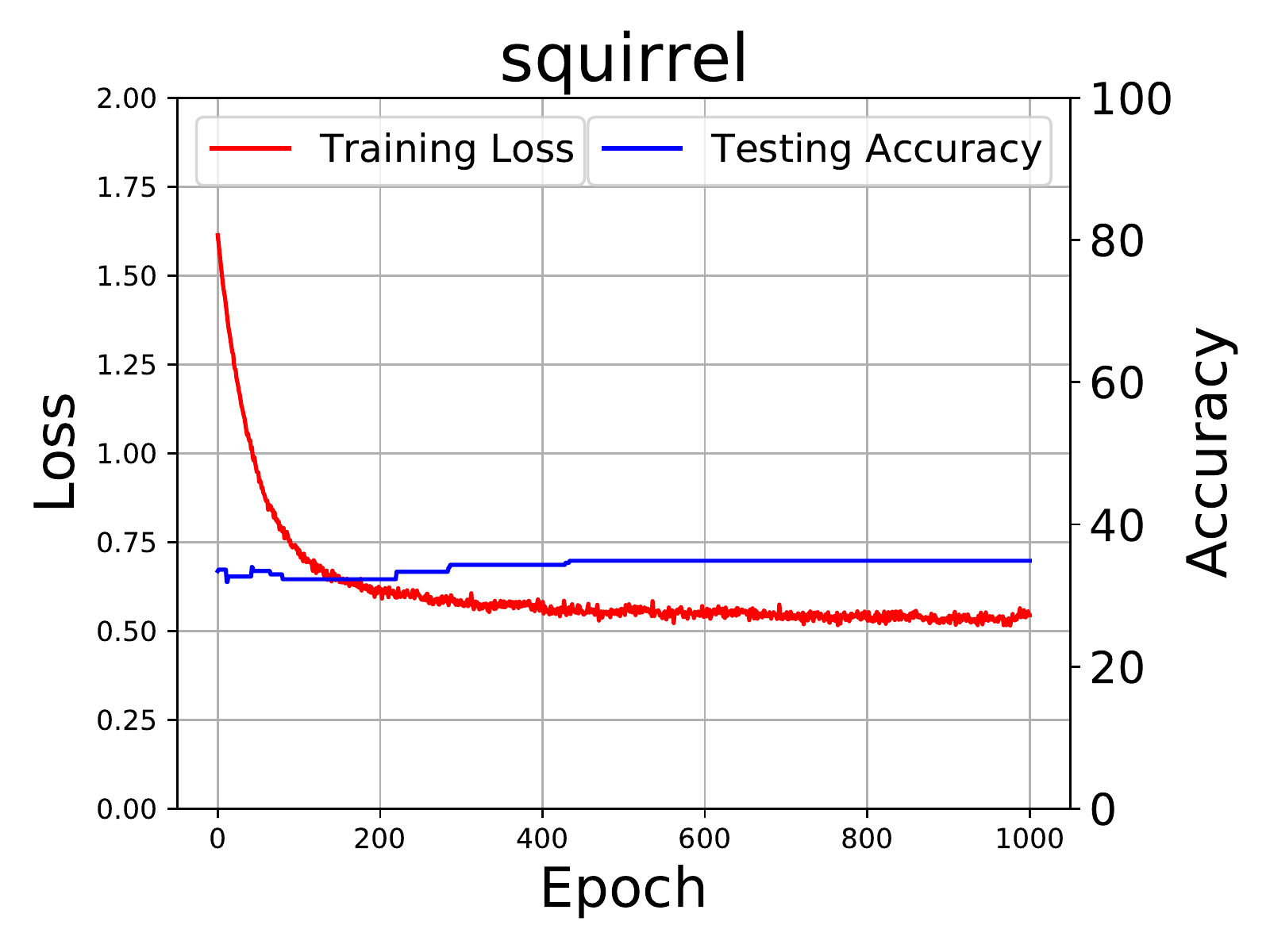}
        \includegraphics[width=0.24\textwidth]{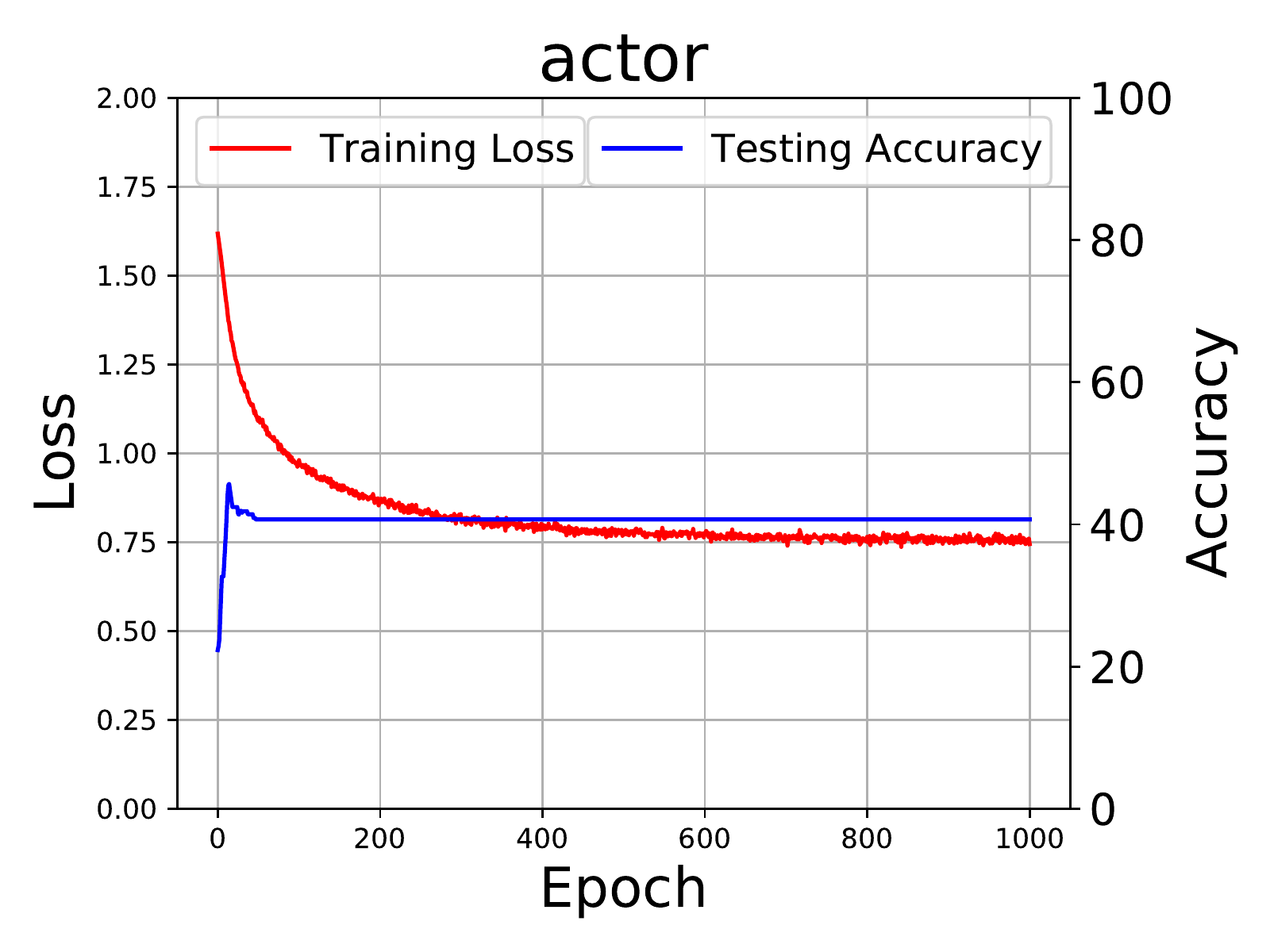}
        \includegraphics[width=0.24\textwidth]{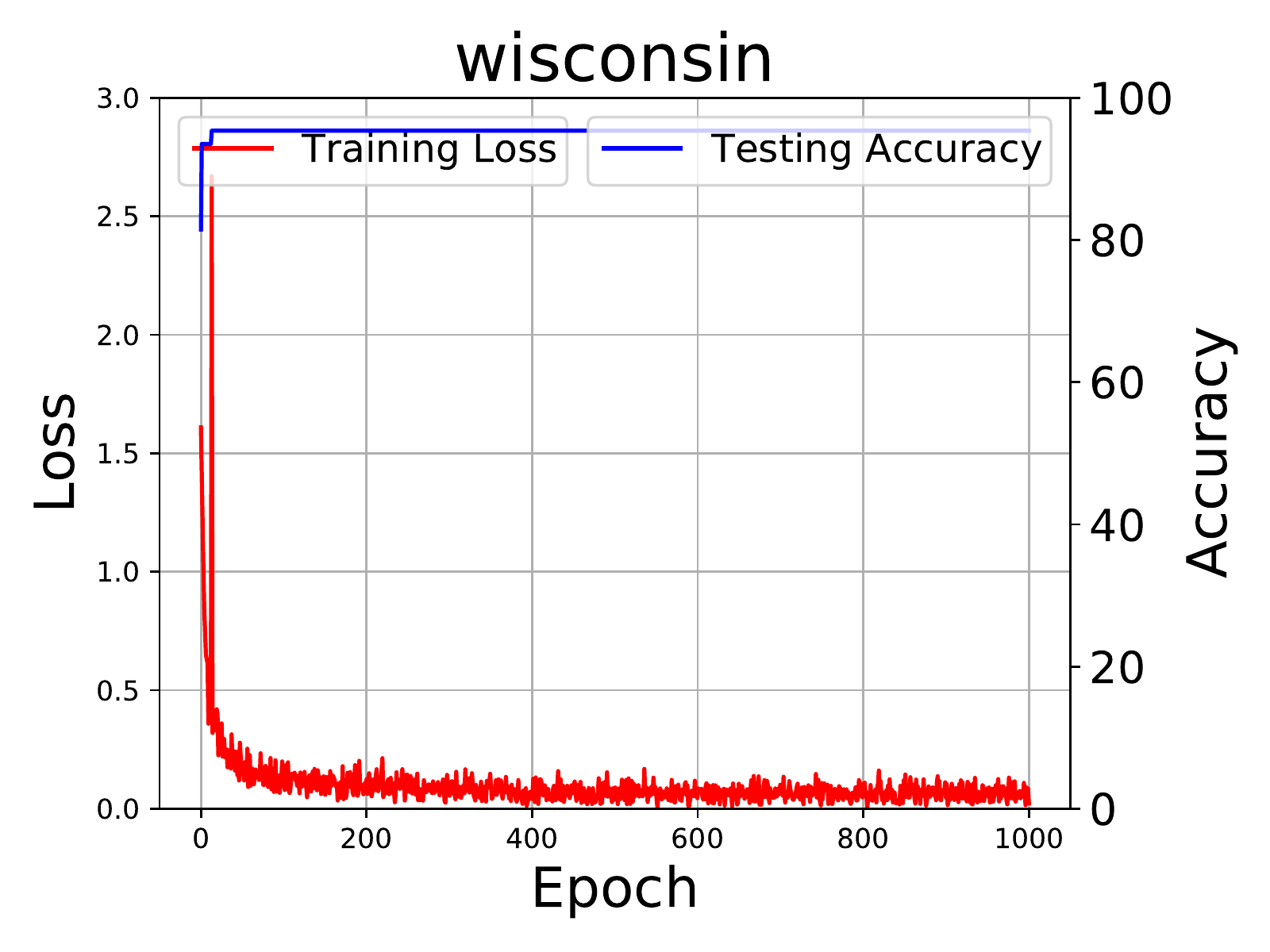}
        \includegraphics[width=0.24\textwidth]{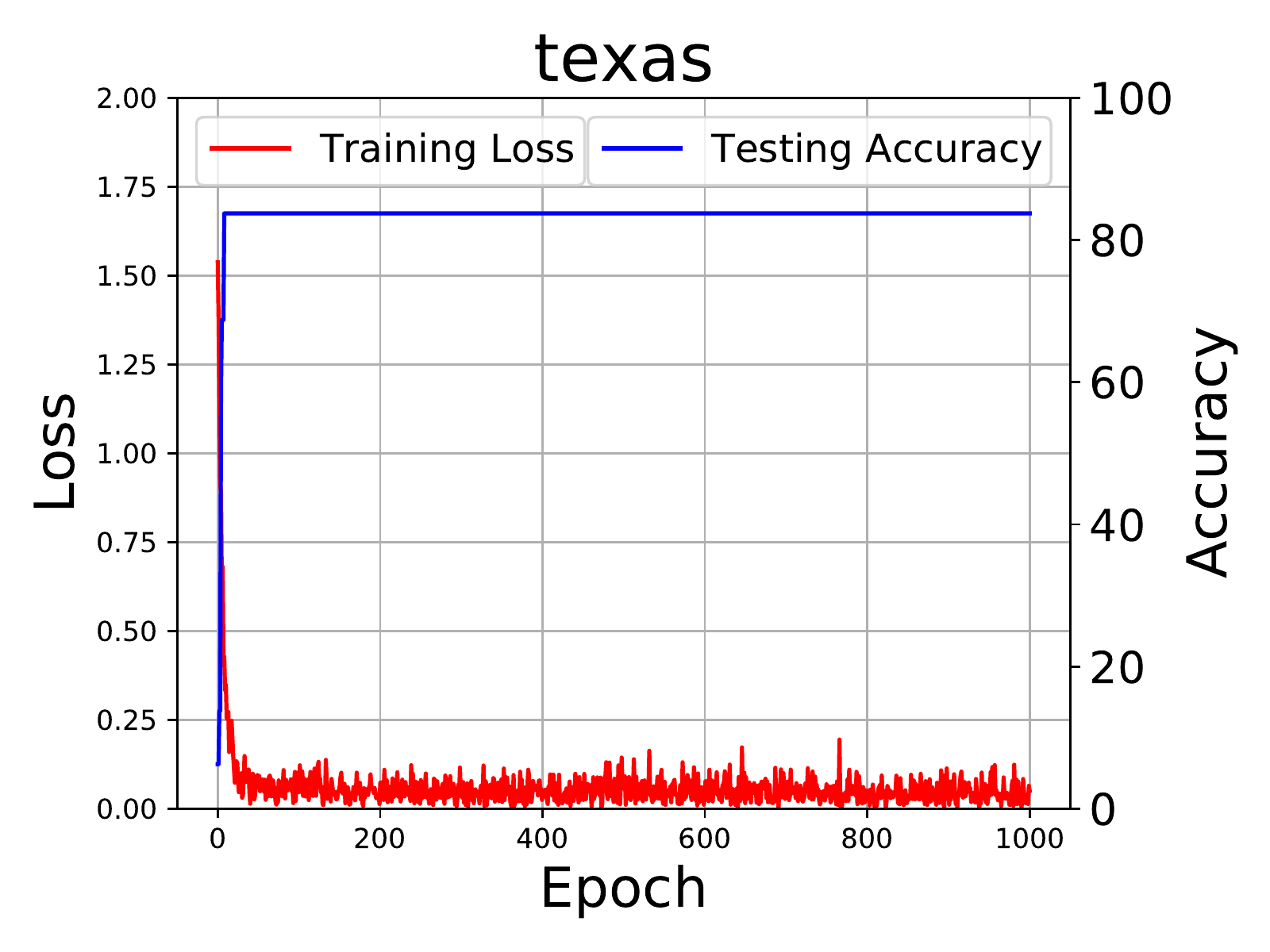}
    \end{subfigure}
    \begin{subfigure}[b]{\textwidth}
		\centering
        \includegraphics[width=0.24\textwidth]{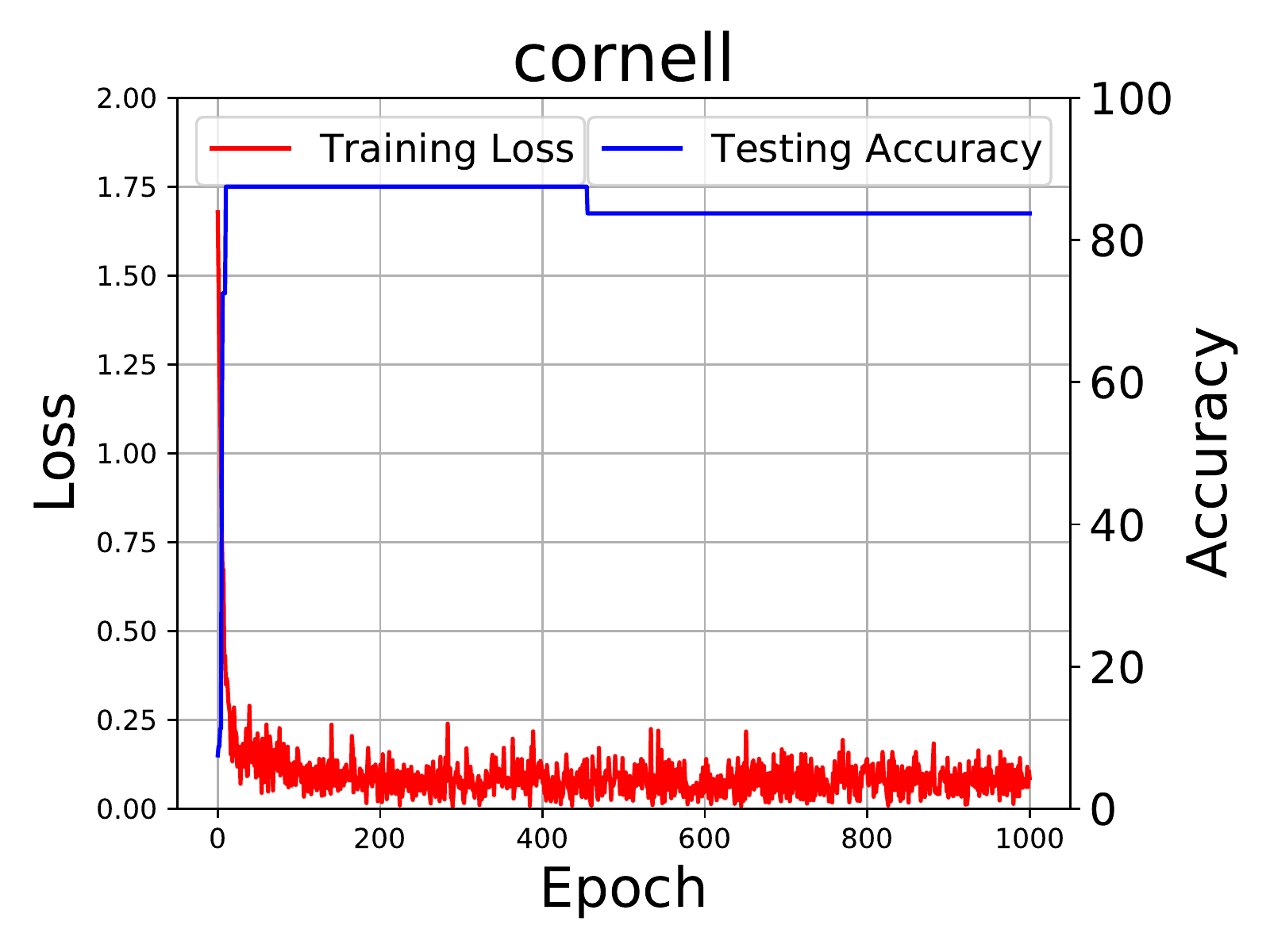}
    \end{subfigure}
	\vspace{-5pt}
	\caption{The curves of training loss and testing accuracy for $p=1$.}
	\vspace{-3.cm}
\end{figure}

\newpage
\subsection{Visualization Results of Node Embeddings}
\begin{figure}[htp]
 	\centering
	\begin{subfigure}[b]{0.28\textwidth}
		\centering
		\includegraphics[width=\textwidth]{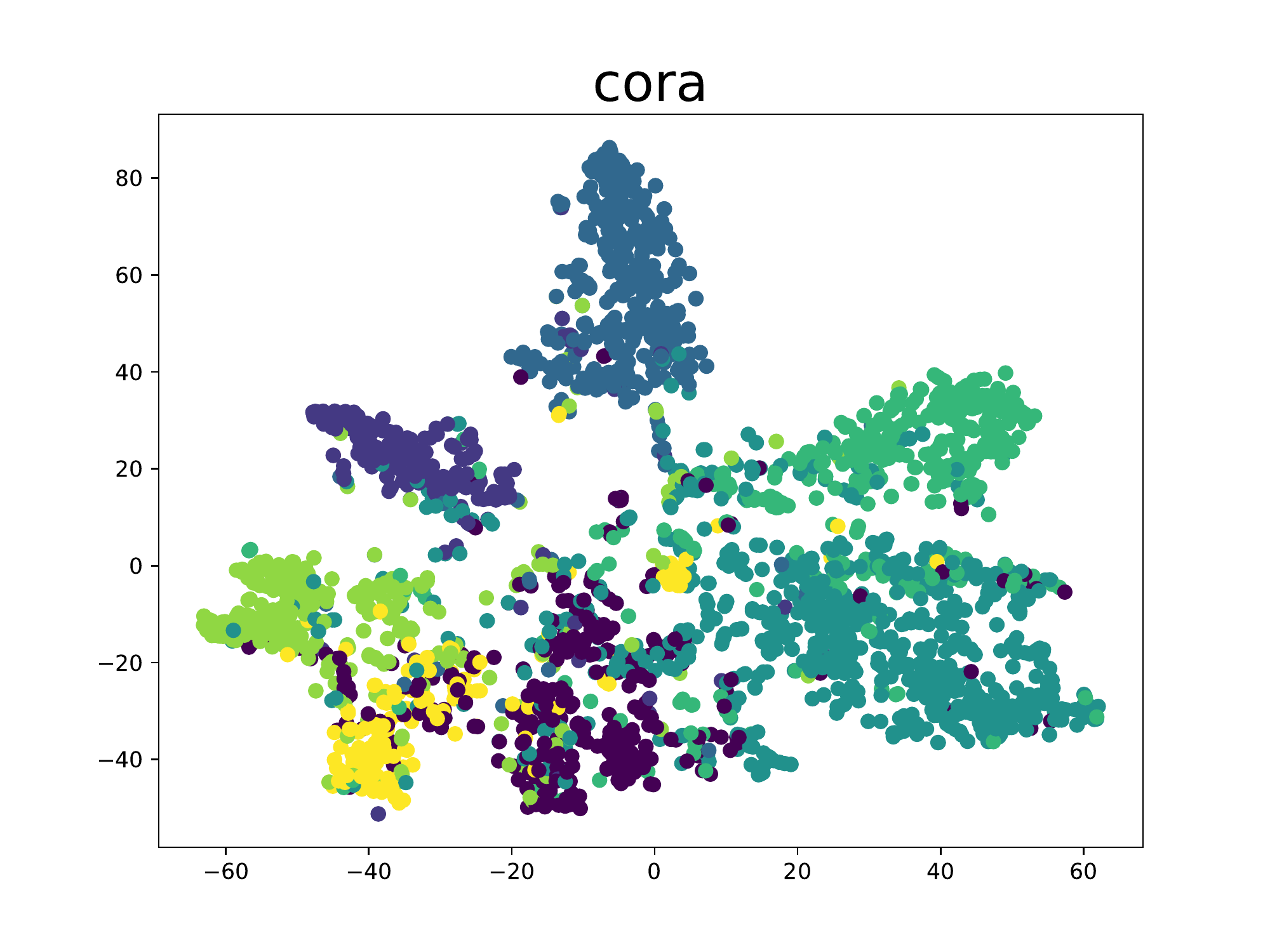}
        \caption{$^{1.0}$GNN.}
	\end{subfigure}
	\begin{subfigure}[b]{0.28\textwidth}
		\centering
		\includegraphics[width=\textwidth]{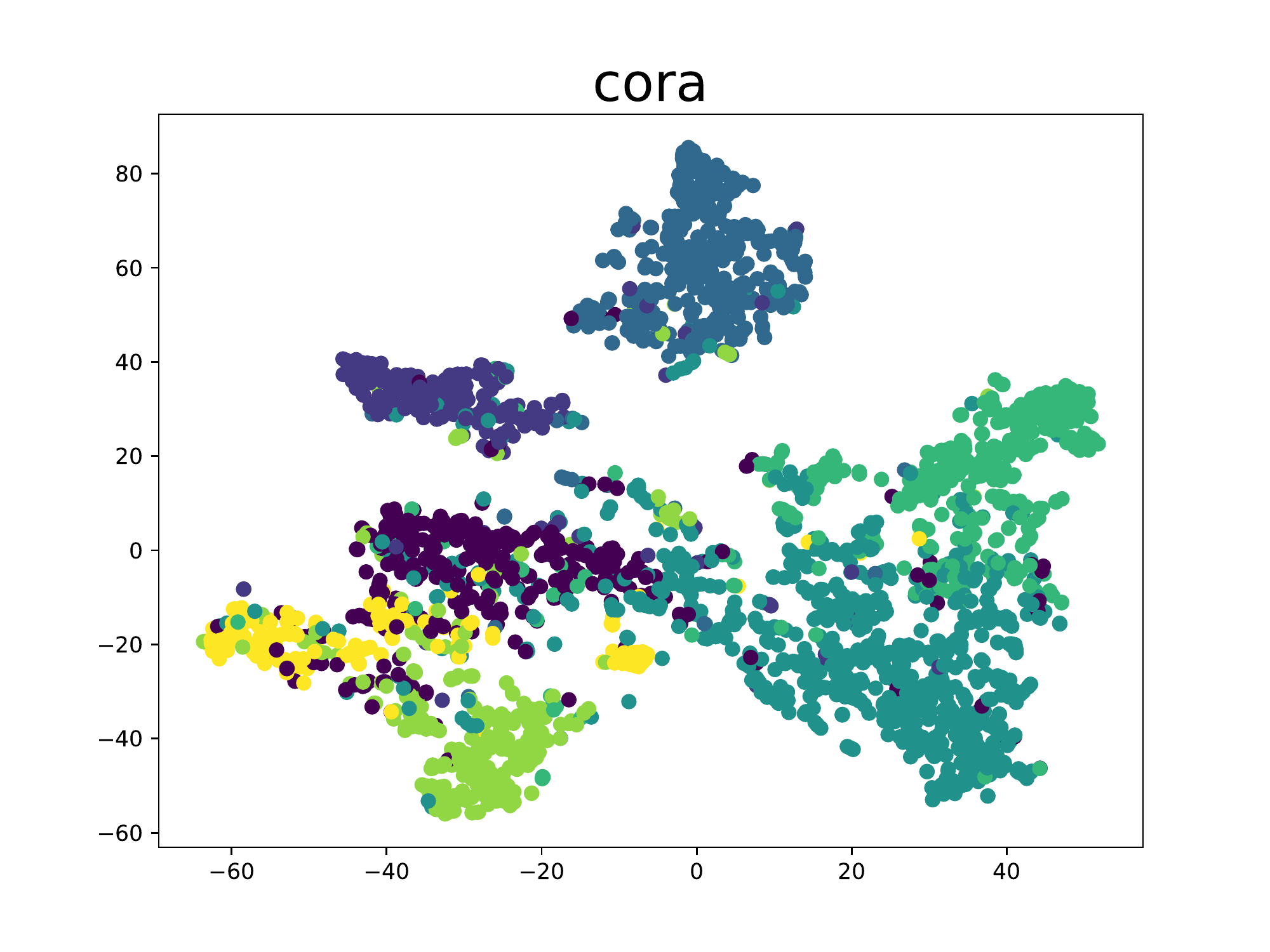}
        \caption{$^{1.5}$GNN.}
	\end{subfigure}
	\begin{subfigure}[b]{0.28\textwidth}
		\centering
		\includegraphics[width=\textwidth]{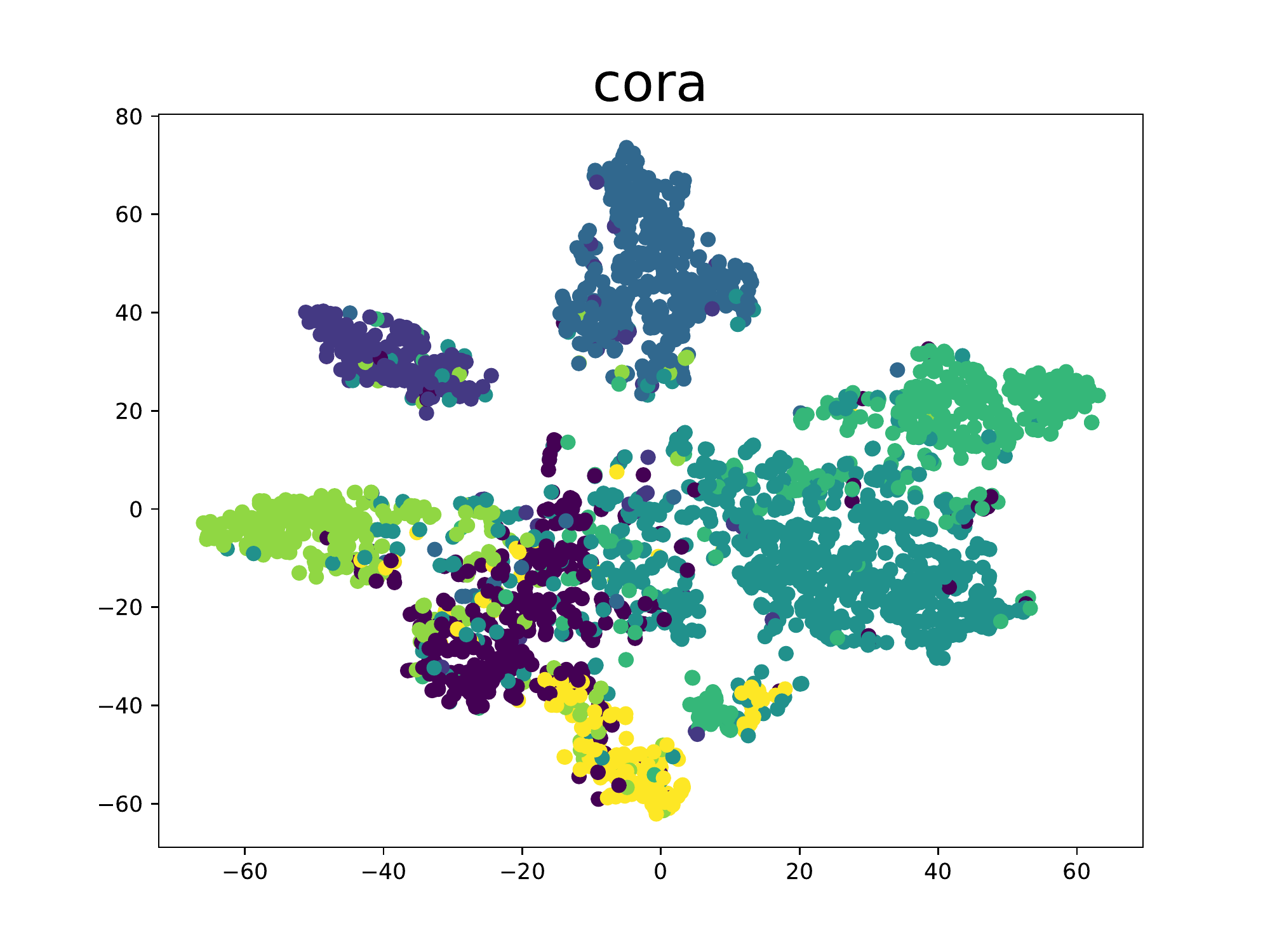}
        \caption{$^{2.0}$GNN.}
	\end{subfigure}
	\begin{subfigure}[b]{0.28\textwidth}
		\centering
		\includegraphics[width=\textwidth]{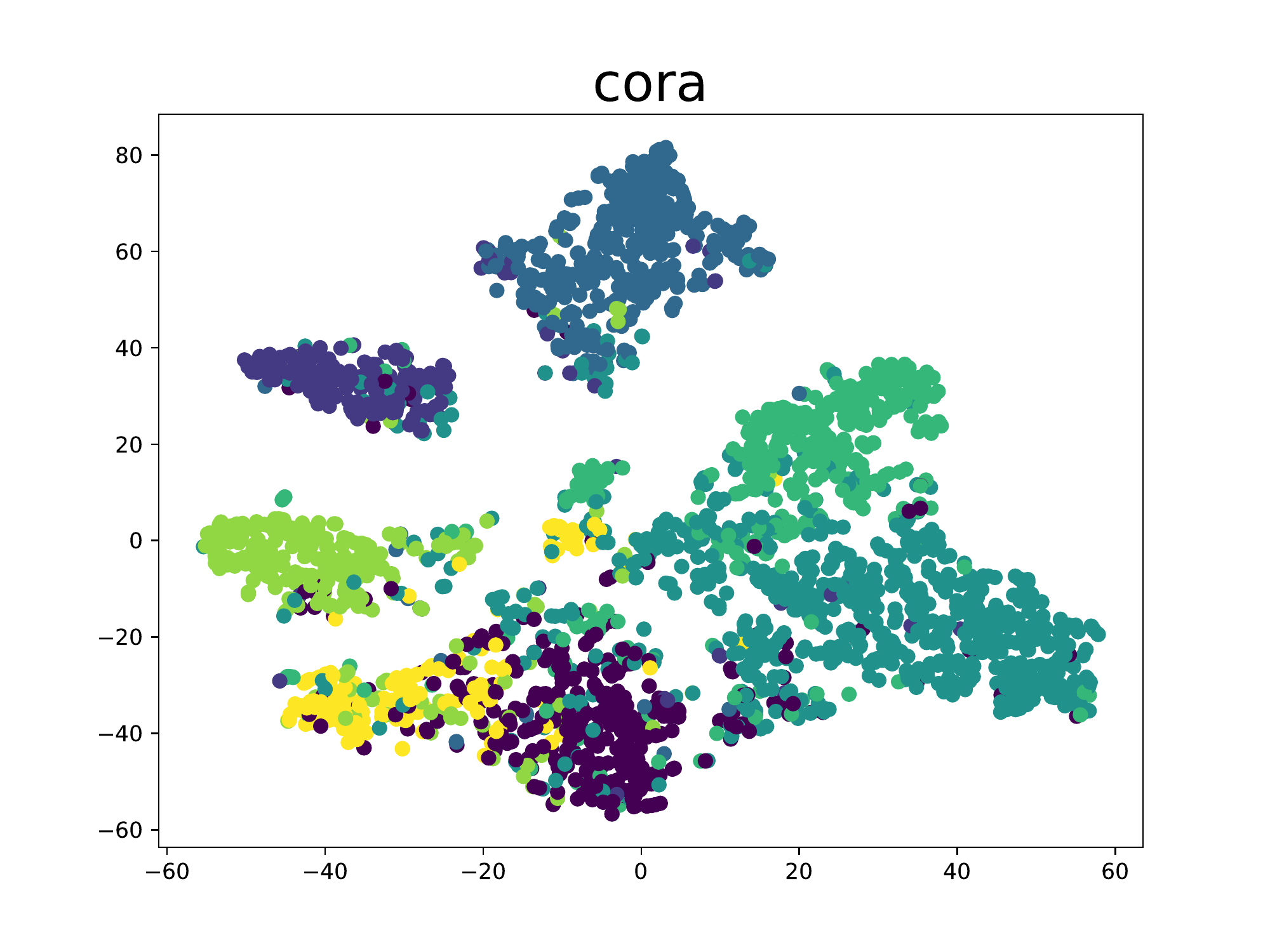}
        \caption{$^{2.5}$GNN.}
	\end{subfigure}
	\begin{subfigure}[b]{0.28\textwidth}
		\centering
		\includegraphics[width=\textwidth]{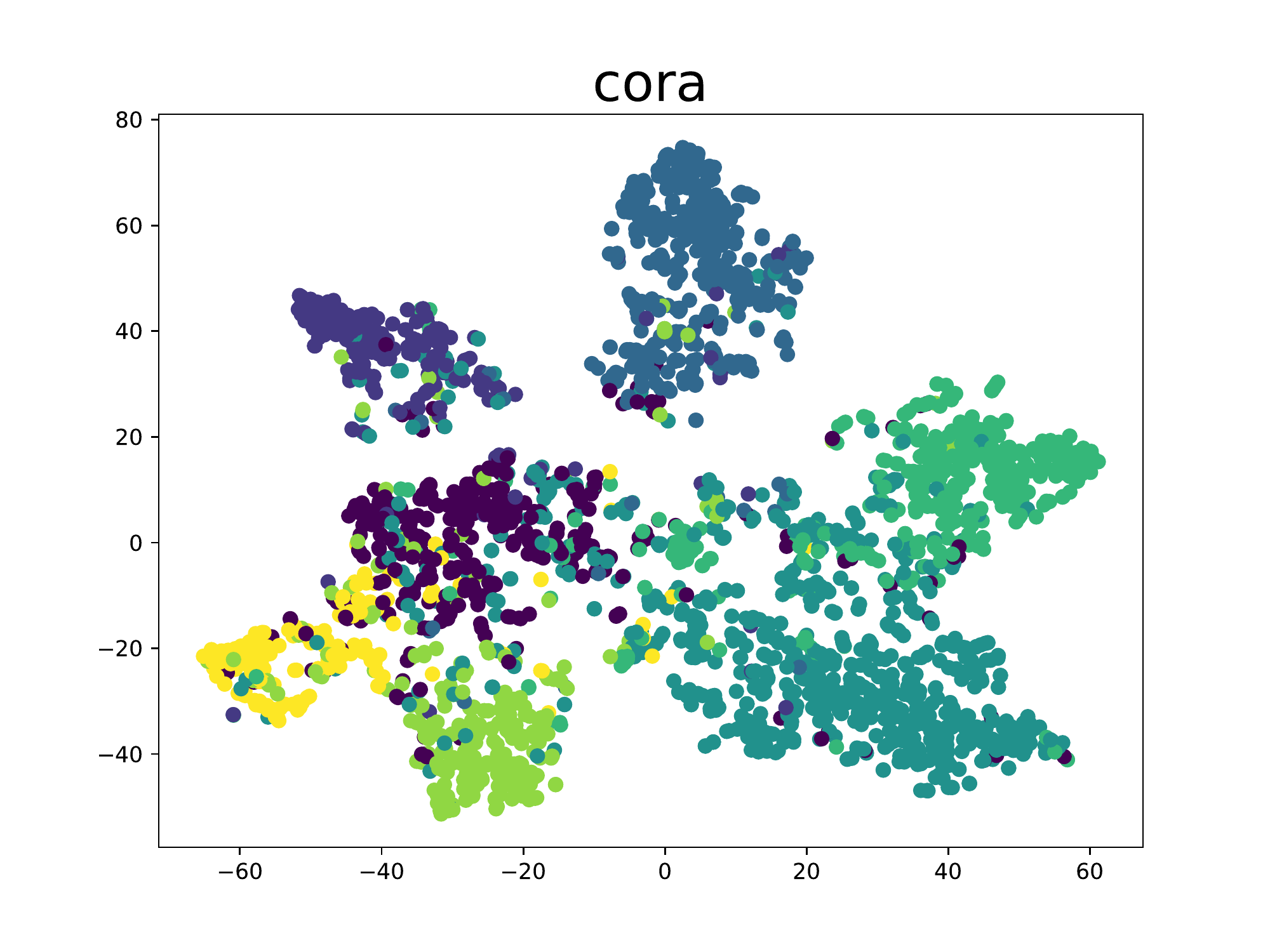}
        \caption{GCN.}
	\end{subfigure}
    \vspace{-5pt}
    \caption{Visualization of node embeddings for Cora dataset using t-SNE~\citep{JMLR:v9:vandermaaten08a}}
\end{figure}

\begin{figure}[htp]
    \centering
	\begin{subfigure}[b]{0.28\textwidth}
		\centering
		\includegraphics[width=\textwidth]{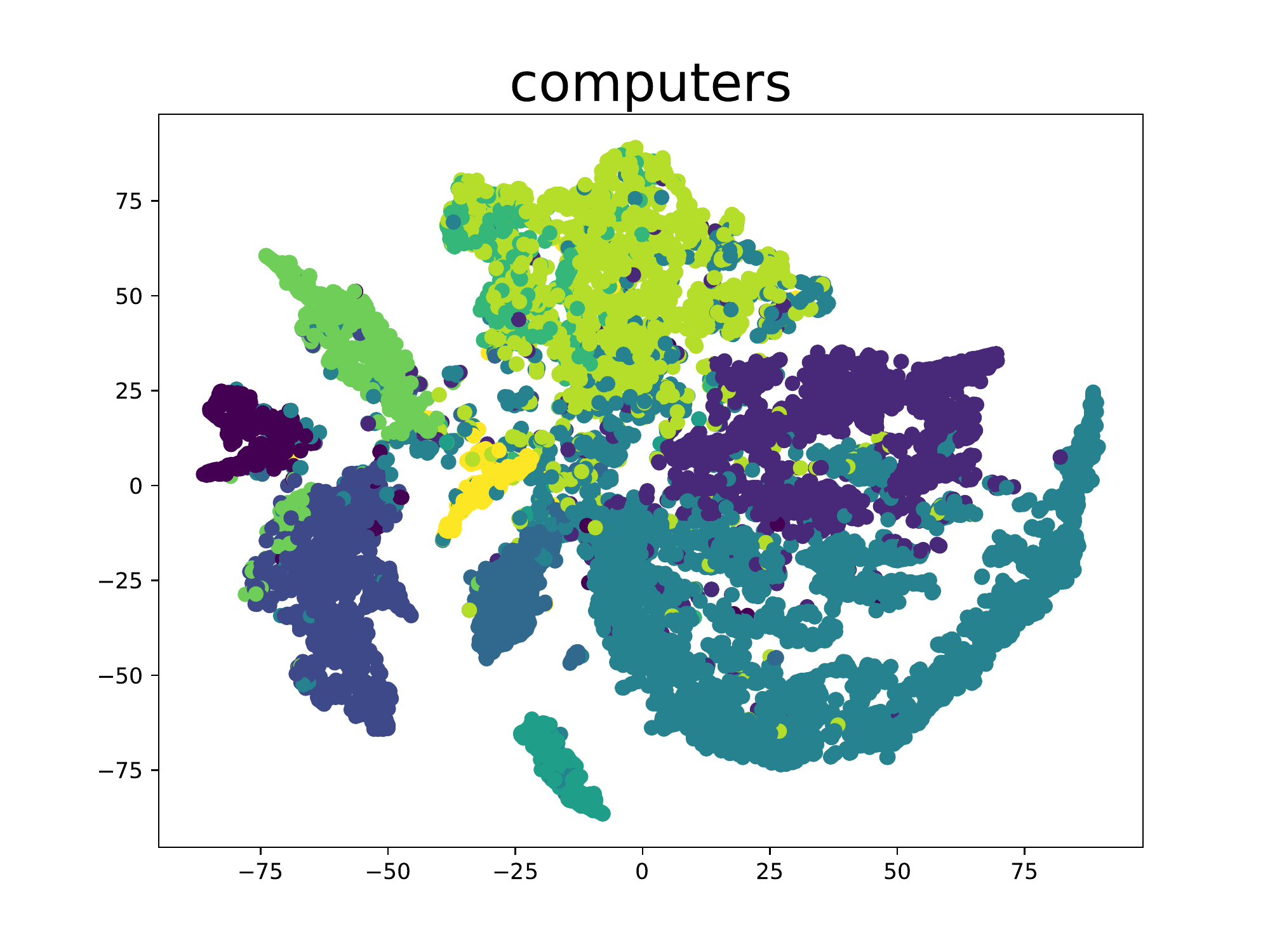}
        \caption{$^{1.0}$GNN.}
	\end{subfigure}
	\begin{subfigure}[b]{0.28\textwidth}
		\centering
		\includegraphics[width=\textwidth]{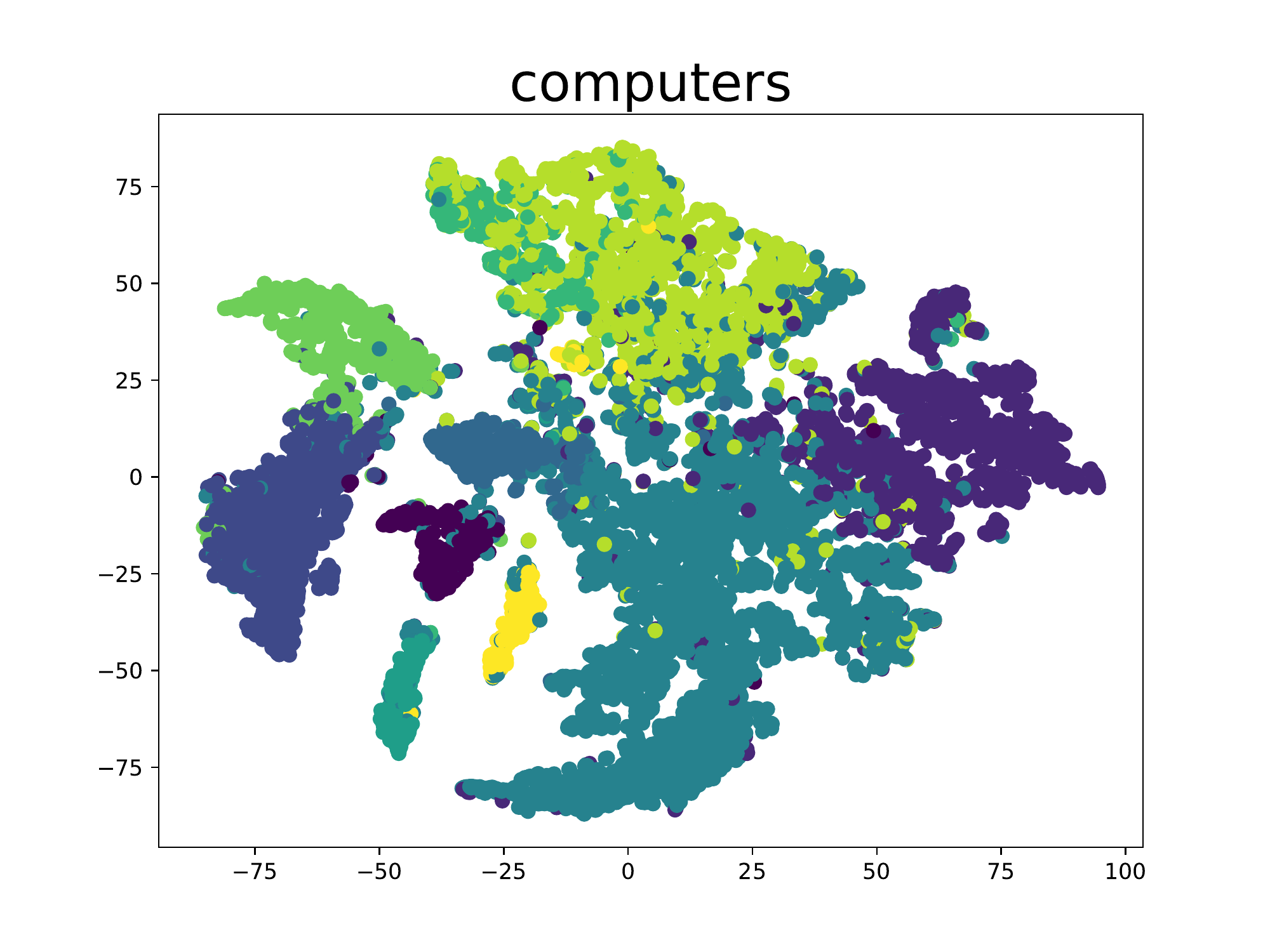}
        \caption{$^{1.5}$GNN.}
	\end{subfigure}
	\begin{subfigure}[b]{0.28\textwidth}
		\centering
		\includegraphics[width=\textwidth]{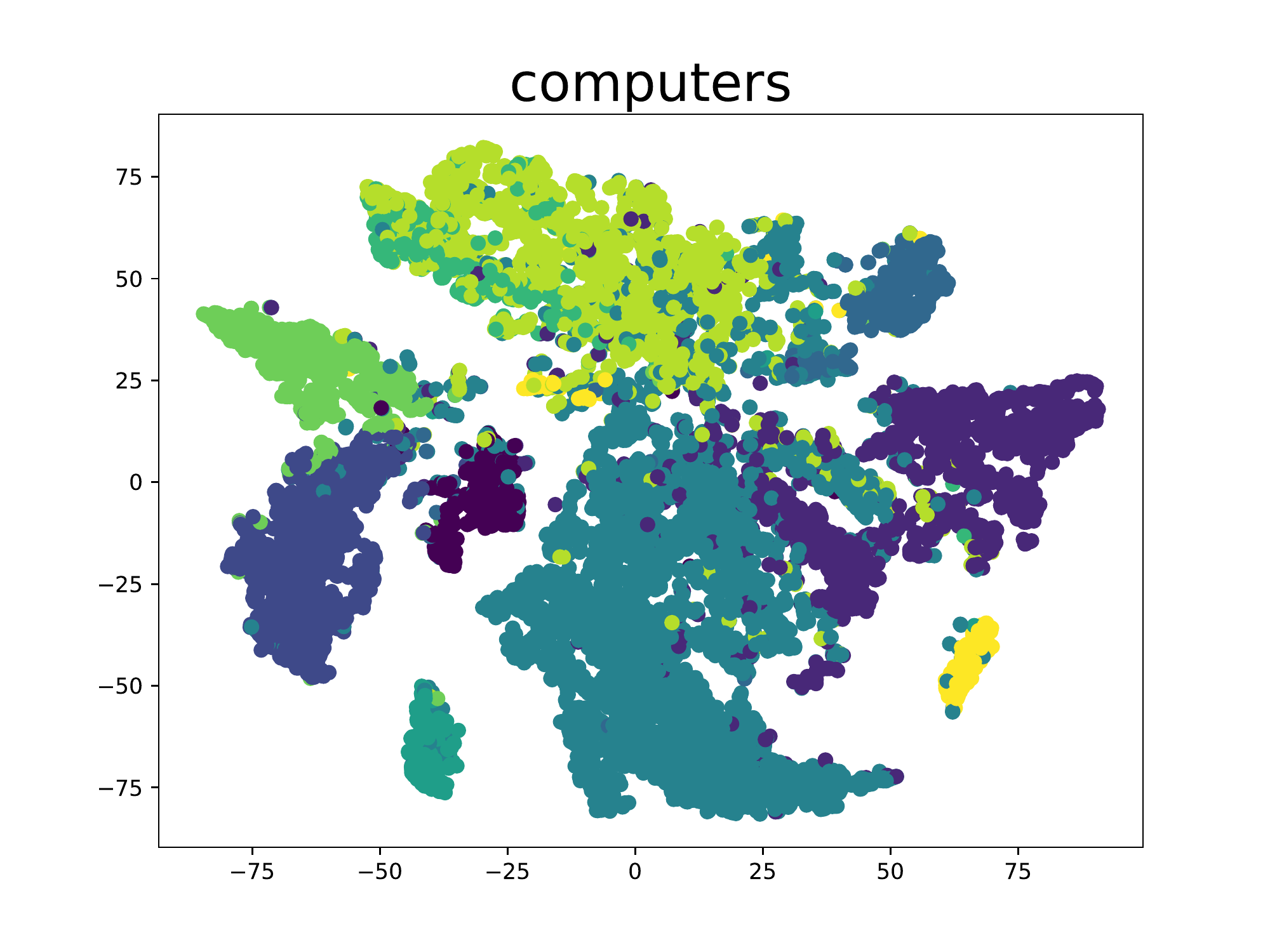}
        \caption{$^{2.0}$GNN.}
	\end{subfigure}
	\begin{subfigure}[b]{0.28\textwidth}
		\centering
		\includegraphics[width=\textwidth]{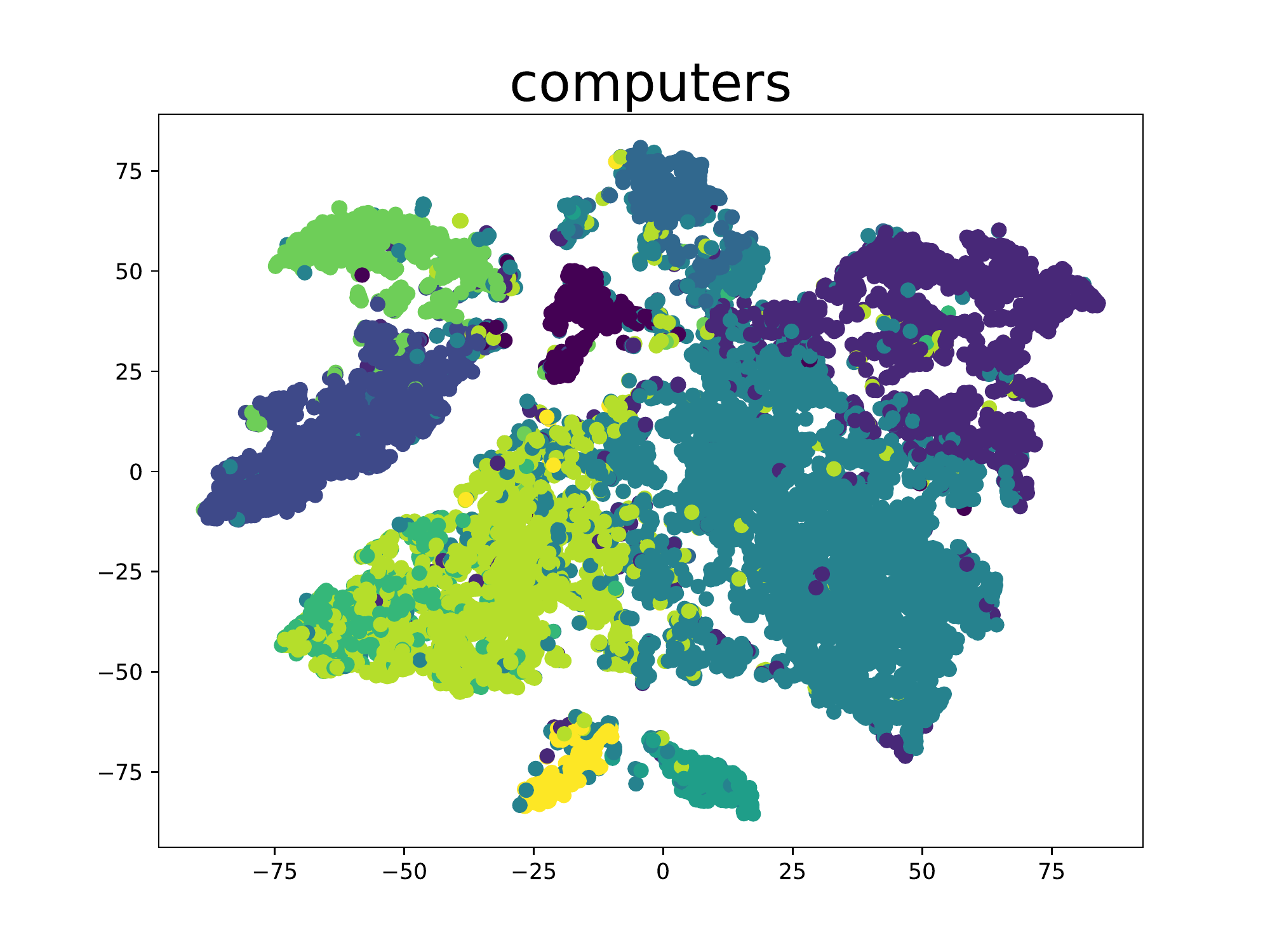}
        \caption{$^{2.5}$GNN.}
	\end{subfigure}
	\begin{subfigure}[b]{0.28\textwidth}
		\centering
		\includegraphics[width=\textwidth]{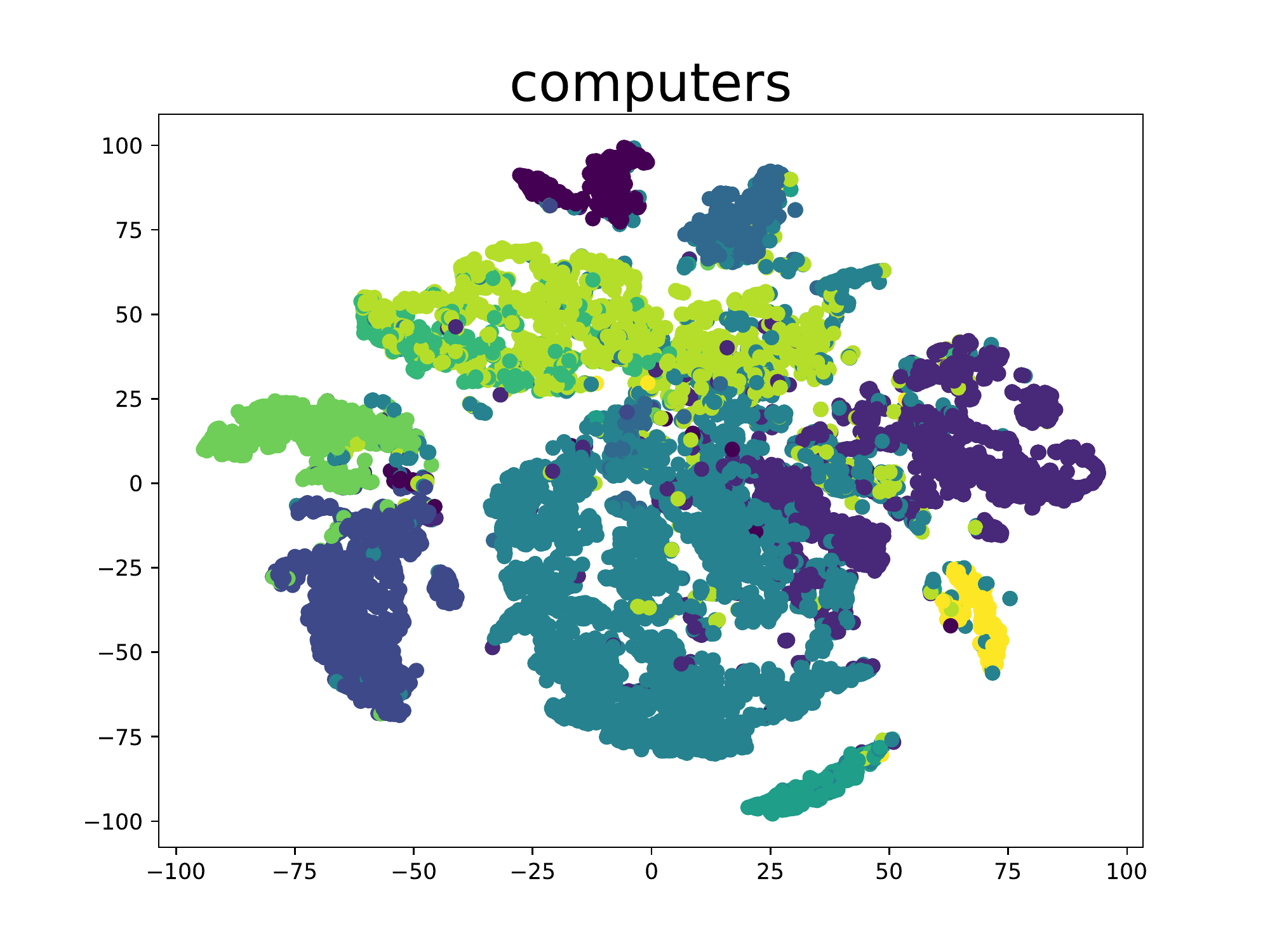}
        \caption{GCN.}
	\end{subfigure}
    \vspace{-5pt}
    \caption{Visualization of node embeddings for Computers dataset using t-SNE.}
    \vspace{-1.cm}
\end{figure}

\begin{figure}[htp]
    \centering
	\begin{subfigure}[b]{0.28\textwidth}
		\centering
		\includegraphics[width=\textwidth]{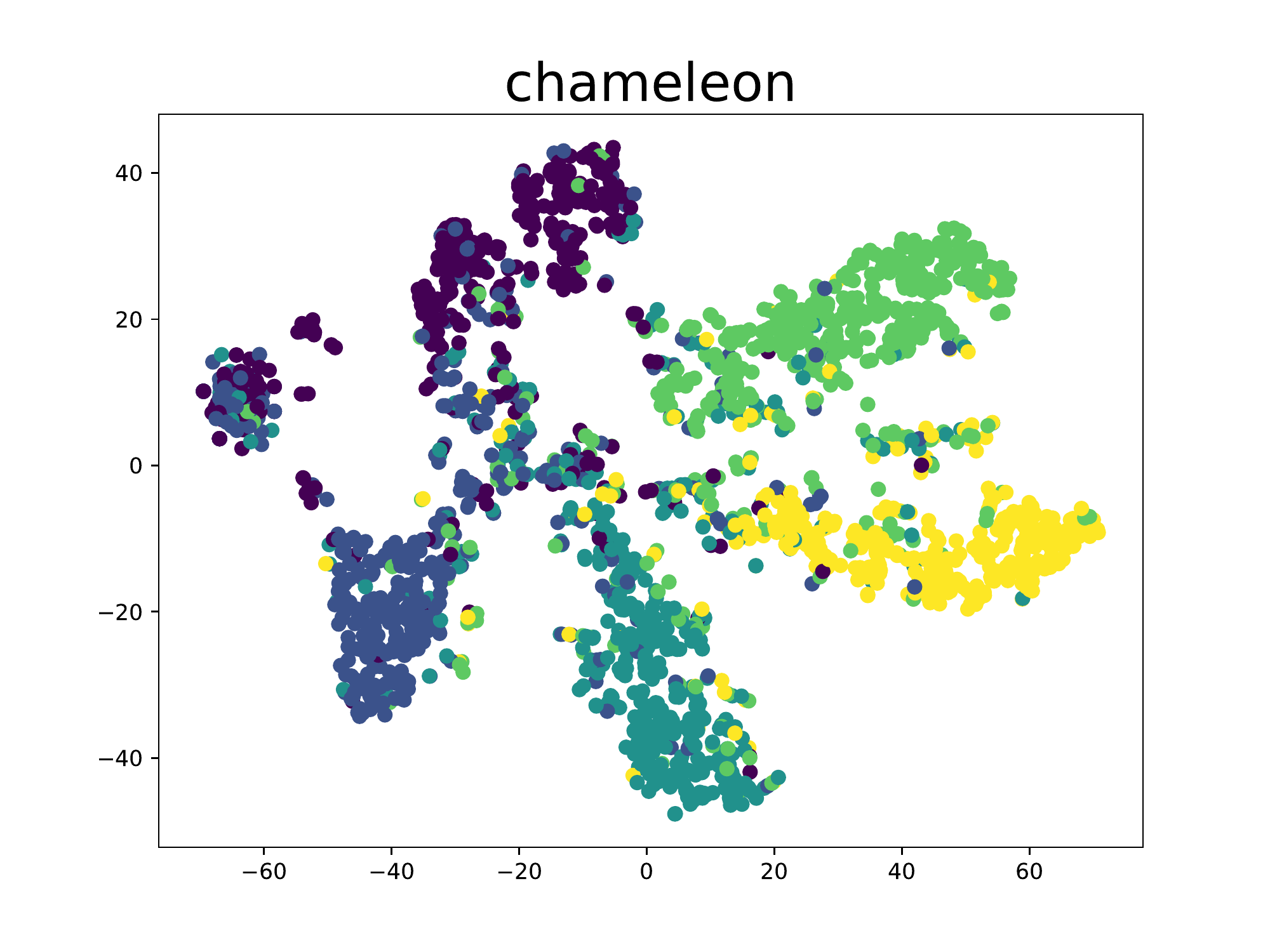}
        \caption{$^{1.0}$GNN.}
	\end{subfigure}
	\begin{subfigure}[b]{0.28\textwidth}
		\centering
		\includegraphics[width=\textwidth]{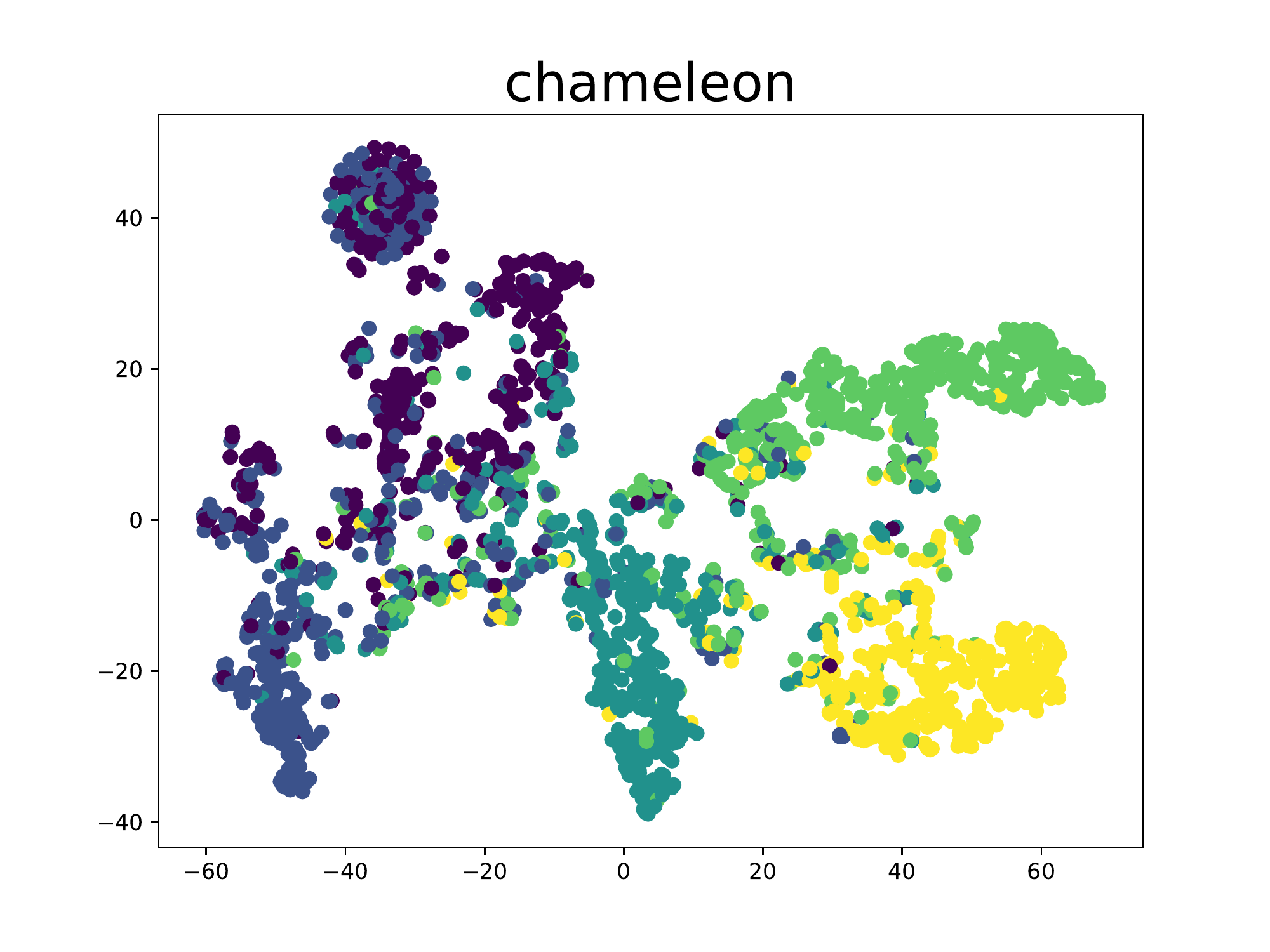}
        \caption{$^{1.5}$GNN.}
	\end{subfigure}
	\begin{subfigure}[b]{0.28\textwidth}
		\centering
		\includegraphics[width=\textwidth]{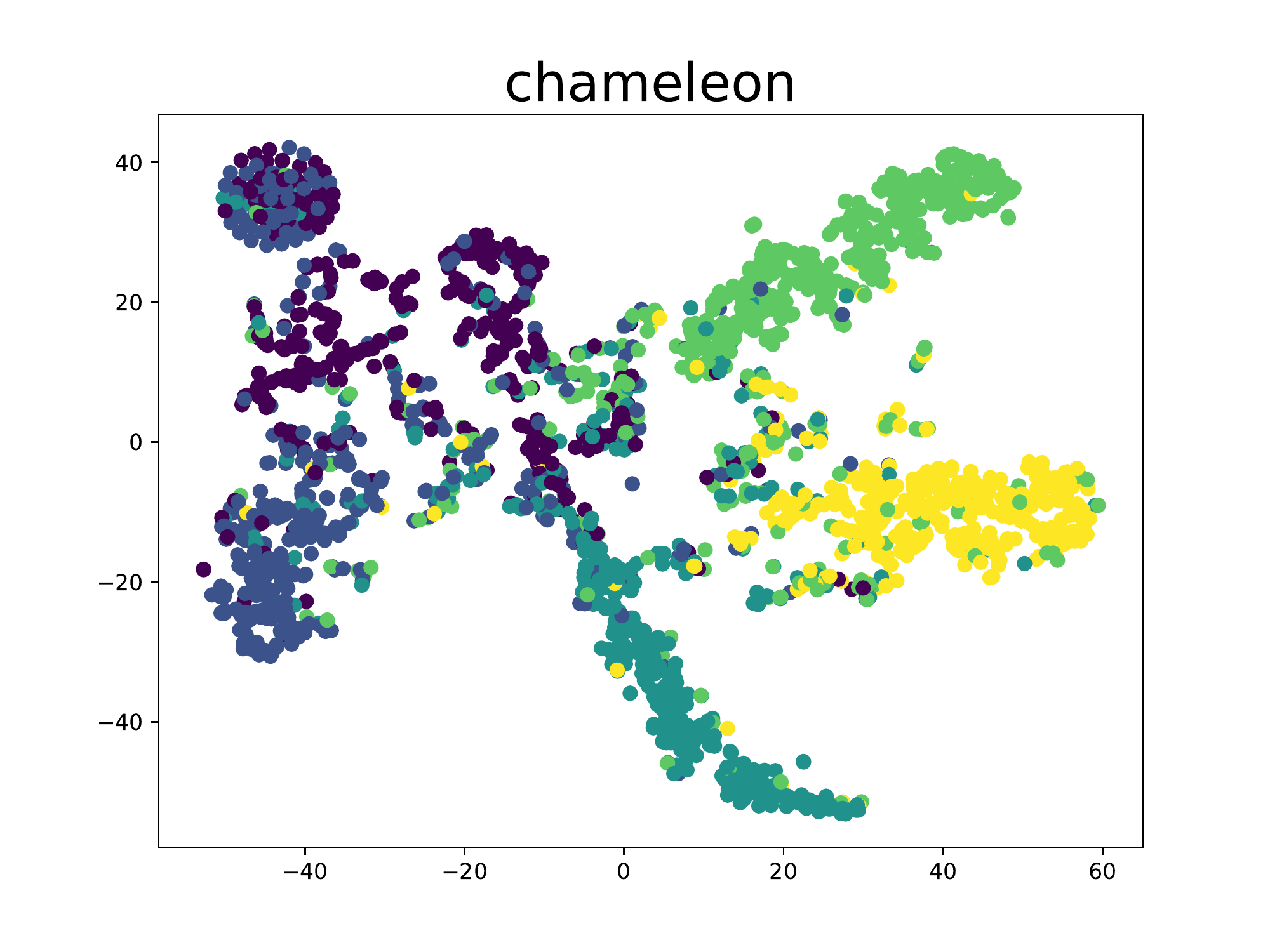}
        \caption{$^{2.0}$GNN.}
	\end{subfigure}
	\begin{subfigure}[b]{0.28\textwidth}
		\centering
		\includegraphics[width=\textwidth]{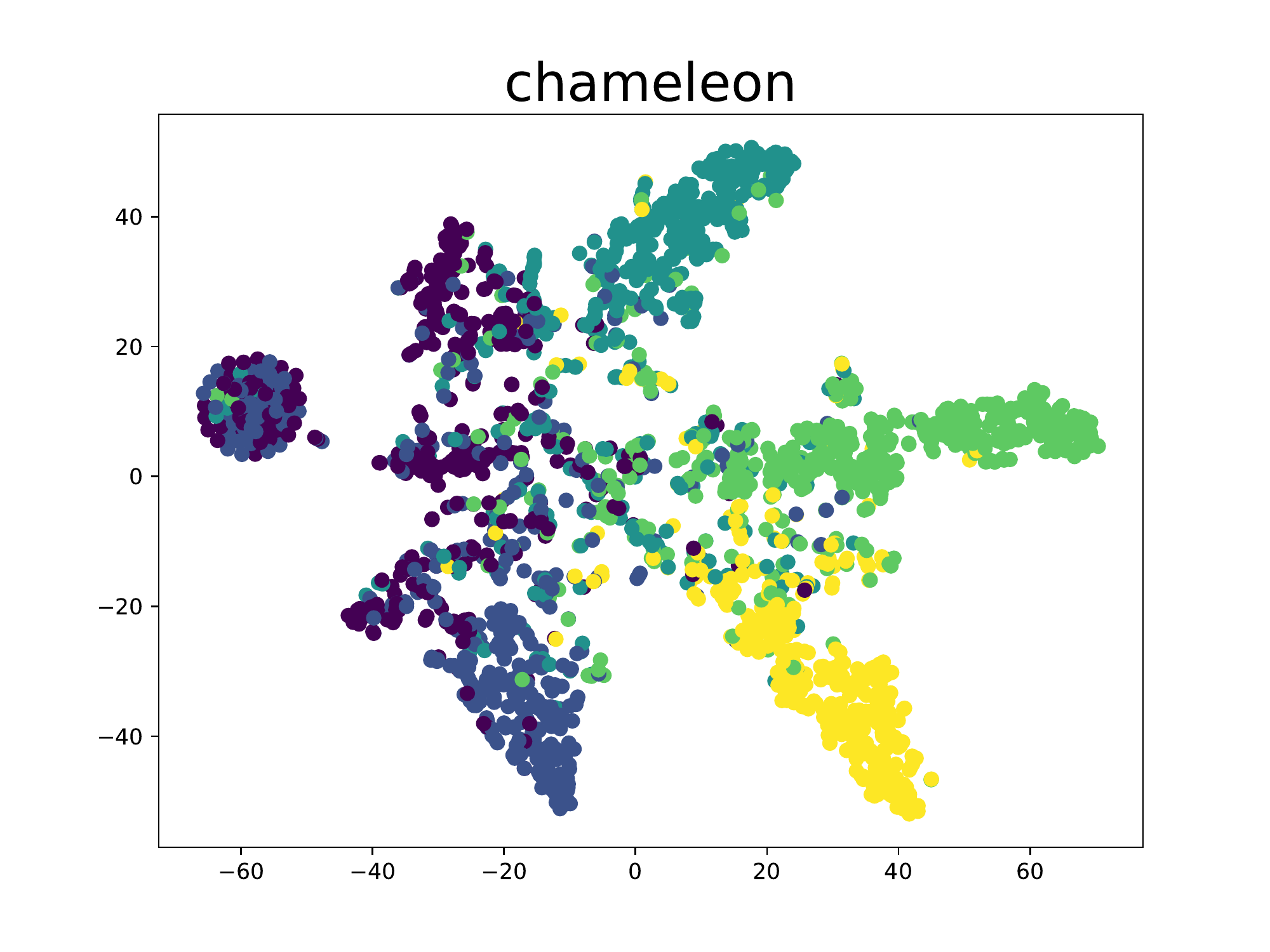}
        \caption{$^{2.5}$GNN.}
	\end{subfigure}
	\begin{subfigure}[b]{0.28\textwidth}
		\centering
		\includegraphics[width=\textwidth]{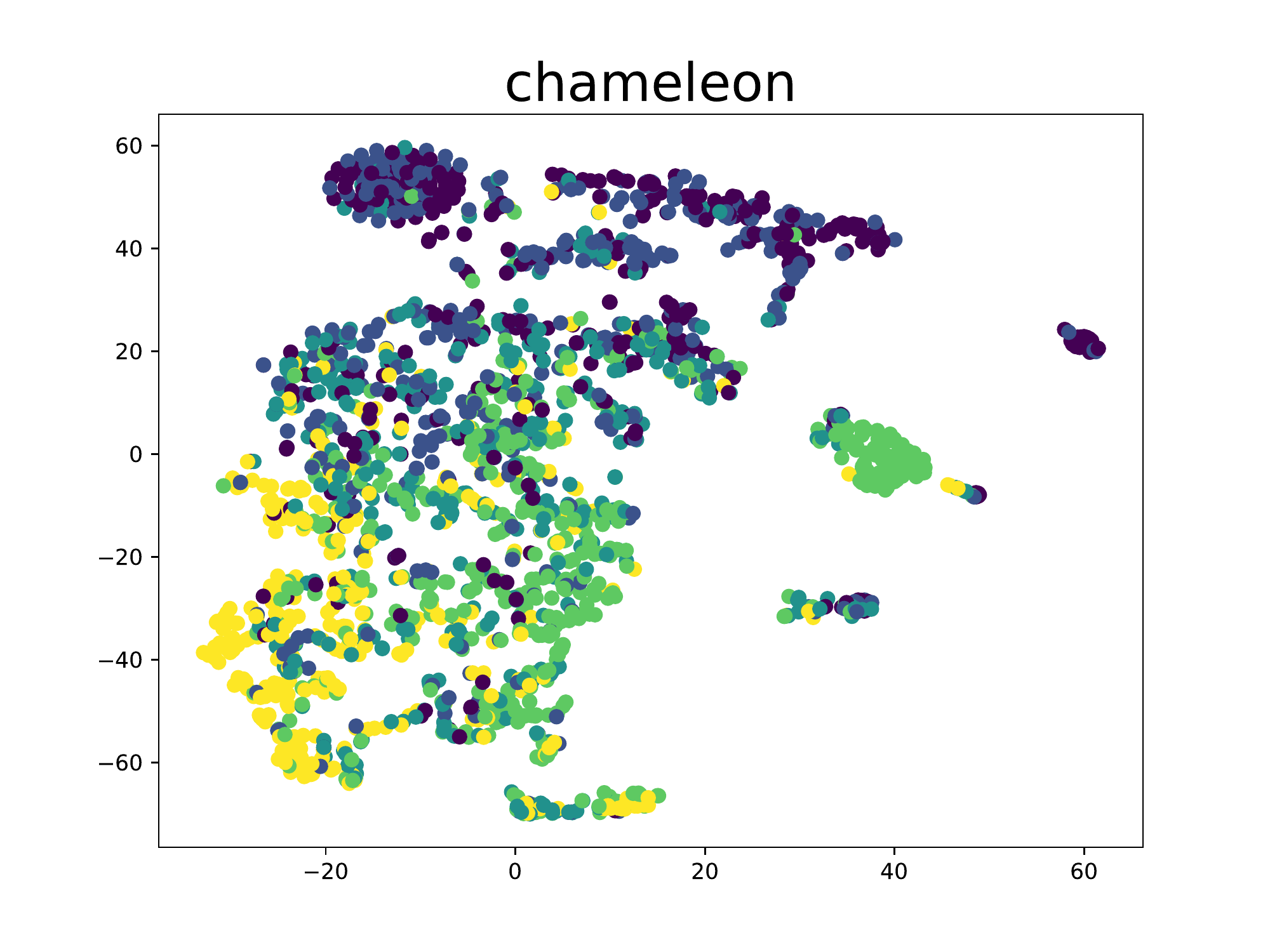}
        \caption{GCN.}
	\end{subfigure}
    \vspace{-5pt}
    \caption{Visualization of node embeddings for Chameleon dataset using t-SNE.}
\end{figure}

\begin{figure}[htp]
    \centering
	\begin{subfigure}[b]{0.28\textwidth}
		\centering
		\includegraphics[width=\textwidth]{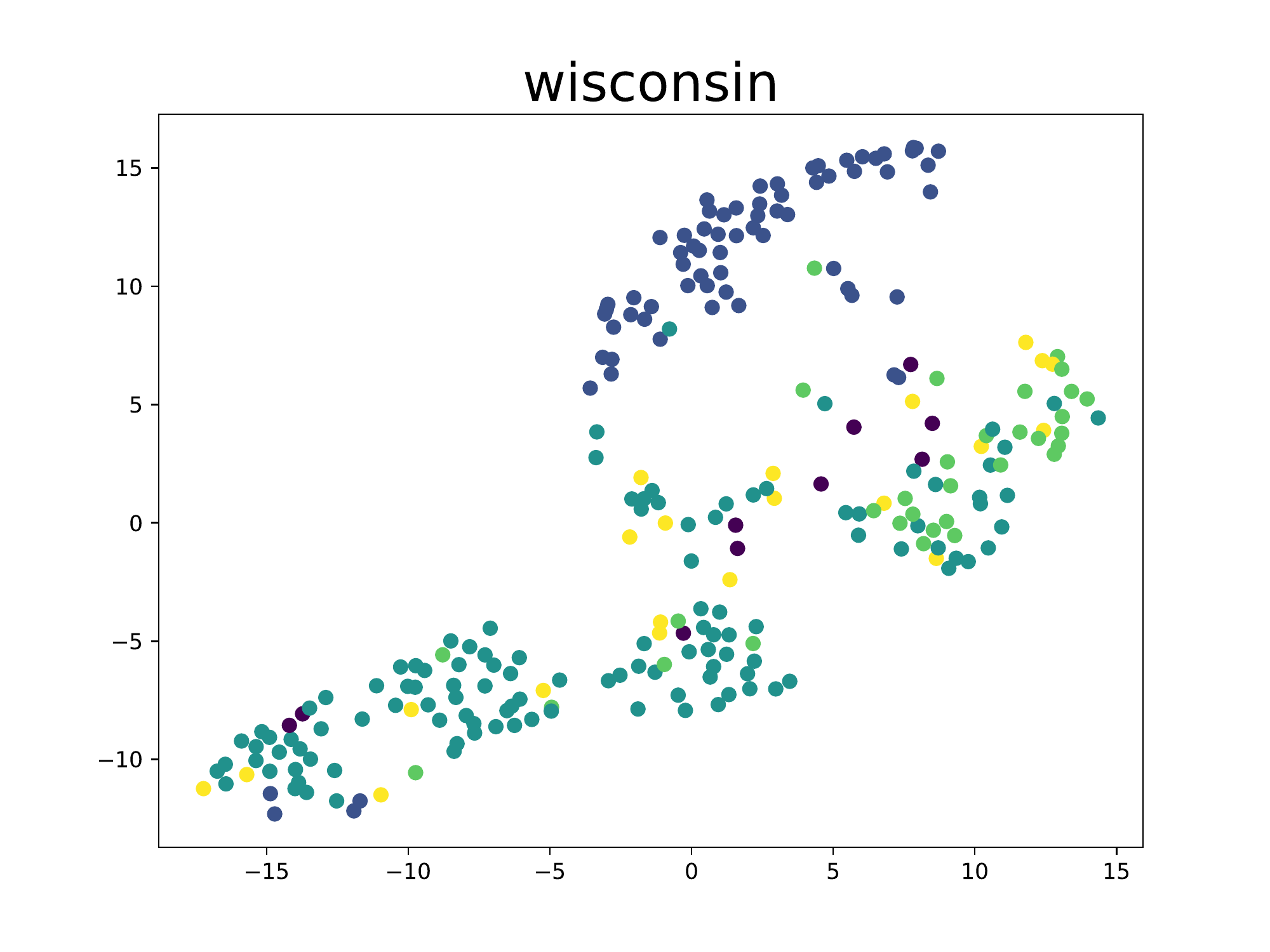}
        \caption{$^{1.0}$GNN.}
	\end{subfigure}
	\begin{subfigure}[b]{0.28\textwidth}
		\centering
		\includegraphics[width=\textwidth]{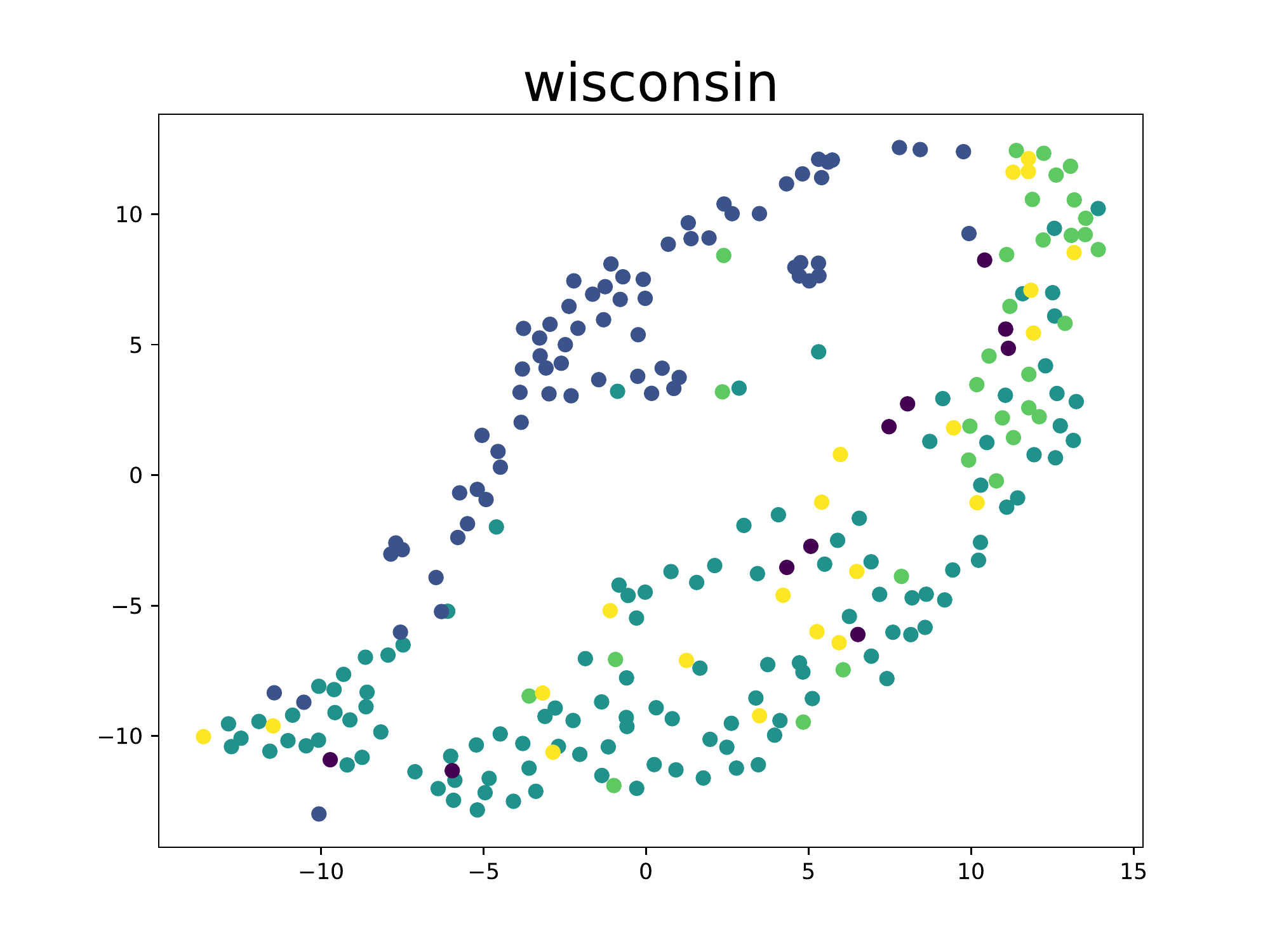}
        \caption{$^{1.5}$GNN.}
	\end{subfigure}
	\begin{subfigure}[b]{0.28\textwidth}
		\centering
		\includegraphics[width=\textwidth]{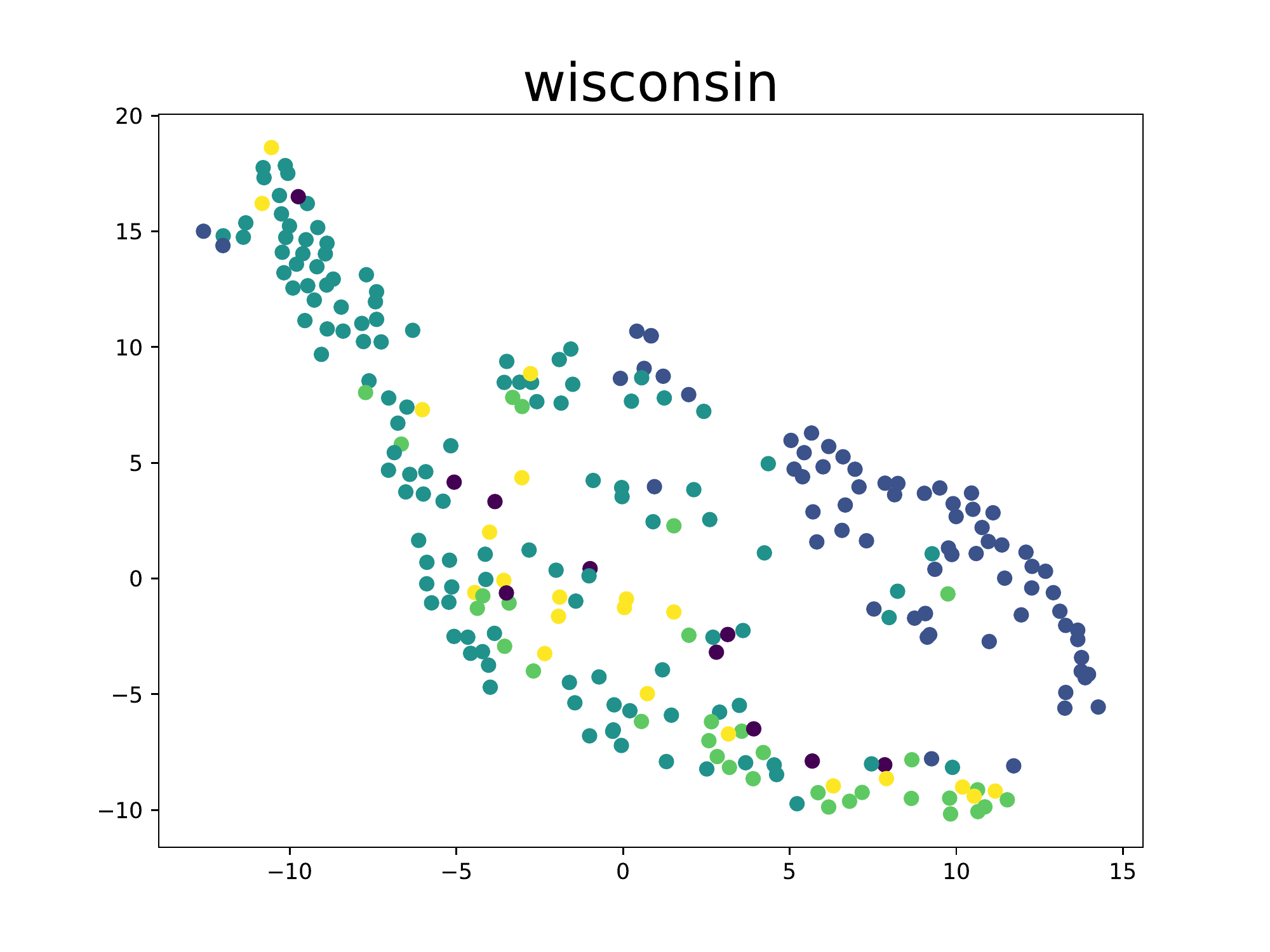}
        \caption{$^{2.0}$GNN.}
	\end{subfigure}
	\begin{subfigure}[b]{0.28\textwidth}
		\centering
		\includegraphics[width=\textwidth]{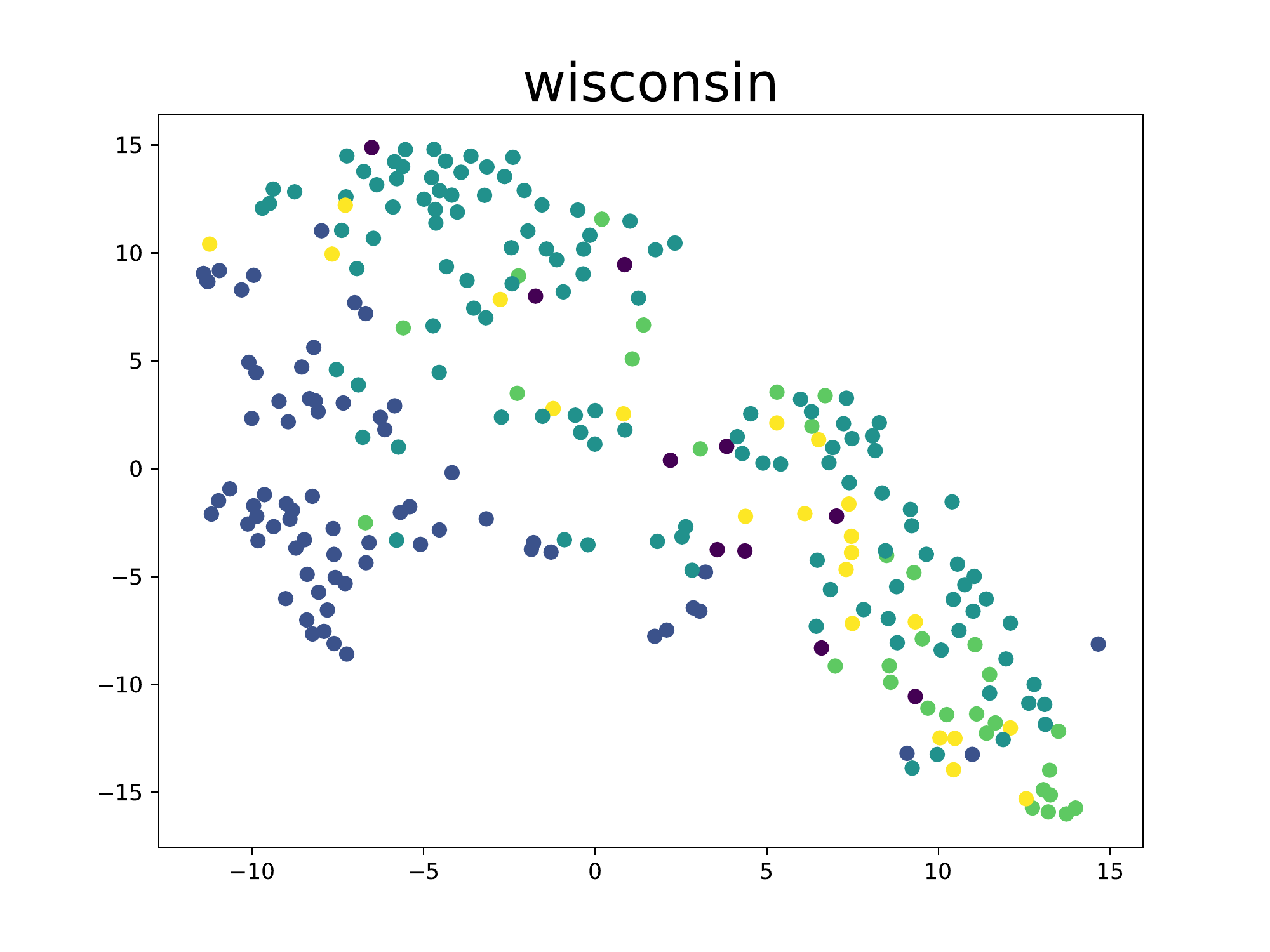}
        \caption{$^{2.5}$GNN.}
	\end{subfigure}
	\begin{subfigure}[b]{0.28\textwidth}
		\centering
		\includegraphics[width=\textwidth]{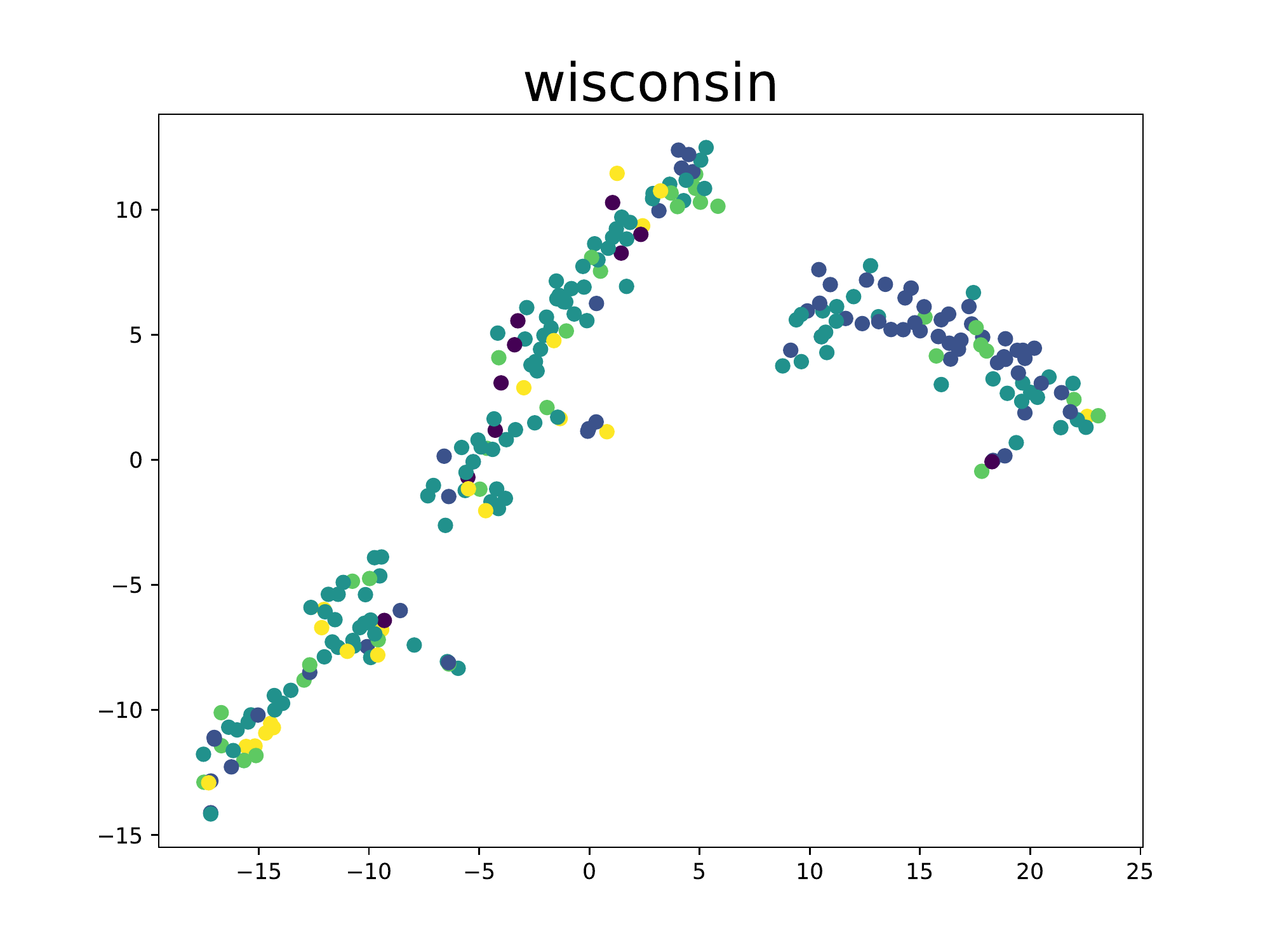}
        \caption{GCN.}
	\end{subfigure}
    \vspace{-5pt}
    \caption{Visualization of node embeddings for Wisconsin Dataset using t-SNE.}
\end{figure}

%% file: main.bbl
\begin{thebibliography}{75}
\providecommand{\natexlab}[1]{#1}
\providecommand{\url}[1]{\texttt{#1}}
\expandafter\ifx\csname urlstyle\endcsname\relax
  \providecommand{\doi}[1]{doi: #1}\else
  \providecommand{\doi}{doi: \begingroup \urlstyle{rm}\Url}\fi

\bibitem[Abu{-}El{-}Haija et~al.(2018)Abu{-}El{-}Haija, Perozzi, Al{-}Rfou, and
  Alemi]{DBLP:conf/nips/Abu-El-HaijaPAA18}
Abu{-}El{-}Haija, S., Perozzi, B., Al{-}Rfou, R., and Alemi, A.~A.
\newblock Watch your step: Learning node embeddings via graph attention.
\newblock In Bengio, S., Wallach, H.~M., Larochelle, H., Grauman, K.,
  Cesa{-}Bianchi, N., and Garnett, R. (eds.), \emph{Advances in Neural
  Information Processing Systems 31, {NeurIPS} 2018, December 3-8, 2018,
  Montr{\'{e}}al, Canada}, pp.\  9198--9208, 2018.

\bibitem[Abu{-}El{-}Haija et~al.(2019)Abu{-}El{-}Haija, Kapoor, Perozzi, and
  Lee]{DBLP:conf/uai/Abu-El-HaijaKPL19}
Abu{-}El{-}Haija, S., Kapoor, A., Perozzi, B., and Lee, J.
\newblock {N-GCN:} multi-scale graph convolution for semi-supervised node
  classification.
\newblock In Globerson, A. and Silva, R. (eds.), \emph{Proceedings of the
  Thirty-Fifth Conference on Uncertainty in Artificial Intelligence, {UAI}
  2019, Tel Aviv, Israel, July 22-25, 2019}, volume 115 of \emph{Proceedings of
  Machine Learning Research}, pp.\  841--851. {AUAI} Press, 2019.

\bibitem[Amghibech(2003)]{DBLP:journals/arscom/Amghibech03}
Amghibech, S.
\newblock Eigenvalues of the discrete p-laplacian for graphs.
\newblock \emph{Ars Combinatoria}, 67:\penalty0 283--302, 2003.

\bibitem[Atwood \& Towsley(2016)Atwood and Towsley]{DBLP:conf/nips/AtwoodT16}
Atwood, J. and Towsley, D.
\newblock Diffusion-convolutional neural networks.
\newblock In Lee, D.~D., Sugiyama, M., von Luxburg, U., Guyon, I., and Garnett,
  R. (eds.), \emph{Advances in Neural Information Processing Systems 29,
  {NeurIPS} 2016, December 5-10, 2016, Barcelona, Spain}, pp.\  1993--2001,
  2016.

\bibitem[Bach \& Jordan(2004)Bach and Jordan]{bach2004learning}
Bach, F. and Jordan, M.
\newblock Learning spectral clustering.
\newblock \emph{Advances in Neural Information Processing Systems, {NIPS}
  2004}, 16\penalty0 (2):\penalty0 305--312, 2004.

\bibitem[Battaglia et~al.(2018)Battaglia, Hamrick, Bapst, Sanchez{-}Gonzalez,
  Zambaldi, Malinowski, Tacchetti, Raposo, Santoro, Faulkner,
  G{\"{u}}l{\c{c}}ehre, Song, Ballard, Gilmer, Dahl, Vaswani, Allen, Nash,
  Langston, Dyer, Heess, Wierstra, Kohli, Botvinick, Vinyals, Li, and
  Pascanu]{DBLP:journals/corr/abs-1806-01261}
Battaglia, P.~W., Hamrick, J.~B., Bapst, V., Sanchez{-}Gonzalez, A., Zambaldi,
  V.~F., Malinowski, M., Tacchetti, A., Raposo, D., Santoro, A., Faulkner, R.,
  G{\"{u}}l{\c{c}}ehre, {\c{C}}., Song, H.~F., Ballard, A.~J., Gilmer, J.,
  Dahl, G.~E., Vaswani, A., Allen, K.~R., Nash, C., Langston, V., Dyer, C.,
  Heess, N., Wierstra, D., Kohli, P., Botvinick, M., Vinyals, O., Li, Y., and
  Pascanu, R.
\newblock Relational inductive biases, deep learning, and graph networks.
\newblock \emph{CoRR}, abs/1806.01261, 2018.

\bibitem[Belkin \& Niyogi(2008)Belkin and Niyogi]{DBLP:journals/jcss/BelkinN08}
Belkin, M. and Niyogi, P.
\newblock Towards a theoretical foundation for laplacian-based manifold
  methods.
\newblock \emph{Journal of Computer and System Sciences}, 74\penalty0
  (8):\penalty0 1289--1308, 2008.

\bibitem[Belkin et~al.(2004)Belkin, Matveeva, and
  Niyogi]{DBLP:conf/colt/BelkinMN04}
Belkin, M., Matveeva, I., and Niyogi, P.
\newblock Regularization and semi-supervised learning on large graphs.
\newblock In \emph{The 17th Annual Conference on Learning Theory, {COLT} 2004,
  Banff, Canada, July 1-4, 2004}, volume 3120, pp.\  624--638, 2004.

\bibitem[Belkin et~al.(2005)Belkin, Niyogi, and
  Sindhwani]{DBLP:conf/aistats/BelkinNS05}
Belkin, M., Niyogi, P., and Sindhwani, V.
\newblock On manifold regularization.
\newblock In Cowell, R.~G. and Ghahramani, Z. (eds.), \emph{Proceedings of the
  Tenth International Workshop on Artificial Intelligence and Statistics,
  {AISTATS} 2005, Bridgetown, Barbados, January 6-8, 2005}. Society for
  Artificial Intelligence and Statistics, 2005.

\bibitem[Bruna et~al.(2014)Bruna, Zaremba, Szlam, and
  LeCun]{DBLP:journals/iclr/BrunaZSL14}
Bruna, J., Zaremba, W., Szlam, A., and LeCun, Y.
\newblock Spectral networks and locally connected networks on graphs.
\newblock In \emph{The 2nd International Conference on Learning
  Representations, {ICLR} 2014, Banff, AB, Canada, April 14-16, 2014}, 2014.

\bibitem[B{\"{u}}hler \& Hein(2009)B{\"{u}}hler and
  Hein]{DBLP:conf/icml/BuhlerH09}
B{\"{u}}hler, T. and Hein, M.
\newblock Spectral clustering based on the graph \emph{p}-laplacian.
\newblock In Danyluk, A.~P., Bottou, L., and Littman, M.~L. (eds.),
  \emph{Proceedings of the 26th Annual International Conference on Machine
  Learning, {ICML} 2009, Montreal, Quebec, Canada, June 14-18, 2009}, volume
  382 of \emph{{ACM} International Conference Proceeding Series}, pp.\  81--88.
  {ACM}, 2009.

\bibitem[Chen et~al.(2018)Chen, Ma, and Xiao]{DBLP:conf/iclr/ChenMX18}
Chen, J., Ma, T., and Xiao, C.
\newblock Fastgcn: Fast learning with graph convolutional networks via
  importance sampling.
\newblock In \emph{6th International Conference on Learning Representations,
  {ICLR} 2018, Vancouver, BC, Canada, April 30 - May 3, 2018, Conference Track
  Proceedings}, 2018.

\bibitem[Chien et~al.(2021)Chien, Peng, Li, and
  Milenkovic]{DBLP:conf/iclr/ChienP0M21}
Chien, E., Peng, J., Li, P., and Milenkovic, O.
\newblock Adaptive universal generalized pagerank graph neural network.
\newblock In \emph{9th International Conference on Learning Representations,
  {ICLR} 2021, Virtual Event, Austria, May 3-7, 2021}. OpenReview.net, 2021.

\bibitem[Defferrard et~al.(2016)Defferrard, Bresson, and
  Vandergheynst]{DBLP:conf/nips/DefferrardBV16}
Defferrard, M., Bresson, X., and Vandergheynst, P.
\newblock Convolutional neural networks on graphs with fast localized spectral
  filtering.
\newblock In \emph{Advances in Neural Information Processing Systems 29,
  {NeurIPS} 2016, December 5-10, 2016, Barcelona, Spain}, pp.\  3837--3845,
  2016.

\bibitem[Deshpande et~al.(2018)Deshpande, Sen, Montanari, and
  Mossel]{DBLP:conf/nips/DeshpandeSMM18}
Deshpande, Y., Sen, S., Montanari, A., and Mossel, E.
\newblock Contextual stochastic block models.
\newblock In Bengio, S., Wallach, H.~M., Larochelle, H., Grauman, K.,
  Cesa{-}Bianchi, N., and Garnett, R. (eds.), \emph{Advances in Neural
  Information Processing Systems 31, {NeurIPS 2018}, December 3-8, 2018,
  Montr{\'{e}}al, Canada}, pp.\  8590--8602, 2018.

\bibitem[Duvenaud et~al.(2015)Duvenaud, Maclaurin, Aguilera{-}Iparraguirre,
  G{\'{o}}mez{-}Bombarelli, Hirzel, Aspuru{-}Guzik, and
  Adams]{DBLP:conf/nips/DuvenaudMABHAA15}
Duvenaud, D., Maclaurin, D., Aguilera{-}Iparraguirre, J.,
  G{\'{o}}mez{-}Bombarelli, R., Hirzel, T., Aspuru{-}Guzik, A., and Adams,
  R.~P.
\newblock Convolutional networks on graphs for learning molecular fingerprints.
\newblock In Cortes, C., Lawrence, N.~D., Lee, D.~D., Sugiyama, M., and
  Garnett, R. (eds.), \emph{Advances in Neural Information Processing Systems
  28, {NeurIPS 2015} December 7-12, 2015, Montreal, Quebec, Canada}, pp.\
  2224--2232, 2015.

\bibitem[Fey \& Lenssen(2019)Fey and
  Lenssen]{DBLP:journals/corr/abs-1903-02428}
Fey, M. and Lenssen, J.~E.
\newblock Fast graph representation learning with pytorch geometric.
\newblock \emph{CoRR}, abs/1903.02428, 2019.

\bibitem[Fu et~al.(2020)Fu, Hou, Zhang, Ma, Kamhoua, and
  Cheng]{DBLP:journals/corr/abs-2006-04386}
Fu, G., Hou, Y., Zhang, J., Ma, K., Kamhoua, B.~F., and Cheng, J.
\newblock Understanding graph neural networks from graph signal denoising
  perspectives.
\newblock \emph{CoRR}, abs/2006.04386, 2020.

\bibitem[Gama et~al.(2020)Gama, Bruna, and Ribeiro]{DBLP:journals/tsp/GamaBR20}
Gama, F., Bruna, J., and Ribeiro, A.
\newblock Stability properties of graph neural networks.
\newblock \emph{{IEEE} Transactions on Signal Processing}, 68:\penalty0
  5680--5695, 2020.

\bibitem[Garg et~al.(2020)Garg, Jegelka, and Jaakkola]{DBLP:conf/icml/GargJJ20}
Garg, V.~K., Jegelka, S., and Jaakkola, T.~S.
\newblock Generalization and representational limits of graph neural networks.
\newblock In \emph{Proceedings of the 37th International Conference on Machine
  Learning, {ICML} 2020, 13-18 July 2020, Virtual Event}, volume 119 of
  \emph{Proceedings of Machine Learning Research}, pp.\  3419--3430. {PMLR},
  2020.

\bibitem[Grandvalet \& Bengio(2004)Grandvalet and
  Bengio]{DBLP:conf/nips/GrandvaletB04}
Grandvalet, Y. and Bengio, Y.
\newblock Semi-supervised learning by entropy minimization.
\newblock In \emph{Advances in Neural Information Processing Systems 17, {NIPS}
  2004, December 13-18, 2004, Vancouver, British Columbia, Canada]}, pp.\
  529--536, 2004.

\bibitem[Hamilton et~al.(2017)Hamilton, Ying, and
  Leskovec]{DBLP:conf/nips/HamiltonYL17}
Hamilton, W.~L., Ying, Z., and Leskovec, J.
\newblock Inductive representation learning on large graphs.
\newblock In \emph{Advances in Neural Information Processing Systems 30,
  {NeurIPS} 2017, 4-9 December 2017, Long Beach, CA, {USA}}, pp.\  1024--1034,
  2017.

\bibitem[Hein(2006)]{DBLP:conf/colt/Hein06}
Hein, M.
\newblock Uniform convergence of adaptive graph-based regularization.
\newblock In \emph{The 19th Annual Conference on Learning Theory, {COLT} 2006,
  Pittsburgh, PA, USA, June 22-25, 2006}, volume 4005, pp.\  50--64, 2006.

\bibitem[Henaff et~al.(2015)Henaff, Bruna, and
  LeCun]{DBLP:journals/corr/HenaffBL15}
Henaff, M., Bruna, J., and LeCun, Y.
\newblock Deep convolutional networks on graph-structured data.
\newblock \emph{CoRR}, abs/1506.05163, 2015.

\bibitem[Hu et~al.(2020)Hu, Fey, Zitnik, Dong, Ren, Liu, Catasta, and
  Leskovec]{DBLP:conf/nips/HuFZDRLCL20}
Hu, W., Fey, M., Zitnik, M., Dong, Y., Ren, H., Liu, B., Catasta, M., and
  Leskovec, J.
\newblock Open graph benchmark: Datasets for machine learning on graphs.
\newblock In Larochelle, H., Ranzato, M., Hadsell, R., Balcan, M., and Lin, H.
  (eds.), \emph{Advances in Neural Information Processing Systems 33, {NeurIPS}
  2020, December 6-12, 2020, virtual}, 2020.

\bibitem[Huang et~al.(2021)Huang, He, Singh, Lim, and
  Benson]{DBLP:conf/iclr/HuangHSLB21}
Huang, Q., He, H., Singh, A., Lim, S., and Benson, A.~R.
\newblock Combining label propagation and simple models out-performs graph
  neural networks.
\newblock In \emph{9th International Conference on Learning Representations,
  {ICLR} 2021, Virtual Event, Austria, May 3-7, 2021}. OpenReview.net, 2021.

\bibitem[Kipf \& Welling(2017)Kipf and Welling]{DBLP:conf/iclr/KipfW17}
Kipf, T.~N. and Welling, M.
\newblock Semi-supervised classification with graph convolutional networks.
\newblock In \emph{5th International Conference on Learning Representations,
  {ICLR} 2017, Toulon, France, April 24-26, 2017, Conference Track
  Proceedings}, 2017.

\bibitem[Klicpera et~al.(2019)Klicpera, Bojchevski, and
  G{\"{u}}nnemann]{DBLP:conf/iclr/KlicperaBG19}
Klicpera, J., Bojchevski, A., and G{\"{u}}nnemann, S.
\newblock Predict then propagate: Graph neural networks meet personalized
  pagerank.
\newblock In \emph{7th International Conference on Learning Representations,
  {ICLR} 2019, New Orleans, LA, USA, May 6-9, 2019}. OpenReview.net, 2019.

\bibitem[Li et~al.(2019)Li, M{\"{u}}ller, Thabet, and
  Ghanem]{DBLP:conf/iccv/Li0TG19}
Li, G., M{\"{u}}ller, M., Thabet, A.~K., and Ghanem, B.
\newblock Deepgcns: Can gcns go as deep as cnns?
\newblock In \emph{2019 {IEEE/CVF} International Conference on Computer Vision,
  {ICCV} 2019, Seoul, Korea (South), October 27 - November 2, 2019}, pp.\
  9266--9275. {IEEE}, 2019.

\bibitem[Li et~al.(2018)Li, Han, and Wu]{DBLP:conf/aaai/LiHW18}
Li, Q., Han, Z., and Wu, X.
\newblock Deeper insights into graph convolutional networks for semi-supervised
  learning.
\newblock In \emph{Thirty-Second {AAAI} Conference on Artificial Intelligence,
  {AAAI} 2018, New Orleans, Louisiana, USA, February 2-7, 2018}, pp.\
  3538--3545, 2018.

\bibitem[Liao et~al.(2019)Liao, Zhao, Urtasun, and
  Zemel]{DBLP:conf/iclr/LiaoZUZ19}
Liao, R., Zhao, Z., Urtasun, R., and Zemel, R.~S.
\newblock Lanczosnet: Multi-scale deep graph convolutional networks.
\newblock In \emph{7th International Conference on Learning Representations,
  {ICLR} 2019, New Orleans, LA, USA, May 6-9, 2019}. OpenReview.net, 2019.

\bibitem[Liu et~al.(2014)Liu, Chakraborty, Li, Liu, and
  Lozano]{liu2014bayesian}
Liu, F., Chakraborty, S., Li, F., Liu, Y., and Lozano, A.~C.
\newblock Bayesian regularization via graph laplacian.
\newblock \emph{Bayesian Analysis}, 9\penalty0 (2):\penalty0 449--474, 2014.

\bibitem[Liu \& Han(2013)Liu and Han]{DBLP:books/crc/aggarwal13/LiuH13}
Liu, J. and Han, J.
\newblock Spectral clustering.
\newblock In Aggarwal, C.~C. and Reddy, C.~K. (eds.), \emph{Data Clustering:
  Algorithms and Applications}, pp.\  177--200. {CRC} Press, 2013.

\bibitem[Liu et~al.(2021)Liu, Jin, Ma, Li, Liu, Wang, Yan, and
  Tang]{DBLP:conf/icml/LiuJ0LLW0T21}
Liu, X., Jin, W., Ma, Y., Li, Y., Liu, H., Wang, Y., Yan, M., and Tang, J.
\newblock Elastic graph neural networks.
\newblock In Meila, M. and Zhang, T. (eds.), \emph{Proceedings of the 38th
  International Conference on Machine Learning, {ICML} 2021, 18-24 July 2021,
  Virtual Event}, volume 139 of \emph{Proceedings of Machine Learning
  Research}, pp.\  6837--6849. {PMLR}, 2021.

\bibitem[Loukas(2020)]{DBLP:conf/iclr/Loukas20}
Loukas, A.
\newblock What graph neural networks cannot learn: depth vs width.
\newblock In \emph{8th International Conference on Learning Representations,
  {ICLR} 2020, Addis Ababa, Ethiopia, April 26-30, 2020}. OpenReview.net, 2020.

\bibitem[Luo et~al.(2010)Luo, Huang, Ding, and Nie]{DBLP:journals/ml/LuoHDN10}
Luo, D., Huang, H., Ding, C. H.~Q., and Nie, F.
\newblock On the eigenvectors of \emph{p}-laplacian.
\newblock \emph{Machine Learning}, 81\penalty0 (1):\penalty0 37--51, 2010.

\bibitem[Marcheggiani \& Titov(2017)Marcheggiani and
  Titov]{DBLP:conf/emnlp/MarcheggianiT17}
Marcheggiani, D. and Titov, I.
\newblock Encoding sentences with graph convolutional networks for semantic
  role labeling.
\newblock In Palmer, M., Hwa, R., and Riedel, S. (eds.), \emph{Proceedings of
  the 2017 Conference on Empirical Methods in Natural Language Processing,
  {EMNLP} 2017, Copenhagen, Denmark, September 9-11, 2017}, pp.\  1506--1515.
  Association for Computational Linguistics, 2017.

\bibitem[Nadler et~al.(2009)Nadler, Srebro, and Zhou]{nadler2009semi}
Nadler, B., Srebro, N., and Zhou, X.
\newblock Semi-supervised learning with the graph laplacian: The limit of
  infinite unlabelled data.
\newblock \emph{Advances in Neural Information Processing Systems, {NIPS}
  2009}, 22:\penalty0 1330--1338, 2009.

\bibitem[Niepert et~al.(2016)Niepert, Ahmed, and
  Kutzkov]{DBLP:conf/icml/NiepertAK16}
Niepert, M., Ahmed, M., and Kutzkov, K.
\newblock Learning convolutional neural networks for graphs.
\newblock In Balcan, M. and Weinberger, K.~Q. (eds.), \emph{Proceedings of the
  33nd International Conference on Machine Learning, {ICML} 2016, New York
  City, NY, USA, June 19-24, 2016}, volume~48 of \emph{{JMLR} Workshop and
  Conference Proceedings}, pp.\  2014--2023. JMLR.org, 2016.

\bibitem[Niyogi(2013)]{DBLP:journals/jmlr/Niyogi13}
Niyogi, P.
\newblock Manifold regularization and semi-supervised learning: some
  theoretical analyses.
\newblock \emph{Journal of Machine Learning Research}, 14\penalty0
  (1):\penalty0 1229--1250, 2013.

\bibitem[NT \& Maehara(2019)NT and Maehara]{DBLP:journals/corr/abs-1905-09550}
NT, H. and Maehara, T.
\newblock Revisiting graph neural networks: All we have is low-pass filters.
\newblock \emph{CoRR}, abs/1905.09550, 2019.

\bibitem[Oono \& Suzuki(2020)Oono and Suzuki]{DBLP:conf/iclr/OonoS20}
Oono, K. and Suzuki, T.
\newblock Graph neural networks exponentially lose expressive power for node
  classification.
\newblock In \emph{8th International Conference on Learning Representations,
  {ICLR} 2020, Addis Ababa, Ethiopia, April 26-30, 2020}. OpenReview.net, 2020.

\bibitem[Page et~al.(1999)Page, Brin, Motwani, and Winograd]{page1999pagerank}
Page, L., Brin, S., Motwani, R., and Winograd, T.
\newblock The pagerank citation ranking: Bringing order to the web.
\newblock Technical report, Stanford InfoLab, 1999.

\bibitem[Pei et~al.(2020)Pei, Wei, Chang, Lei, and
  Yang]{DBLP:conf/iclr/PeiWCLY20}
Pei, H., Wei, B., Chang, K.~C., Lei, Y., and Yang, B.
\newblock Geom-gcn: Geometric graph convolutional networks.
\newblock In \emph{8th International Conference on Learning Representations,
  {ICLR} 2020, Addis Ababa, Ethiopia, April 26-30, 2020}. OpenReview.net, 2020.

\bibitem[Rozemberczki et~al.(2021)Rozemberczki, Allen, and
  Sarkar]{DBLP:journals/compnet/RozemberczkiAS21}
Rozemberczki, B., Allen, C., and Sarkar, R.
\newblock Multi-scale attributed node embedding.
\newblock \emph{Journal of Complex Networks}, 9\penalty0 (2), 2021.

\bibitem[Satorras \& Estrach(2018)Satorras and
  Estrach]{DBLP:conf/iclr/SatorrasE18}
Satorras, V.~G. and Estrach, J.~B.
\newblock Few-shot learning with graph neural networks.
\newblock In \emph{6th International Conference on Learning Representations,
  {ICLR} 2018, Vancouver, BC, Canada, April 30 - May 3, 2018, Conference Track
  Proceedings}. OpenReview.net, 2018.

\bibitem[Sen et~al.(2008)Sen, Namata, Bilgic, Getoor, Gallagher, and
  Eliassi{-}Rad]{DBLP:journals/aim/SenNBGGE08}
Sen, P., Namata, G., Bilgic, M., Getoor, L., Gallagher, B., and Eliassi{-}Rad,
  T.
\newblock Collective classification in network data.
\newblock \emph{{AI} Magazine}, 29\penalty0 (3):\penalty0 93--106, 2008.

\bibitem[Shchur et~al.(2018)Shchur, Mumme, Bojchevski, and
  G{\"{u}}nnemann]{DBLP:journals/corr/abs-1811-05868}
Shchur, O., Mumme, M., Bojchevski, A., and G{\"{u}}nnemann, S.
\newblock Pitfalls of graph neural network evaluation.
\newblock \emph{CoRR}, abs/1811.05868, 2018.

\bibitem[Sindhwani et~al.(2005)Sindhwani, Niyogi, Belkin, and
  Keerthi]{sindhwani2005linear}
Sindhwani, V., Niyogi, P., Belkin, M., and Keerthi, S.
\newblock Linear manifold regularization for large scale semi-supervised
  learning.
\newblock In \emph{Proceedings of the 22nd ICML Workshop on Learning with
  Partially Classified Training Data}, volume~28, 2005.

\bibitem[Slepcev \& Thorpe(2017)Slepcev and
  Thorpe]{DBLP:journals/corr/SlepcevT17}
Slepcev, D. and Thorpe, M.
\newblock Analysis of $p$-laplacian regularization in semi-supervised learning.
\newblock \emph{CoRR}, abs/1707.06213, 2017.

\bibitem[Smola \& Kondor(2003)Smola and Kondor]{DBLP:conf/colt/SmolaK03}
Smola, A.~J. and Kondor, R.
\newblock Kernels and regularization on graphs.
\newblock In Sch{\"{o}}lkopf, B. and Warmuth, M.~K. (eds.), \emph{Computational
  Learning Theory and Kernel Machines, 16th Annual Conference on Computational
  Learning Theory and 7th Kernel Workshop, COLT/Kernel 2003, Washington, DC,
  USA, August 24-27, 2003, Proceedings}, volume 2777 of \emph{Lecture Notes in
  Computer Science}, pp.\  144--158. Springer, 2003.

\bibitem[Thekumparampil et~al.(2018)Thekumparampil, Wang, Oh, and
  Li]{DBLP:journals/corr/abs-1803-03735}
Thekumparampil, K.~K., Wang, C., Oh, S., and Li, L.
\newblock Attention-based graph neural network for semi-supervised learning.
\newblock \emph{CoRR}, abs/1803.03735, 2018.

\bibitem[van~der Maaten \& Hinton(2008)van~der Maaten and
  Hinton]{JMLR:v9:vandermaaten08a}
van~der Maaten, L. and Hinton, G.
\newblock Visualizing data using t-sne.
\newblock \emph{Journal of Machine Learning Research}, 9\penalty0
  (86):\penalty0 2579--2605, 2008.

\bibitem[van Engelen \& Hoos(2020)van Engelen and
  Hoos]{DBLP:journals/ml/EngelenH20}
van Engelen, J.~E. and Hoos, H.~H.
\newblock A survey on semi-supervised learning.
\newblock \emph{Machine Learning}, 109\penalty0 (2):\penalty0 373--440, 2020.

\bibitem[Velickovic et~al.(2018)Velickovic, Cucurull, Casanova, Romero,
  Li{\`{o}}, and Bengio]{DBLP:conf/iclr/VelickovicCCRLB18}
Velickovic, P., Cucurull, G., Casanova, A., Romero, A., Li{\`{o}}, P., and
  Bengio, Y.
\newblock Graph attention networks.
\newblock In \emph{6th International Conference on Learning Representations,
  {ICLR} 2018, Vancouver, BC, Canada, April 30 - May 3, 2018}, 2018.

\bibitem[Velickovic et~al.(2019)Velickovic, Fedus, Hamilton, Li{\`{o}}, Bengio,
  and Hjelm]{DBLP:conf/iclr/VelickovicFHLBH19}
Velickovic, P., Fedus, W., Hamilton, W.~L., Li{\`{o}}, P., Bengio, Y., and
  Hjelm, R.~D.
\newblock Deep graph infomax.
\newblock In \emph{7th International Conference on Learning Representations,
  {ICLR} 2019, New Orleans, LA, USA, May 6-9, 2019}. OpenReview.net, 2019.

\bibitem[Verma \& Zhang(2019)Verma and Zhang]{DBLP:conf/kdd/VermaZ19}
Verma, S. and Zhang, Z.
\newblock Stability and generalization of graph convolutional neural networks.
\newblock In \emph{The 25th {ACM} {SIGKDD} International Conference on
  Knowledge Discovery and Data Mining, {KDD} 2019, Anchorage, AK, USA, August
  4-8, 2019}, pp.\  1539--1548, 2019.

\bibitem[von Luxburg(2007)]{DBLP:journals/sac/Luxburg07}
von Luxburg, U.
\newblock A tutorial on spectral clustering.
\newblock \emph{Statistics and Computing}, 17\penalty0 (4):\penalty0 395--416,
  2007.

\bibitem[Wang \& Leskovec(2020)Wang and
  Leskovec]{DBLP:journals/corr/abs-2002-06755}
Wang, H. and Leskovec, J.
\newblock Unifying graph convolutional neural networks and label propagation.
\newblock \emph{CoRR}, abs/2002.06755, 2020.

\bibitem[Wu et~al.(2019)Wu, Jr., Zhang, Fifty, Yu, and
  Weinberger]{DBLP:conf/icml/WuSZFYW19}
Wu, F., Jr., A. H.~S., Zhang, T., Fifty, C., Yu, T., and Weinberger, K.~Q.
\newblock Simplifying graph convolutional networks.
\newblock In \emph{Proceedings of the 36th International Conference on Machine
  Learning, {ICML} 2019, 9-15 June 2019, Long Beach, California, {USA}}, pp.\
  6861--6871, 2019.

\bibitem[Wu et~al.(2021)Wu, Pan, Chen, Long, Zhang, and
  Yu]{DBLP:journals/tnn/WuPCLZY21}
Wu, Z., Pan, S., Chen, F., Long, G., Zhang, C., and Yu, P.~S.
\newblock A comprehensive survey on graph neural networks.
\newblock \emph{{IEEE} Transactions on Neural Networks and Learning System},
  32\penalty0 (1):\penalty0 4--24, 2021.

\bibitem[Xinyi \& Chen(2019)Xinyi and Chen]{DBLP:conf/iclr/XinyiC19}
Xinyi, Z. and Chen, L.
\newblock Capsule graph neural network.
\newblock In \emph{7th International Conference on Learning Representations,
  {ICLR} 2019, New Orleans, LA, USA, May 6-9, 2019}. OpenReview.net, 2019.

\bibitem[Xu et~al.(2018)Xu, Li, Tian, Sonobe, Kawarabayashi, and
  Jegelka]{DBLP:conf/icml/XuLTSKJ18}
Xu, K., Li, C., Tian, Y., Sonobe, T., Kawarabayashi, K., and Jegelka, S.
\newblock Representation learning on graphs with jumping knowledge networks.
\newblock In \emph{Proceedings of the 35th International Conference on Machine
  Learning, {ICML} 2018, Stockholmsm{\"{a}}ssan, Stockholm, Sweden, July 10-15,
  2018}, volume~80 of \emph{Proceedings of Machine Learning Research}, pp.\
  5449--5458. {PMLR}, 2018.

\bibitem[Xu et~al.(2019)Xu, Hu, Leskovec, and Jegelka]{DBLP:conf/iclr/XuHLJ19}
Xu, K., Hu, W., Leskovec, J., and Jegelka, S.
\newblock How powerful are graph neural networks?
\newblock In \emph{7th International Conference on Learning Representations,
  {ICLR} 2017 New Orleans, LA, USA, May 6-9, 2019, Conference Track
  Proceedings}, 2019.

\bibitem[Ying et~al.(2018)Ying, You, Morris, Ren, Hamilton, and
  Leskovec]{DBLP:conf/nips/YingY0RHL18}
Ying, Z., You, J., Morris, C., Ren, X., Hamilton, W.~L., and Leskovec, J.
\newblock Hierarchical graph representation learning with differentiable
  pooling.
\newblock In Bengio, S., Wallach, H.~M., Larochelle, H., Grauman, K.,
  Cesa{-}Bianchi, N., and Garnett, R. (eds.), \emph{Advances in Neural
  Information Processing Systems 31, {NeurIPS} 2018, December 3-8, 2018,
  Montr{\'{e}}al, Canada}, pp.\  4805--4815, 2018.

\bibitem[Ying et~al.(2019)Ying, Bourgeois, You, Zitnik, and
  Leskovec]{DBLP:conf/nips/YingBYZL19}
Ying, Z., Bourgeois, D., You, J., Zitnik, M., and Leskovec, J.
\newblock Gnnexplainer: Generating explanations for graph neural networks.
\newblock In \emph{Advances in Neural Information Processing Systems 32,
  {NeurIPS} 2019, 8-14 December 2019, Vancouver, BC, Canada}, pp.\  9240--9251,
  2019.

\bibitem[Zeng et~al.(2020)Zeng, Zhou, Srivastava, Kannan, and
  Prasanna]{DBLP:conf/iclr/ZengZSKP20}
Zeng, H., Zhou, H., Srivastava, A., Kannan, R., and Prasanna, V.~K.
\newblock Graphsaint: Graph sampling based inductive learning method.
\newblock In \emph{8th International Conference on Learning Representations,
  {ICLR} 2020, Addis Ababa, Ethiopia, April 26-30, 2020}. OpenReview.net, 2020.

\bibitem[Zhou \& Sch{\"{o}}lkopf(2005)Zhou and
  Sch{\"{o}}lkopf]{DBLP:conf/dagm/ZhouS05}
Zhou, D. and Sch{\"{o}}lkopf, B.
\newblock Regularization on discrete spaces.
\newblock In \emph{The 27th {DAGM} Symposium, Vienna, Austria, August 31 -
  September 2, 2005}, volume 3663, pp.\  361--368, 2005.

\bibitem[Zhou et~al.(2003)Zhou, Bousquet, Lal, Weston, and
  Sch{\"{o}}lkopf]{DBLP:conf/nips/ZhouBLWS03}
Zhou, D., Bousquet, O., Lal, T.~N., Weston, J., and Sch{\"{o}}lkopf, B.
\newblock Learning with local and global consistency.
\newblock In Thrun, S., Saul, L.~K., and Sch{\"{o}}lkopf, B. (eds.),
  \emph{Advances in Neural Information Processing Systems 16, {NIPS} 2003,
  December 8-13, 2003, Vancouver and Whistler, British Columbia, Canada]}, pp.\
   321--328. {MIT} Press, 2003.

\bibitem[Zhou et~al.(2020)Zhou, Cui, Hu, Zhang, Yang, Liu, Wang, Li, and
  Sun]{DBLP:journals/aiopen/ZhouCHZYLWLS20}
Zhou, J., Cui, G., Hu, S., Zhang, Z., Yang, C., Liu, Z., Wang, L., Li, C., and
  Sun, M.
\newblock Graph neural networks: {A} review of methods and applications.
\newblock \emph{{AI} Open}, 1:\penalty0 57--81, 2020.

\bibitem[Zhou \& Belkin(2011)Zhou and Belkin]{DBLP:journals/jmlr/ZhouB11}
Zhou, X. and Belkin, M.
\newblock Semi-supervised learning by higher order regularization.
\newblock In Gordon, G.~J., Dunson, D.~B., and Dud{\'{\i}}k, M. (eds.),
  \emph{Proceedings of the Fourteenth International Conference on Artificial
  Intelligence and Statistics, {AISTATS} 2011, Fort Lauderdale, USA, April
  11-13, 2011}, volume~15 of \emph{{JMLR} Proceedings}, pp.\  892--900.
  JMLR.org, 2011.

\bibitem[Zhu et~al.(2020)Zhu, Yan, Zhao, Heimann, Akoglu, and
  Koutra]{DBLP:conf/nips/ZhuYZHAK20}
Zhu, J., Yan, Y., Zhao, L., Heimann, M., Akoglu, L., and Koutra, D.
\newblock Beyond homophily in graph neural networks: Current limitations and
  effective designs.
\newblock In \emph{Advances in Neural Information Processing Systems 33,
  {NeurIPS} 2020, December 6-12, 2020, virtual}, 2020.

\bibitem[Zhu et~al.(2021)Zhu, Rossi, Rao, Mai, Lipka, Ahmed, and
  Koutra]{DBLP:conf/aaai/ZhuR0MLAK21}
Zhu, J., Rossi, R.~A., Rao, A., Mai, T., Lipka, N., Ahmed, N.~K., and Koutra,
  D.
\newblock Graph neural networks with heterophily.
\newblock In \emph{35th {AAAI} Conference on Artificial Intelligence, {AAAI}
  2021, Virtual Event, February 2-9, 2021}, pp.\  11168--11176. {AAAI} Press,
  2021.

\bibitem[Zhu et~al.(2003)Zhu, Ghahramani, and Lafferty]{DBLP:conf/icml/ZhuGL03}
Zhu, X., Ghahramani, Z., and Lafferty, J.~D.
\newblock Semi-supervised learning using gaussian fields and harmonic
  functions.
\newblock In Fawcett, T. and Mishra, N. (eds.), \emph{Machine Learning,
  Proceedings of the Twentieth International Conference, {ICML} 2003, August
  21-24, 2003, Washington, DC, {USA}}, pp.\  912--919. {AAAI} Press, 2003.

\bibitem[Zitnik \& Leskovec(2017)Zitnik and
  Leskovec]{DBLP:journals/bioinformatics/ZitnikL17}
Zitnik, M. and Leskovec, J.
\newblock Predicting multicellular function through multi-layer tissue
  networks.
\newblock \emph{Bioinform.}, 33\penalty0 (14):\penalty0 i190--i198, 2017.

\end{thebibliography}
